\documentclass{article}

\usepackage[preprint,nonatbib]{neurips_2025}
\usepackage{orcidlink}

\usepackage[utf8]{inputenc}   %
\usepackage[T1]{fontenc}       %
\usepackage{hyperref}          %
\usepackage{url}               %
\usepackage{booktabs}          %
\usepackage{multirow}          %
\usepackage{makecell}          %
\usepackage{adjustbox}         %
\usepackage{array}             %
\usepackage{subcaption}        %
\usepackage{amsfonts}          %
\usepackage{amsmath}           %
\usepackage{amssymb}           %
\usepackage{nicefrac}          %
\usepackage{microtype}         %
\usepackage{xcolor}            %
\usepackage{colortbl}          %
\usepackage{graphicx}          %
\usepackage{tikz}              %
\usetikzlibrary{arrows.meta, positioning, calc}

\usepackage{snapshot} %

\definecolor{scoreblue}{RGB}{80,160,255}
\definecolor{arrowcolor}{RGB}{60,60,60}
\usepackage[square,numbers]{natbib} %

\definecolor{LightGray}{gray}{0.97}
\definecolor{jupgreen}{RGB}{0,128,0}        %
\definecolor{jupred}{RGB}{186,33,33}         %
\definecolor{jupcomment}{RGB}{64,128,128}    %
\definecolor{jupbuiltin}{RGB}{0,128,0}       %
\definecolor{codegray}{rgb}{0.5,0.5,0.5}

\usepackage{listings}              %
\lstdefinestyle{mystyle}{
  backgroundcolor=\color{LightGray},
  commentstyle=\itshape\color{jupcomment},
  keywordstyle=\bfseries\color{jupgreen},
  numberstyle=\tiny\color{codegray},
  stringstyle=\color{jupred},
  basicstyle=\ttfamily\small,
  breakatwhitespace=false,
  breaklines=true,
  captionpos=b,
  keepspaces=true,
  numbers=left,
  numbersep=5pt,
  showspaces=false,
  showstringspaces=false,
  showtabs=false,
  tabsize=2,
  frame=lines,
  framesep=2mm,
  xleftmargin=1em,
  xrightmargin=1em,
  basewidth=0.5em
}
\newcommand{\conditionalpagebreak}{%
  \ifdim\dimexpr\pagegoal-\pagetotal<0.25\textheight
    \pagebreak
  \fi
}

\title{GenSBI: Generative Methods for Simulation-Based Inference in JAX}

\author{%
  Aurelio Amerio{\hypersetup{hidelinks}\orcidlink{0000-0001-9083-0910}} \\
  Instituto de F\'isica Corpuscular (IFIC) \\
  Universitat de Val\`encia \& CSIC \\
  Val\`encia, Spain \\
  \texttt{aure.amerio@gmail.com}
}

\newcommand{\panel}[1]{\textit{#1}}
\newcommand{\bestcell}[1]{\cellcolor{gray!20}\bfseries\boldmath #1}
\newcommand{\secondcell}[1]{$\textit{#1}$}

\newcolumntype{C}{>{\centering\arraybackslash}p{0.085\textwidth}}
\newcolumntype{M}{>{\raggedright\arraybackslash}p{0.16\textwidth}}

\begin{document}

\pagenumbering{roman}

\maketitle

\begin{abstract}
  Flow and diffusion generative models have established themselves as widely adopted density estimators for simulation-based inference (SBI), extending naturally from neural posterior estimation to likelihood and joint density estimation. Their principled optimization objectives and freedom from architectural constraints have driven rapid adoption across the natural sciences. Yet the most widely used SBI libraries remain PyTorch-based, leaving researchers who develop their forward models and analysis pipelines in JAX without a native option.

We present GenSBI, an open-source library that implements flow matching, score matching, and denoising diffusion entirely in JAX. The library offers three transformer-based architectures --- SimFormer, Flux1, and a novel Flux1Joint that extends gate-modulated transformer blocks to joint density estimation --- all interchangeable through a unified interface that decouples generative method, neural backbone, and inference mode. GenSBI provides an end-to-end workflow from training through posterior calibration (SBC, TARP, LC2ST) and supports custom architectures with domain-specific embedding networks.

We validate the framework on standard SBI benchmarks, achieving near-ideal mean C2ST scores (0.50--0.56, where 0.50 is ideal) on SBIBM tasks with minimal per-task tuning and well-calibrated posterior coverage across all tested configurations. The code is publicly available at \url{https://github.com/aurelio-amerio/GenSBI}.

\end{abstract}

\newpage
\vspace{0.75em}
{
  \hypersetup{hidelinks}
  \makeatletter
  \renewcommand{\@dotsep}{10000}
  \let\orig@l@section\l@section
  \renewcommand{\l@section}[2]{%
    \vspace{-0.8em}\orig@l@section{#1}{#2}%
  }
  \let\orig@l@subsection\l@subsection
  \renewcommand{\l@subsection}[2]{%
    \vspace{-0.3em}\orig@l@subsection{#1}{#2}%
  }
  \makeatother
  \tableofcontents
}
\vspace{0.75em}
\newpage

\pagenumbering{arabic}

\section{Introduction}
\label{sec:introduction}

Modern physics has produced some of the most sophisticated forward simulators in science.
$N$-body simulations evolve billions of dark-matter particles under gravity and produce synthetic maps of the cosmic web~\cite{Springel:2017illustris, Villaescusa_Navarro:2021camels}.
Monte Carlo event generators propagate the products of high-energy collisions through cathedral-sized particle detectors, reproducing the energy deposits and particle tracks that experimentalists measure~\cite{Cranmer:2019eaq}.
Waveform models predict the gravitational-wave signals emitted by merging black holes and neutron stars with sub-radian phase accuracy~\cite{Dax:2024fmpe, Dax:2021real-time}.
Beyond physics, cell-invasion models simulate how tumour populations spread through tissue~\cite{Carr:2021invasion, Wang:2024cguide}, brain-network models reproduce the electroencephalography rhythms and seizure dynamics recorded from cortex~\cite{Rodrigues:2021hnpe, Hashemi:2023epilepsy}, and compartmental epidemic models tracked the COVID-19 infection waves from contact rates and intervention timings~\cite{Radev:2021outbreakflow}.
In every case, the scientist controls a set of parameters~$\theta$~--- cosmological densities, coupling constants, compact-object masses, tumour invasion rates, epidemic contact rates~--- and the simulator returns synthetic data~$x$ that is often indistinguishable from real observations.
And yet, plugging real data back into these simulators to ask \emph{which parameters produced this observation?} is, in most cases, an open problem.
This section introduces the inference challenge that motivates the field (Section~\ref{sec:intro:sbi}), surveys the current software landscape and identifies the gap that this work fills (Section~\ref{sec:intro:landscape}), and states the contributions of GenSBI together with an outline of the paper (Section~\ref{sec:intro:contributions}).

\subsection{The Simulation-Based Inference Problem}
\label{sec:intro:sbi}

The inverse problem is fundamentally harder than the forward one.
Given a real observation $x_\mathrm{obs}$, we want to infer which parameter values $\theta$ could plausibly have produced it, together with a principled quantification of the associated uncertainty.
In Bayesian statistics, this amounts to computing the posterior distribution
\begin{equation}
    p(\theta \mid x_\mathrm{obs}) = \frac{p(x_\mathrm{obs} \mid \theta)\, p(\theta)}{p(x_\mathrm{obs})},
    \label{eq:bayes}
\end{equation}
where $p(\theta)$ is the prior encoding our knowledge before seeing data, and $p(x|\theta)$ is the likelihood~--- the probability of observing~$x$ given parameters~$\theta$, and $p(x_\mathrm{obs})$ is the evidence.
For models whose likelihood is analytically tractable, the posterior can be explored with standard methods, such as Markov chain Monte Carlo (MCMC)~\cite{Metropolis:1953am, Hastings:1970aa}, Hamiltonian Monte Carlo (HMC)~\cite{Duane:1987de}, microcanonical Monte Carlo~\cite{Robnik:2022bzs}, or nested sampling~\cite{Skilling:2006gxv}.
For the simulators described above, however, the likelihood is defined only implicitly: it can be \emph{sampled} by running the simulator, but it cannot be \emph{evaluated} as a function of~$\theta$ for a given observation.
To see why, consider a cosmological $N$-body simulation. Computing the likelihood of an observed galaxy catalogue would require integrating over every possible realization of the initial density field, every stochastic step in the halo-finding and galaxy-assignment pipeline, and every source of observational noise~--- an integral over a latent space with millions of dimensions.
The same structure recurs across physics: in particle collisions, the detector response introduces billions of random latent variables, making the full, field-level likelihood inaccessible; in gravitational-wave astronomy, the mapping from binary parameters to a noisy strain time-series involves stochastic processes that, similarly, often preclude closed-form likelihood evaluation.
The likelihood is, in short, intractable~\cite{Cranmer:2019eaq}.

Before the advent of neural methods, two classical strategies dominated the landscape of likelihood-free inference, each with well-documented limitations~\cite{Cranmer:2019eaq}.
Approximate Bayesian Computation (ABC)~\cite{Sisson:2018abc} bypasses the likelihood entirely by simulating data from prior draws and retaining only those whose output falls within a small tolerance of the real observation~--- effectively rejection sampling in simulation space.
Classical surrogate-likelihood approaches instead train a parametric emulator~--- a Gaussian process, or the Gaussian approximation of summary statistics used in Synthetic Likelihood~\cite{Wood:2010sl}~--- to stand in for the intractable likelihood, and then run MCMC on this emulator.
Both strategies share a common set of weaknesses.
ABC's acceptance rate drops sharply with the dimensionality of the data, and the choice of summary statistics, distance metric, and tolerance threshold requires careful calibration~\cite{Sisson:2018abc}.
Classical surrogates impose rigid parametric assumptions and must remain globally accurate across the entire region of parameter space explored by the Markov chain; in high dimensions this requirement is difficult to satisfy, and even small approximation errors can bias the posterior in ways that are hard to diagnose~\cite{Cranmer:2019eaq}.
Both approaches also suffer from the curse of dimensionality in their reliance on hand-crafted, low-dimensional summary statistics, discarding information that may be critical for constraining the parameters of interest.

Neural simulation-based inference (SBI) replaces these classical tools with a fundamentally different paradigm~\cite{Cranmer:2019eaq, Greenberg:2019apt, Radev:2020bayesflow}.
The central insight is that the simulated pairs $\{(\theta^{(i)}, x^{(i)})\}_{i=1}^N$ produced during forward modelling already encode the full statistical relationship between parameters and data.
Rather than discarding simulations through an acceptance criterion or approximating the likelihood with a rigid parametric model, one trains a flexible neural density estimator to learn the target distribution directly from the simulated dataset.
Four complementary estimation strategies have emerged within this paradigm.
Neural posterior estimation (NPE)~\cite{Papamakarios:2016ctj, Greenberg:2019apt} learns the posterior $q_\phi(\theta|x) \approx p(\theta|x)$ directly, producing samples in a single network evaluation.
Neural likelihood estimation (NLE)~\cite{Papamakarios:2018zoy} learns the likelihood $q_\phi(x|\theta) \approx p(x|\theta)$ using expressive neural density estimators~--- the modern successor to Synthetic Likelihood, but free of parametric assumptions~--- and feeds it into standard MCMC samplers; because the learned likelihood factorises over independent observations, NLE scales naturally to problems with multiple i.i.d.\ trials and decouples the trained model from the choice of prior~\cite{Deistler:2025sbi_tutorial}.
Neural ratio estimation (NRE)~\cite{Hermans:2019ioj} learns the likelihood-to-evidence ratio, sidestepping explicit density estimation altogether.
More recently, neural joint estimation (NJE)~\cite{uria2013rnade,uria2016neural, Gloeckler:2024simformer} learns the full joint density $p(\theta, x)$, enabling posterior, likelihood, and marginal queries from a single model.
Systematic benchmarks confirm that these neural methods outperform classical ABC and Synthetic Likelihood approaches across a range of tasks, though no single strategy dominates universally~\cite{Lueckmann:2021sbibm}.

Among these, NPE is fully \emph{amortised}: the upfront cost of simulation and training is paid once, and afterward the network can produce posterior estimates for any new observation~$x_\mathrm{obs}$ in a single forward pass (or, for flow-based and diffusion-based estimators, by solving a short ordinary differential equation).
No new simulations, no new MCMC chains, no new likelihood evaluations are required.
This property~--- amortisation~--- is what makes neural SBI transformative for experimental sciences where new observations arrive continuously: a new gravitational-wave event every few days, a new cosmological survey data release, a new LHC collision dataset, a new epidemic surveillance report~--- each requiring a new, well-calibrated posterior estimate.

\subsection{The Current Landscape}
\label{sec:intro:landscape}

The neural SBI paradigm has matured into a well-developed software ecosystem (see Section~\ref{sec:related}).
Among the most widely used tools, the \texttt{sbi} library~\cite{BoeltsDeistler_sbi_2025} is the most comprehensive toolkit, implementing neural posterior estimation (NPE), neural likelihood estimation (NLE), and neural ratio estimation (NRE) within a unified PyTorch~\cite{Paszke:2019pytorch} interface.
Complementary tools serve more specialised needs: \texttt{swyft}~\cite{Miller2022} provides truncated marginal neural ratio estimation for high-dimensional data, and \texttt{BayesFlow}~\cite{Radev:2020bayesflow} focuses on amortised workflows with invertible networks and built-in summary statistics learning, among others.

A common thread across these libraries is the choice of density estimator.
Normalizing flows~--- masked autoregressive flows (MAFs) and neural spline flows (NSFs) in particular~--- have been the default backbone for NPE and NLE since the earliest neural SBI papers~\cite{Greenberg:2019apt}.
Normalizing flows remain the method of choice when explicit, fast likelihood evaluation is required, since they compute exact log-probabilities in a single forward pass.
However, their reliance on bijective architectures limits expressiveness: the Jacobian constraint restricts the class of neural networks that can be used and makes scaling to high-dimensional, structured data difficult.

Flow matching~\cite{Lipman:2023fm} and diffusion models~\cite{Song:2020sde, Karras:2022edm} offer a modern alternative that lifts these architectural constraints entirely.
These methods parameterise velocity fields or score functions with unconstrained architectures~--- including transformers~--- and generate samples by solving ordinary or stochastic differential equations.
They have achieved state-of-the-art sample quality on image generation benchmarks~\cite{Karras:2022edm, Lipman:2023fm} and have been applied to molecular design~\cite{Wang:2025diffmol}, computational biology~\cite{Wang:2024cguide, Dingeldein:2025sbi}, and simulation-based inference across the physical sciences~\cite{Yang:2022diffsurvey, Arruda:2025dsbi}.
Among these continuous-time methods, flow matching stands out: its optimal-transport objective produces straight, nearly-linear probability paths between the prior and the posterior, which reduce numerical integration error and enable accurate sampling with fewer ODE steps than the curved stochastic trajectories of diffusion models~\cite{Lipman:2023fm, Dax:2024fmpe}.
In practice, flow matching tends to produce mass-covering posteriors, while diffusion models trained with likelihood weighting offer stronger formal statistical guarantees through the variational lower bound; the two families are therefore complementary, and their relative strengths are compared in detail in Section~\ref{sec:density_estimator_choice}.
Beyond posterior estimation, the same generative models can serve as neural emulators: conditioning on parameters and sampling the learned distribution yields synthetic observations drawn from an approximation of the forward model $p(x|\theta)$, without running the original simulator.
Furthermore, exact log-likelihood evaluation is available through the continuous change-of-variables formula applied to the probability flow ODE, enabling neural likelihood estimation (NLE) workflows within the same framework.
In practice, however, each likelihood evaluation requires solving an ODE, so for inference pipelines that couple density evaluation tightly with MCMC --- where the likelihood is queried at every chain step --- normalizing flows retain an efficiency advantage thanks to their single-pass evaluation.
Neural likelihood estimation, whether backed by normalizing flows for fast single-pass evaluation or by the more expressive flow matching and diffusion architectures, remains a compelling and actively studied alternative to direct posterior estimation~\cite{Lueckmann:2021sbibm, Deistler:2025sbi_tutorial}.
Recent work on joint estimation further extends these advantages: the SimFormer~\cite{Gloeckler:2024simformer} uses a transformer-based diffusion model with random masking to learn all conditionals of the joint distribution simultaneously. Replacing the underlying score model with a flow matching objective is a natural next step, and two independent efforts pursued it concurrently: OneFlowSBI~\cite{Nautiyal:2026oneflowsbi} and GenSBI's \texttt{Flux1Joint} architecture, both released in January 2026. The two implementations share the same masked conditional flow matching formulation but differ in the network backbone: OneFlowSBI uses a lightweight residual MLP, while \texttt{Flux1Joint} uses a transformer whose self-attention mechanism enables explicit pairwise interactions between all variables in the joint sequence; Section~\ref{sec:nn_architectures} provides a detailed comparison.

Despite these advances, the existing SBI ecosystem is almost exclusively built on PyTorch.
Across the JAX~\cite{Bradbury:2018jax} ecosystem~--- which offers composable function transformations (\texttt{jit}, \texttt{vmap}, \texttt{grad}), native TPU support, and tight integration with tools such as BlackJAX~\cite{Cabezas:2024blackjax} for MCMC and NumPyro~\cite{Phan:2019numpyro} for probabilistic programming~--- no library exists that provides flow matching and diffusion-based neural posterior estimation.
\texttt{sbijax}~\cite{Dirmeier:2024sbijax} implements a subset of SBI methods in JAX, but does not support the modern transformer-based architectures or the full range of generative formulations that have proven effective for density estimation.

\subsection{Contributions and Paper Outline}
\label{sec:intro:contributions}

We present \texttt{GenSBI}, an open-source, JAX-native library for simulation-based inference using flow matching and diffusion models, built on the \texttt{Flax}  ~\cite{Flax:2024} neural network framework.
GenSBI fills the gap identified above by providing a modular, extensible framework for neural posterior estimation that operates entirely within the JAX ecosystem.
The main contributions of this work are:

\begin{enumerate}
    \item \textbf{Three generative formulations.} GenSBI implements flow matching, score matching, and denoising diffusion (EDM) as interchangeable density estimation methods under a unified strategy-pattern interface. The same neural network backbone can be used across all three methods with no architectural modifications.
    
    \item \textbf{State-of-the-art transformer architectures.} The library provides three transformer-based models: \texttt{SimFormer} (adapted from Gloeckler et al.~\cite{Gloeckler:2024simformer}), \texttt{Flux1} (adapted from the FLUX.1 architecture~\cite{BlackForestLabs:2025flux} and repurposed for arbitrary data), and \texttt{Flux1Joint}~--- a new architecture that combines Flux1's single-stream gate modulation with SimFormer's masking mechanism, enabling joint density estimation and post-training conditioning on any subset of variables.
    
    \item \textbf{Built-in calibration diagnostics.} GenSBI includes simulation-based calibration (SBC), TARP with Jeffreys confidence intervals, LC2ST, and marginal coverage tests as first-class components. For scientific applications, well-calibrated posteriors are non-negotiable, and these diagnostics are integrated rather than treated as afterthoughts.
    
    \item \textbf{Benchmark validation.} We demonstrate that GenSBI achieves well-calibrated posteriors and high sample quality on standard SBI benchmark tasks (Section~\ref{sec:benchmarks}).
\end{enumerate}

The remainder of this paper is organised as follows.
Section~\ref{sec:sbi} defines the inference problem formally and introduces neural density estimation as the solution paradigm.
Section~\ref{sec:generative_models} provides a self-contained exposition of the three generative modelling frameworks implemented in GenSBI, following the historical development of the field.
Section~\ref{sec:software} describes the library's software architecture and features.
Section~\ref{sec:benchmarks} validates GenSBI on standard SBI benchmark tasks.
Section~\ref{sec:related} positions GenSBI relative to existing tools, and Section~\ref{sec:conclusion} concludes with current limitations and future directions.

\section{Simulation-Based Inference}
\label{sec:sbi}

The previous section motivated this work from a practical standpoint: many scientific models rely on complex simulators whose likelihoods cannot be evaluated analytically, and existing tools for neural SBI are almost exclusively built on PyTorch. We now turn to the formal setting that underpins simulation-based inference and the density estimation methods that make it tractable.

This section proceeds in three steps. Section~\ref{sec:inference_problem} states the inference problem in precise terms and identifies the assumptions under which it can be addressed with neural networks. Section~\ref{sec:nde_sbi} introduces neural density estimation as the overarching paradigm and describes four estimation strategies --- neural posterior estimation, neural likelihood estimation, neural ratio estimation, and neural joint estimation --- together with their respective tradeoffs; of these, GenSBI implements NPE, NLE, and NJE, while NRE is discussed for completeness. Section~\ref{sec:density_estimator_choice} then motivates the choice of flow matching and diffusion models as the generative backbone, contrasting them with the normalizing flows that have historically dominated the field.

\subsection{The Inference Problem}
\label{sec:inference_problem}

We formalise the setting introduced in Section~\ref{sec:intro:sbi}. Consider a parametric model of a system, specified by a vector of parameters $\theta \in \Theta$ and a stochastic simulator that produces synthetic observations $x \in \mathcal{X}$. The simulator defines an implicit generative model: given a parameter value $\theta$ and a random seed, it executes a deterministic program augmented with internal stochastic sampling steps and returns a synthetic observation $x$. Repeated calls with the same $\theta$ but different seeds produce different realisations, so the simulator implicitly defines a conditional distribution $p(x|\theta)$ over observations. Together with a prior $p(\theta)$ encoding our beliefs about the parameters before any data are observed, this yields a joint probability model
\begin{equation}
  p(\theta, x) = p(x \mid \theta)\, p(\theta).
\end{equation}

The quantity $p(x|\theta)$ is the likelihood --- the probability of the simulator producing observation $x$ when run at parameters $\theta$. Given an actual observation $x_\mathrm{obs}$, the goal of Bayesian inference is to compute the posterior distribution
\begin{equation}
\label{eq:posterior}
  p(\theta \mid x_\mathrm{obs})
    = \frac{p(x_\mathrm{obs} \mid \theta)\, p(\theta)}{p(x_\mathrm{obs})},
  \qquad
  p(x_\mathrm{obs})
    = \int p(x_\mathrm{obs} \mid \theta)\,p(\theta)\,d\theta,
\end{equation}
which updates our knowledge of the parameters in light of the data. The denominator $p(x_\mathrm{obs})$~--- the evidence~--- is a normalisation constant which can in principle be obtained by integrating the likelihood against the prior. In most practical settings this integral is itself intractable, but many inference algorithms (MCMC, for example) require likelihood evaluations only up to a normalisation constant, so the evidence need not be computed explicitly.

The challenge specific to simulation-based inference is more fundamental: it is not the evidence that is unavailable, but the likelihood itself. Standard Bayesian methods~--- Markov chain Monte Carlo, Hamiltonian Monte Carlo, variational inference, nested sampling~--- all share a common requirement: the ability to evaluate the likelihood $p(x|\theta)$ pointwise at any proposed parameter value \cite{Diggle:1984implicit, Cranmer:2019eaq}. In the simulator-based setting this evaluation is unavailable. The simulator can generate independent draws $x \sim p(x|\theta)$ for any given $\theta$, but the density $p(x|\theta)$ cannot be written down or computed as a function. Internally, the simulator may marginalise over high-dimensional latent variables, perform numerical integrations whose intermediate states are discarded, or follow branching logic paths that depend on the random seed.

As a concrete example, consider inferring cosmological parameters from observations of how matter is distributed across the universe. To predict the observable universe for a given set of these parameters, cosmologists run gravitational $N$-body simulations. These codes start from a random initial configuration of matter~--- a snapshot of the tiny density fluctuations present shortly after the Big Bang~--- and evolve it forward in time under gravity, producing a synthetic map of the present-day distribution of galaxies and dark matter \cite{Springel:2017illustris, Villaescusa_Navarro:2021camels}. The key difficulty is that the initial configuration is itself a high-dimensional random variable: for any fixed cosmological parameters, infinitely many starting configurations are equally valid, and each leads to a different present-day map. The simulator handles this by drawing a fresh random realisation of the initial conditions at each run, effectively marginalising over them. As a result, the same cosmological parameters produce different outputs on every call, and the relationship between input parameters and output map is mediated by a chain of non-linear gravitational evolution that cannot be inverted or expressed in closed form. The cumulative effect of these operations defines a valid probability distribution over observations, but one that exists only as a sampling procedure, not as an evaluable function.
The same structure arises beyond physics: compartmental epidemic models define their likelihood only implicitly through simulation of disease dynamics~\cite{Radev:2021outbreakflow}, and stochastic cell-invasion models marginalise over random proliferation and migration events at the cell level~\cite{Carr:2021invasion}.

This intractability rules out the direct application of standard Bayesian methods and has historically motivated two classes of workarounds. Approximate Bayesian Computation (ABC) \cite{Sisson:2018abc} bypasses the likelihood entirely by simulating data from prior draws and retaining only those parameter values whose simulations are sufficiently close to the observation. ABC is broadly applicable~--- any simulator can be used without modification~--- but its acceptance rate falls sharply with the dimensionality of the data, and the choice of summary statistics, distance metric, and tolerance threshold requires careful calibration that can introduce bias \cite{Sisson:2018abc, Cranmer:2019eaq}. Classical surrogate-likelihood approaches~--- Synthetic Likelihood with a Gaussian approximation of the summary statistics \cite{Wood:2010sl}, Gaussian process emulators, and similar parametric regression models~--- instead train an emulator to approximate $p(x|\theta)$ and then run MCMC on this emulator. These methods can work well in low-dimensional problems, but they impose rigid parametric assumptions on the likelihood surface, must remain accurate across the entire parameter space explored by the Markov chain, and in some formulations require new simulations at every MCMC step; in high dimensions, satisfying these requirements becomes exponentially harder \cite{Cranmer:2019eaq}.

\begin{figure}[t]
  \centering
  \includegraphics[width=\textwidth]{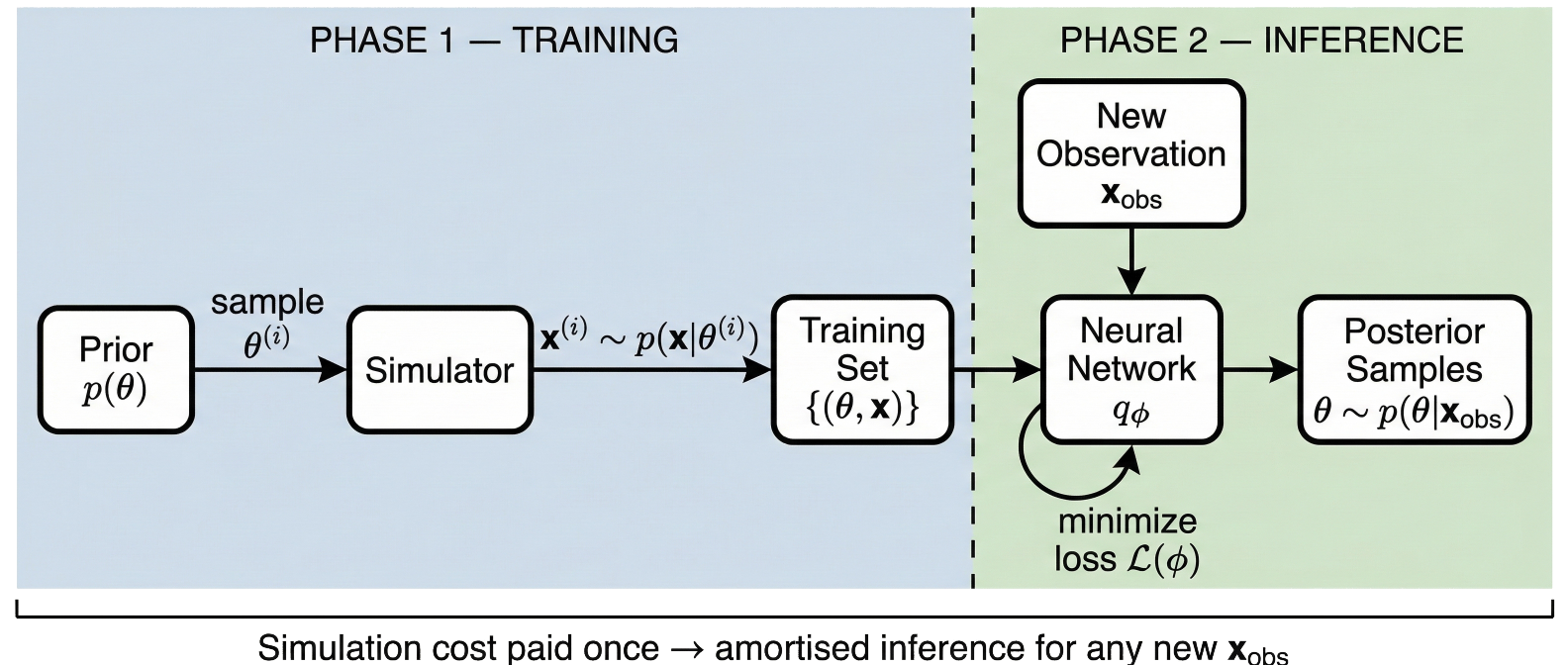}
  \caption{The simulation-based inference pipeline. \textbf{Phase~1 (Training):} parameters are sampled from the prior $p(\theta)$, passed through the simulator to generate synthetic observations, and the resulting pairs $\{(\theta^{(i)}, x^{(i)})\}$ are used to train a neural density estimator $q_\phi$ by minimising a suitable loss $\mathcal{L}(\phi)$. \textbf{Phase~2 (Inference):} given a new observation $x_\mathrm{obs}$, the trained network produces posterior samples directly, with no further simulations required. The simulation cost is paid once; inference for any subsequent observation is amortised.}
  \label{fig:sbi_pipeline}
\end{figure}

The neural SBI paradigm~--- the approach pursued in this work~--- takes a fundamentally different path. Rather than filtering simulations through an acceptance criterion or approximating the likelihood with a rigid parametric model, it uses the forward-simulated pairs $\{(\theta^{(i)}, x^{(i)})\}_{i=1}^N$ as a supervised training set for expressive neural networks that learn tractable density approximations directly from data \cite{Papamakarios:2016ctj, Greenberg:2019apt, Cranmer:2019eaq}. The training data are generated once: parameters are sampled from the prior, $\theta^{(i)} \sim p(\theta)$, and each is passed through the simulator to produce a corresponding observation, $x^{(i)} \sim p(x|\theta^{(i)})$. The network then learns the statistical relationship between parameters and observations from these examples. Which density the network targets~--- the posterior, the likelihood, the likelihood-to-evidence ratio, or the full joint~--- defines the estimation strategy, as described in detail in Section~\ref{sec:nde_sbi}. In all cases the heavy computational cost of simulation is paid once during training; the trained model then yields estimates for any new observation without further simulator calls~--- a property known as amortisation, discussed in Section~\ref{sec:amortisation}. The complete pipeline is illustrated in Figure~\ref{fig:sbi_pipeline}.

The following subsection surveys these estimation strategies and their respective tradeoffs.

\subsection{Neural Density Estimation for SBI}
\label{sec:nde_sbi}

\begin{figure}[t]
  \centering
  \includegraphics[width=\textwidth]{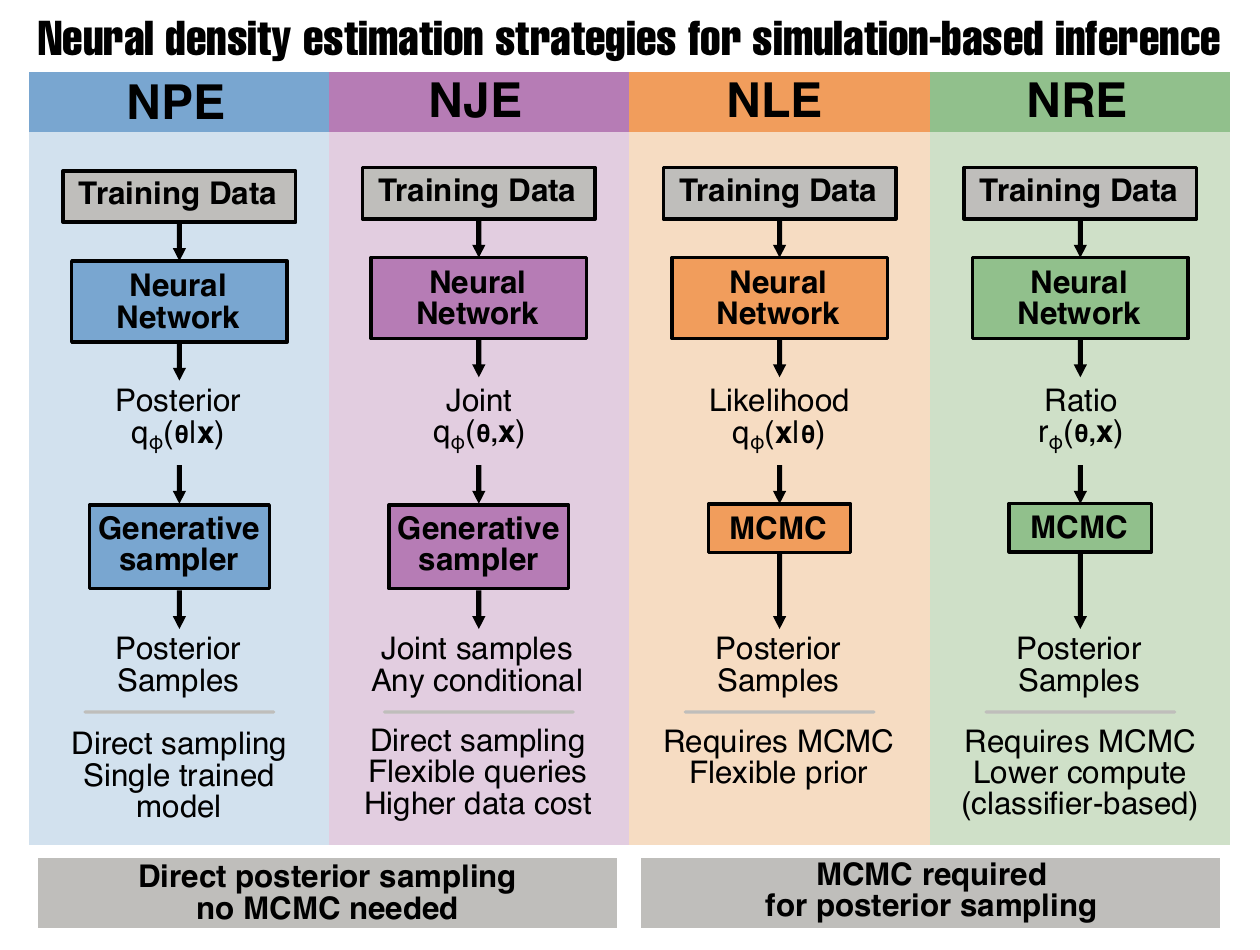}
  \caption{Comparison of the four neural density estimation strategies for simulation-based inference. In every case a neural network learns a tractable surrogate for an otherwise intractable quantity~--- the posterior (NPE), the joint (NJE), the likelihood (NLE), or the likelihood-to-evidence ratio (NRE). The strategies differ in this target and in how posterior samples are then obtained: NPE and NJE draw them directly from the trained generative model, while NLE and NRE require an additional MCMC step. Each column shows the target quantity, inference flow, and key properties of the corresponding strategy. GenSBI implements NPE, NLE, and NJE; NRE is included in the comparison for completeness.}
  \label{fig:nde_comparison}
\end{figure}

The forward-simulation strategy outlined above~--- sampling $\theta^{(i)} \sim p(\theta)$, running $x^{(i)} \sim \mathrm{Sim}(\theta^{(i)})$, and collecting the pairs into a training set~--- opens the door to a family of methods known collectively as neural density estimation (NDE). The common idea is to train a neural network on these simulated pairs so that it learns a tractable approximation of a target density that would otherwise be inaccessible \cite{Cranmer:2019eaq}. Which density the network targets defines the estimation strategy and shapes the resulting inference workflow. Figure~\ref{fig:nde_comparison} provides a side-by-side overview of the four strategies discussed below.

\subsubsection{Neural Posterior Estimation}
\label{sec:npe}

The most direct approach is to train a conditional generative model $q_\phi(\theta \mid x)$ that approximates the posterior $p(\theta \mid x)$ itself. The network parameters $\phi$ are (usually) optimised by minimising the forward Kullback--Leibler divergence between the true and approximate posteriors, averaged over observations drawn from the joint:
\begin{equation}
\label{eq:npe_loss}
  \mathcal{L}_\mathrm{NPE}(\phi)
    = -\mathbb{E}_{p(\theta, x)}\!\left[\log q_\phi(\theta \mid x)\right].
\end{equation}

This objective is equivalent to maximising the expected conditional log-likelihood of the parameters given the data under the learned model \cite{Papamakarios:2016ctj, Greenberg:2019apt}. When the generative model is a flow matching or diffusion model, the loss takes the form of a conditional score matching or flow matching objective, providing a tractable training signal without explicit density evaluation \cite{Dax:2024fmpe}.

NPE is the primary inference mode in GenSBI and the most widely adopted SBI strategy in practice, owing to several distinctive strengths \cite{Deistler:2025sbi_tutorial}. First, it provides fully amortised inference: once trained, posterior samples for any new observation $x_\mathrm{obs}$ are generated by a single forward pass through the learned generative model, with no further simulations, likelihood evaluations, or MCMC runs required. This makes inference essentially instantaneous~--- typically on the order of milliseconds~--- and eliminates the need for sampler hyperparameter tuning (chain length, step size, warm-up) that NLE and NRE introduce through their MCMC or variational inference components \cite{Cranmer:2019eaq, Deistler:2025sbi_tutorial}. Second, because NPE takes observations $x$ as input, it naturally accommodates high-dimensional data through embedding networks~--- convolutional, recurrent, or attention-based architectures that learn informative summary features end-to-end with the density estimator \cite{Deistler:2025sbi_tutorial}. NLE, by contrast, takes parameters $\theta$ as input and predicts data, which limits its ability to leverage such embeddings for high-dimensional observations, without having to train expensive autoencoders~\cite{Rombach:2022ldm}. Third, the speed of amortised sampling makes NPE uniquely practical for global posterior validation: diagnostics such as simulation-based calibration (SBC) and expected coverage tests, which require full posterior inference on hundreds or thousands of calibration datasets, become computationally efficient \cite{Deistler:2025sbi_tutorial}. These advantages come at a cost: the training prior is absorbed into the network and cannot be changed without retraining, and if the network interpolates poorly for out-of-distribution observations, the resulting posterior may be silently miscalibrated with no built-in correction mechanism~--- limitations addressed by NLE's alternative design, described next.

\subsubsection{Neural Likelihood Estimation}
\label{sec:nle}

An alternative strategy targets the likelihood rather than the posterior. In neural likelihood estimation (NLE), the network learns a surrogate $q_\phi(x \mid \theta) \approx p(x \mid \theta)$ of the simulator's data-generating process \cite{Papamakarios:2018zoy}. Posterior inference then proceeds by combining the learned likelihood with a prior $p(\theta)$ and running a standard sampling algorithm~--- typically MCMC. The cost is that posterior sampling is no longer amortised; each new observation requires a fresh MCMC run, which can be expensive in high-dimensional parameter spaces. This design, however, offers several advantages over direct posterior estimation.

First, it decouples training from prior specification: because the network approximates $p(x \mid \theta)$ rather than the posterior $p(\theta \mid x)$, the same learned likelihood can be reused with different priors or embedded in hierarchical models without retraining \cite{Cranmer:2019eaq}.

Second, using MCMC to sample from the product $p(\theta)\,q_\phi(x_\mathrm{obs} \mid \theta)$ provides convergence guarantees that amortised posterior estimation lacks. Under standard regularity conditions, the Markov chain converges to the exact distribution defined by the true prior and the approximate likelihood: the prior is incorporated exactly and without distortion, and the chain's stationarity can be monitored with classical diagnostics \cite{Lueckmann:2021sbibm, Deistler:2025sbi_tutorial}. By contrast, an amortised NPE network absorbs the prior into its parameters during training, and if it interpolates or extrapolates poorly, the resulting posterior may be silently miscalibrated with no built-in correction mechanism \cite{Cranmer:2019eaq}.

Third, when the data consist of multiple independent and identically distributed observations~--- for example, repeated experimental trials~--- NLE gains a natural composability advantage. Because the joint likelihood factorises, the total log-likelihood for $N$ observations can be computed by summing the single-observation log-likelihoods estimated by the pretrained network: $\log p(\theta \mid x_1, \ldots, x_N) \propto \log p(\theta) + \sum_{i=1}^N \log q_\phi(x_i \mid \theta)$. An NLE model trained on single-trial simulations can therefore perform inference on datasets of arbitrary and varying size without retraining \cite{Boelts:2022mnle, Deistler:2025sbi_tutorial}. NPE, by contrast, must be trained on sets of $N$ observations per parameter value, increasing the simulation cost by a factor of $N_\mathrm{max}$ and requiring permutation-invariant architectures to handle the set input \cite{Deistler:2025sbi_tutorial}.

Fourth, because NLE provides an explicit likelihood surrogate, it supports workflows that NPE cannot: Bayesian model comparison via the Bayesian evidence, frequentist inference through maximum likelihood estimation, and direct integration with nested sampling or sequential Monte Carlo methods for evidence computation \cite{Cranmer:2019eaq}.

Normalizing flows~--- bijective transformations that provide approximate likelihoods in a single forward pass through the change-of-variables formula~--- are a natural backbone for NLE: each likelihood query is a single network evaluation, so coupling the learned surrogate with MCMC incurs negligible overhead per chain step. Flow matching and diffusion models can in principle also yield approximate likelihoods via the continuous change-of-variables formula applied to the probability flow ODE, but doing so requires solving the ODE numerically and computing the divergence of the velocity field at each integration step. The resulting evaluation is orders of magnitude slower than a normalizing-flow forward pass, making it impractical as an inner loop within MCMC~--- though promising avenues such as flow distillation and one-step methods may close this gap in the future \cite{Song:2020sde, Lipman:2023fm, Ai:2025f2d2}. GenSBI provides a \texttt{log\_prob} method for approximate likelihood computation via the probability flow ODE, making flow-based NLE possible within the framework. Neural likelihood estimation, whether backed by normalizing flows for fast single-pass evaluation or by the more expressive continuous-time architectures, remains a compelling and actively studied alternative to direct posterior estimation \cite{Lueckmann:2021sbibm, Deistler:2025sbi_tutorial}.

\subsubsection{Neural Ratio Estimation}
\label{sec:nre}

A third paradigm recasts inference as binary classification. Neural ratio estimation (NRE) trains a classifier to distinguish joint samples $(\theta, x)$ drawn from the true data-generating process from marginal samples drawn independently from $p(\theta)\,p(x)$. When the classifier is optimal, its output yields the likelihood-to-evidence ratio $r(\theta, x) = p(x \mid \theta) / p(x)$, which is proportional to the posterior density \cite{Hermans:2019ioj, Cranmer:2019eaq}. By reducing inference to classification, NRE avoids the architectural constraints of density estimation~--- any standard classifier (e.g.\ a residual network) suffices, and training is typically cheaper per step than for the generative models required by NPE or NLE \cite{Deistler:2025sbi_tutorial}. Like NLE, NRE learns a quantity that is independent of the prior, so a trained ratio estimator can be reused with different priors without retraining and handles sequential refinement without the proposal-correction issues of sequential NPE \cite{Cranmer:2019eaq, polyswyft}. NRE also shares NLE's composability for multiple i.i.d.\ observations: individual ratios can be multiplied at inference time, enabling a model trained on single trials to handle datasets of arbitrary size \cite{Deistler:2025sbi_tutorial}. Furthermore, since NRE takes observations $x$ as input, it is compatible with deep embedding networks for high-dimensional data~--- a property it shares with NPE but not with NLE \cite{Deistler:2025sbi_tutorial}. Drawing posterior samples, however, still requires MCMC, as in NLE. NRE is not implemented in GenSBI; we refer to \texttt{sbi} \cite{BoeltsDeistler_sbi_2025} and \texttt{swyft} \cite{Miller2022} for mature implementations.

\subsubsection{Neural Joint Estimation}
\label{sec:nje}

Rather than targeting a single conditional distribution, one can train the network on the full joint density $q_\phi(\theta, x) \approx p(\theta, x)$. Gloeckler et al.\ \cite{Gloeckler:2024simformer} showed that a transformer-based diffusion model trained with random masking over subsets of the joint variables can learn all conditionals and marginals of the joint distribution simultaneously. At inference time, any subset of variables can be treated as observed and the remaining ones sampled from the corresponding conditional, without retraining. Replacing the underlying score model with a flow matching objective is a natural extension of this framework: it preserves the masking mechanism and the any-conditional inference capability while inheriting the straighter probability paths and more efficient ODE integration characteristic of flow matching. This step was pursued independently by OneFlowSBI~\cite{Nautiyal:2026oneflowsbi} and by GenSBI's \texttt{Flux1Joint} architecture. The two implementations share the same masked conditional flow matching formulation; they differ in their network architectures and training details, a comparison discussed in Section~\ref{sec:nn_architectures}. GenSBI additionally supports NJE through the \texttt{SimFormer} architecture. The tradeoff of joint estimation is that learning the full joint is more demanding in data and computation, since the model must capture all possible conditionals rather than a single one.

\subsubsection{Amortisation}
\label{sec:amortisation}

A property shared by the methodologies mentioned in the above sections is amortised inference. Once the network is trained, no new simulations are needed to obtain the posterior for a previously unseen observation: posterior samples are produced by evaluating the learned model at the new $x_\mathrm{obs}$. The simulation and training cost is paid once, after which inference for each new observation is essentially free \cite{Cranmer:2019eaq}. Amortisation also enables powerful validation techniques such as simulation-based calibration (SBC), which require evaluating the posterior for hundreds or thousands of test observations~--- an exercise that would be prohibitively expensive with per-observation methods.

The practical speed of amortised inference, however, depends on the underlying density estimator. Normalizing flows produce samples in a single forward pass, but their architectural constraints (bijective layers) can limit expressiveness. Flow matching and diffusion models are more expressive and scale better to complex, high-dimensional posteriors, but sampling requires solving an ODE or SDE over multiple integration steps. NRE, by contrast, evaluates the posterior density at a given $(\theta, x)$ pair quickly~--- a single forward pass through the classifier~--- yet drawing posterior samples still requires MCMC. Similarly, NLE requires a fresh MCMC run for each observation, which is slower than a single forward pass; in return, MCMC provides asymptotic convergence guarantees, exact incorporation of the prior, and chain-level diagnostics that can flag sampling failures. These complementary strengths make the choice of estimator problem-dependent: when fast sampling is the priority and the posterior is complex, flow matching and diffusion models are a natural choice; when fast density evaluation matters more, normalizing flows or NRE may be preferable. The next section examines these tradeoffs in detail and motivates the design choices underlying GenSBI.

\subsection{Choice of Density Estimator}
\label{sec:density_estimator_choice}

The estimation strategies described in the previous section~--- NPE, NLE, NRE, and NJE~--- are in principle agnostic to the generative model used to parameterise $q_\phi$. In practice, however, the choice of density estimator has a significant impact on expressiveness, training stability, and the computational cost of both sampling and likelihood evaluation. This section reviews the two main families of generative models that have been applied to neural SBI~--- normalizing flows and flow matching / diffusion models~--- and motivates the design choices underlying GenSBI.

Normalizing flows have been the standard density estimator for neural SBI since the early formalisation of NPE and NLE. Greenberg et al.\ \cite{Greenberg:2019apt} and Papamakarios et al.\ \cite{Papamakarios:2018zoy} introduced masked autoregressive flows (MAFs) \cite{Papamakarios:2017tec} as the backbone for posterior and likelihood estimation, respectively, and neural spline flows (NSFs) \cite{Durkan:2019nsq} soon became a popular alternative owing to their greater flexibility. Normalizing flows define an invertible, differentiable map between a simple base distribution (typically a standard Gaussian) and the target distribution. This construction yields exact likelihoods in a single forward pass via the change-of-variables formula --- a single network evaluation, compared with the $\mathcal{O}(50\text{--}100)$ sequential evaluations required to solve the probability flow ODE in flow matching or diffusion models. This cost advantage makes normalizing flows particularly well suited to any workflow dominated by repeated density queries: NLE, where the learned likelihood must be evaluated at every step of an MCMC chain; Bayesian model comparison, where evidence ratios $p(x_\mathrm{obs} \mid M_0) / p(x_\mathrm{obs} \mid M_1)$ must be computed; and posterior density evaluation for downstream optimisation or diagnostics. Flow matching and diffusion models can in principle compute the same quantities exactly through the continuous change-of-variables formula, but they are optimised for high-quality sampling rather than for fast density evaluation, and each log-probability query incurs the cost of a full ODE solve. For low-dimensional problems with smooth, unimodal posteriors ($d \lesssim 10$), neural spline flows in particular handle the distribution capably with less computational overhead and shorter training times than continuous-time methods.

The same bijection requirement that grants normalizing flows their tractable likelihoods, however, also constrains their expressiveness. Every layer must be invertible with a tractable Jacobian determinant, limiting the class of transformations the model can represent. In particular, the bijection requirement precludes flexible, modern architectures such as transformers, which lack the invertibility structure that normalizing flows demand. Normalizing flows can therefore struggle with highly multimodal posteriors or with problems whose dimensionality makes the required number of coupling layers impractical, though promising avenues such as flow distillation and one-step methods may close this gap in the future~\cite{Song:2020sde, Lipman:2023fm, Ai:2025f2d2}.

Flow matching and diffusion models circumvent these limitations by abandoning the requirement of an invertible network altogether. Rather than constructing a bijection, these methods train an unconstrained neural network~--- a score function or a velocity field~--- to characterise a stochastic or deterministic process that maps noise to data \cite{Song:2020sde, Lipman:2023fm}. Because the network itself need not be invertible, any architecture can serve as the backbone, including transformers and other high-capacity models that would be incompatible with normalizing flows. This architectural freedom translates into several practical advantages: more stable training at high dimensions, better mode coverage for complex multimodal distributions, and simpler training objectives~--- flow matching, in particular, replaces the ODE simulation required by early continuous normalizing flows with a simulation-free regression loss \cite{Lipman:2023fm}.

Within this continuous-time family, flow matching and diffusion models differ in important ways. Diffusion models train a score network to reverse a noise-injection process and generate samples by simulating a stochastic or deterministic reverse-time trajectory. These trajectories are typically curved: the score function varies rapidly across noise levels, and accurate integration demands many solver steps. Flow matching instead learns a velocity field that transports probability mass along optimal-transport paths~--- straight-line interpolations between noise and data. The resulting ODE trajectories are nearly linear, which reduces numerical integration error and allows sampling in as few as 10--20 function evaluations~\cite{Lipman:2023fm, Dax:2024fmpe}.

In practice, flow matching tends to produce mass-covering posteriors that assign non-zero probability across the full support of the target distribution~\cite{Dax:2024fmpe}. This tendency arises because the conditional flow matching loss is optimised at the marginal velocity field~--- the conditional expectation over all trajectories connecting noise to data~--- which integrates contributions from every mode, biasing the network toward full-support coverage rather than mode collapse~\cite{Lipman:2024fmguide}. This mechanism is an inductive bias of the training objective, not a formal guarantee: the flow matching regression loss does not minimise or bound the forward Kullback--Leibler divergence for deterministic ODEs, and with finite capacity mode-dropping can occur~\cite{albergo2025stochasticinterpolantsunifyingframework, zammit2025neural, Lipman:2024fmguide}. By contrast, diffusion models trained with likelihood weighting~--- where the loss weight is set to $\lambda(t) = g(t)^2$, making the score matching objective equivalent to an evidence lower bound (ELBO)~--- provide a formal upper bound on the negative log-likelihood through the variational lower bound, and the diffusion term in the SDE naturally regularises the learned distribution, giving stronger theoretical guarantees for mass coverage~\cite{Song:2020sde, albergo2025stochasticinterpolantsunifyingframework}. The distinction has practical consequences: on low-dimensional tasks, SDE-based diffusion models can match or slightly outperform flow matching in statistical accuracy, particularly with limited simulation budgets~\cite{Arruda:2025dsbi}; on high-dimensional problems, flow matching's straighter trajectories and more stable training tend to dominate~\cite{Dax:2024fmpe}. Both families therefore offer complementary strengths, and providing them within a single framework allows practitioners to choose the method best suited to the dimensionality, complexity, and data budget of their inference problem.

These benefits come at a cost. Sampling from a trained flow matching or diffusion model requires integrating an ODE or SDE over multiple solver steps, each involving a forward pass through the neural network. While the resulting samples are typically of high quality (the samples faithfully reproduce the posterior shape, modes, and variance without being easily distinguishable from real data), the process is slower than the single-pass generation of normalizing flows. 
More importantly for inference, computing exact log-likelihoods requires solving the probability flow ODE backward in time and evaluating the divergence of the velocity field at each step~--- a procedure that is orders of magnitude more expensive than the change-of-variables formula \cite{Song:2020sde}. This cost makes flow matching and diffusion models impractical for NLE, where the likelihood must be evaluated at every iteration of an MCMC sampler.

For NPE, the choice of density estimator depends on which operation dominates the inference workflow. When the goal is to \emph{evaluate} the posterior density at given parameter values~--- for instance, to compute Bayes factors or to feed into a downstream optimisation~--- normalizing flows retain a clear advantage, since they provide exact log-probabilities in a single forward pass. When the primary goal is instead to \emph{draw samples} from the posterior, flow matching offers greater expressiveness and, on standard posterior benchmarks, consistently lower classifier two-sample test (C2ST) scores than normalizing-flow-based NPE~\cite{Arruda:2025dsbi, Dax:2024fmpe}. Flow matching's combination of straight optimal-transport paths, high quality samples, empirically mass-covering posteriors, and unconstrained architectures makes it particularly well suited to NPE for complex scientific posteriors. The interest in flow matching and diffusion models for density estimation has grown rapidly: Yang et al.\ \cite{Yang:2022diffsurvey} catalogue over 200 works applying diffusion models across vision, natural language, and scientific domains within two years of the foundational papers. In the SBI context specifically, Dax et al.\ \cite{Dax:2024fmpe} demonstrated that flow matching scales to high-dimensional problems more effectively than normalizing flows, and the SimFormer architecture~\cite{Gloeckler:2024simformer} showed that diffusion-based models can perform joint estimation over parameters and observations within a single transformer. Superior sample quality on generative benchmarks, however, does not automatically guarantee posterior accuracy: the loss parameterisation, noise schedule, and sampler all affect the statistical fidelity of the learned distribution~\cite{Arruda:2025dsbi}, and the output should always be validated with posterior calibration diagnostics.

To summarise, a reliable neural posterior estimator should be mass-covering, computationally efficient to sample from, and robust to numerical choices made during training and inference, producing samples of high quality that are neither over nor under-dispersed. Flow matching and diffusion models address these desiderata in complementary ways. Diffusion models trained with likelihood weighting provide a formal variational bound on the negative log-likelihood that guarantees mass coverage, whereas flow matching achieves broad coverage as an empirical inductive bias of its conditional regression objective. In terms of sampling efficiency and numerical stability, flow matching holds an advantage: its nearly straight optimal-transport trajectories allow accurate integration in 10--50 solver steps, whereas diffusion models typically require more steps and are more sensitive to the noise schedule. Both families share key strengths~--- unconstrained architectural backbones (notably transformers), full amortisation, and the ability to compute approximate log-likelihoods via the probability flow ODE~--- and both require external diagnostics such as simulation-based calibration (SBC) and Tests of Accuracy with Random Points (TARP) for posterior calibration verification, as neither training objective alone guarantees calibrated credible intervals.

GenSBI is built around the strengths of flow matching and diffusion models for sampling-based NPE. A deliberate design choice is the inclusion of both families under a single interface: flow matching serves as the default for its practical efficiency and strong empirical performance, while score matching with likelihood weighting provides a variational upper bound on the negative log-likelihood and EDM offers principled preconditioning-based training~--- complementary strengths that users may prefer when theoretical guarantees or training stability are a priority. The choice should be guided by the estimation strategy, the dimensionality of the problem, and the requirements of the inference task at hand. The following section provides a self-contained introduction to the three generative formulations implemented in GenSBI: score-based models, denoising diffusion with EDM preconditioning, and flow matching.

\section{Generative Models for Density Estimation}
\label{sec:generative_models}

The generative models at the heart of this work draw on an idea with deep roots in physics: the irreversible diffusion of structure into disorder. When a drop of ink disperses in water, the second law of thermodynamics ensures that the complex initial pattern dissolves into a featureless equilibrium---an easy process to run forward, yet seemingly impossible to reverse. Sohl-Dickstein et al.~\cite{Sohl-Dickstein:2015dhe} recognised that this asymmetry could be turned into a generative modeling strategy. By decomposing the destruction of data into a long sequence of small, tractable Gaussian perturbations, the reverse of each individual step remains a simple Gaussian whose parameters can be learned. The resulting framework---\emph{diffusion probabilistic models}---converts the hard problem of sampling from an unknown distribution into many easy denoising steps.

Two parallel lines of work brought this idea to practice. Ho et al.~\cite{Ho:2020epu} introduced denoising diffusion probabilistic models (DDPM), which made diffusion practical for high-quality image generation through a simplified noise-prediction training objective. Independently, Song and Ermon~\cite{Song:2019ncsn} proposed score matching with Langevin dynamics (SMLD), showing that a neural network trained to estimate the \emph{score function}---the gradient of the log-density $\nabla_x \log p(x)$---can guide an iterative refinement process from noise to data without an explicit probabilistic model. Song et al.~\cite{Song:2020sde} later unified both approaches within a single continuous-time framework based on stochastic differential equations (SDEs), revealing that DDPM and SMLD are discretisations of the same underlying process with different noise schedules. Section~\ref{sec:score_models} presents this SDE framework in detail.

\begin{figure}[t]
  \centering
  \resizebox{\textwidth}{!}{\input{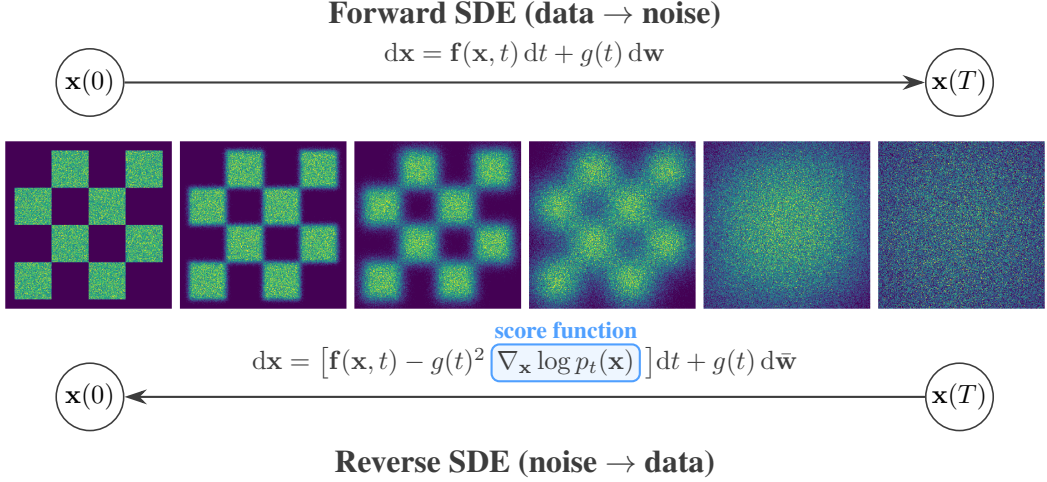}}
  \caption{Schematic of score-based generative modeling. The forward SDE progressively corrupts data into noise; the reverse SDE recovers data by following the learned score function $\nabla_{\mathbf{x}} \log p_t(\mathbf{x})$. The panels show the evolution of a two-dimensional checkerboard distribution under the forward process.}
  \label{fig:SM_schematic}
\end{figure}

\begin{figure}[t]
  \centering
  \resizebox{\textwidth}{!}{\input{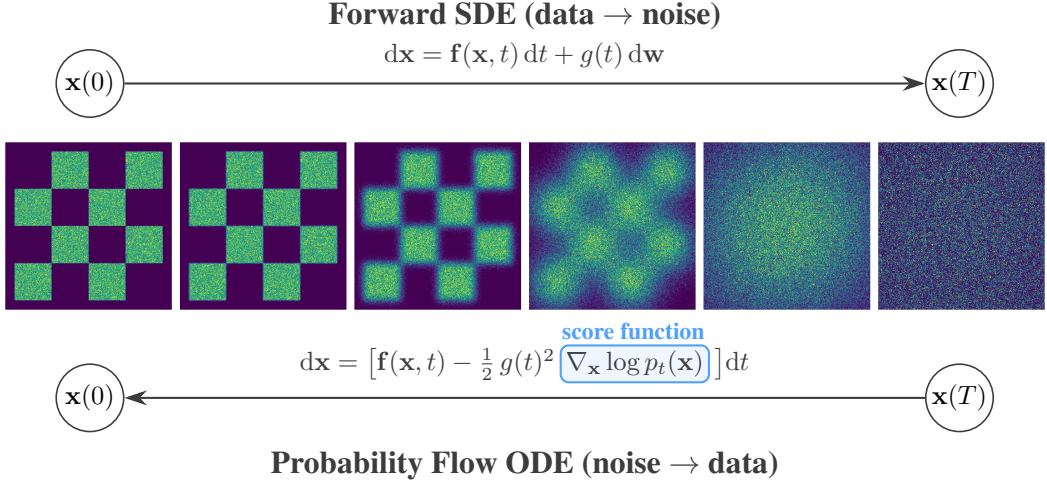}}
  \caption{Schematic of the EDM framework. The forward SDE is the same noise-corruption process as in score matching; the reverse direction is replaced by the deterministic probability flow ODE, whose drift absorbs the stochastic term via a $\tfrac{1}{2}g(t)^2$ factor multiplying the score function. The panels show the forward process on a two-dimensional checkerboard distribution.}
  \label{fig:EDM_schematic}
\end{figure}

In the unified SDE picture, a \emph{forward} stochastic differential equation progressively corrupts data into noise, while a \emph{reverse-time} SDE recovers the data by following the learned score function at every noise level (Figure~\ref{fig:SM_schematic}). A key insight of this framework is that the stochastic reverse process is not the only option: Song et al.~\cite{Song:2020sde} showed that every diffusion SDE has a corresponding deterministic ordinary differential equation---the \emph{probability flow ODE}---whose trajectories trace out the same marginal distributions at every time $t$. This deterministic formulation brings several practical advantages: sampling becomes reproducible for a given initial noise vector, adaptive-step ODE solvers can reduce the number of network evaluations relative to fixed-step SDE discretisations, and the continuous change-of-variables formula enables exact likelihood computation. Karras et al.~\cite{Karras:2022edm} built on this ODE perspective, reformulating diffusion models through a generalised probability flow ODE that encompasses all previously proposed noise schedules---including both the VP (DDPM) and VE (SMLD) variants---as special cases. The resulting framework, commonly known as EDM (from the paper's title, \emph{Elucidating the Design Space of Diffusion-Based Generative Models}), replaces score prediction with a preconditioned denoiser that maps noisy inputs directly back to clean data, and systematically optimises the noise schedule, loss weighting, and network preconditioning around the probability flow ODE (Figure~\ref{fig:EDM_schematic}). Section~\ref{sec:edm} describes the probability flow ODE and the EDM denoising framework.

\begin{figure}[t]
  \centering
  \resizebox{\textwidth}{!}{\input{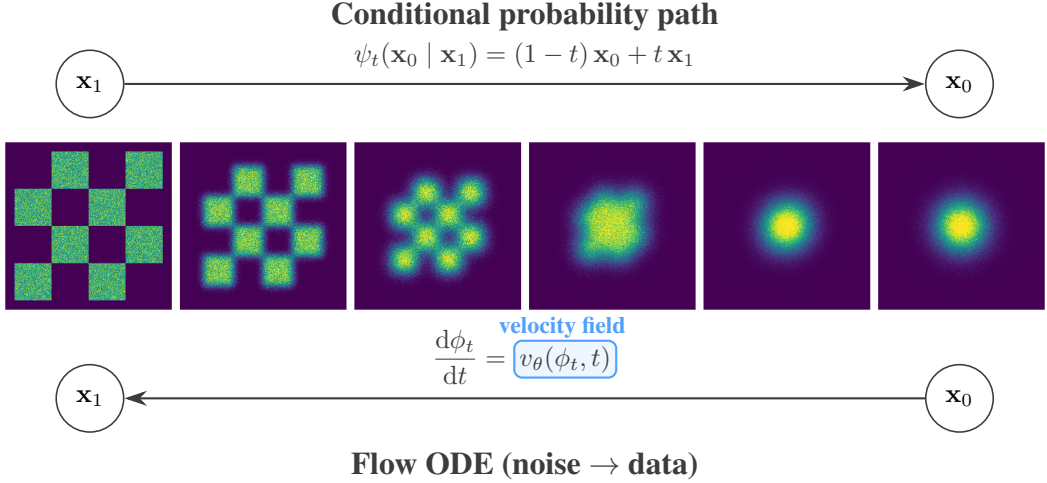}}
  \caption{Schematic of flow matching with the conditional optimal-transport (CondOT) path. The top arrow shows the affine probability path that interpolates between data $\mathbf{x}_1$ and noise $\mathbf{x}_0$; the bottom arrow shows the flow ODE, which transports noise to data by integrating the learned velocity field $v_\theta$. Note the reversed notation relative to diffusion: $\mathbf{x}_0$ is noise and $\mathbf{x}_1$ is data. The panels show the forward (data-to-noise) direction of the same checkerboard distribution.}
  \label{fig:FM_schematic}
\end{figure}

The success of the ODE-based perspective raised a natural question: if the goal is to integrate an ODE that transports noise to data, why derive it from a stochastic diffusion at all? Flow matching~\cite{Lipman:2023fm} answers this by constructing the ODE directly. Rather than learning a score function and plugging it into a probability flow ODE inherited from a diffusion process, flow matching trains a neural network to approximate a \emph{velocity field} that transports a simple prior to the data distribution along prescribed probability paths (Figure~\ref{fig:FM_schematic}). With the conditional optimal-transport path, these trajectories become straight lines, making the ODE easy to integrate and simplifying both training and sampling. Section~\ref{sec:flow_matching} develops the flow matching formalism in full.

This section provides a self-contained introduction to these three generative formulations as implemented in GenSBI: score-based models (Section~\ref{sec:score_models}), denoising diffusion with EDM preconditioning (Section~\ref{sec:edm}), and flow matching (Section~\ref{sec:flow_matching}). Section~\ref{sec:conditional_nde} then shows how all three frameworks extend to conditional density estimation, the bridge from general-purpose generative modeling to simulation-based inference. As discussed in Section~\ref{sec:density_estimator_choice}, both families are well suited for neural posterior estimation; flow matching is currently the preferred default owing to its straighter paths and more efficient ODE integration, while diffusion models offer stronger formal statistical guarantees on convergence.

A note on notation: each subsection follows the conventions established in the corresponding literature. In Sections~\ref{sec:score_models} and~\ref{sec:edm}, data samples are denoted $x(0)$ or $x_0$ and the noisy endpoint is $x(T)$, following the diffusion convention where time runs forward from clean data to noise. In Section~\ref{sec:flow_matching}, the notation is reversed: $x_0$ denotes the source noise and $x_1$ the target data, following the flow matching convention where time runs forward from noise to data. In GenSBI's implementation, all three methods adopt the latter convention ($x_0$ = noise, $x_1$ = data) for internal consistency across the codebase.

\subsection{Score-Based Generative Models}
\label{sec:score_models}

Score-based generative models generate data by learning to reverse a noise-corruption process. The idea, introduced by Sohl-Dickstein et al.\ \cite{Sohl-Dickstein:2015dhe} and refined by Song and Ermon \cite{Song:2019ncsn, Song:2020ncsnpp}, is to define a forward process that gradually transforms a data sample into unstructured noise, and then train a neural network to undo each step of that transformation. Song et al.\ \cite{Song:2020sde} placed both earlier formulations --- denoising diffusion probabilistic models (DDPM) \cite{Ho:2020epu} and score matching with Langevin dynamics (SMLD) \cite{Song:2019ncsn} --- within a single continuous-time framework based on stochastic differential equations (SDEs). This unified perspective is the mathematical foundation of GenSBI's \texttt{ScoreMatchingMethod}.

\paragraph{Forward diffusion as an SDE.}
The forward process perturbs a data point $x(0) \sim p_\mathrm{data}$ by adding noise according to an It\^o SDE (eq.~5 of~\cite{Song:2020sde}),
\begin{equation}
\mathrm{d}x = f(x,t)\,\mathrm{d}t + g(t)\,\mathrm{d}w, \qquad t \in [0, T],
\end{equation}
where $f(x,t)$ is a vector-valued drift coefficient, $g(t)$ is a scalar diffusion coefficient controlling the noise intensity, and $w$ denotes a standard Wiener process. The coefficients are chosen so that the marginal distribution $p_t(x)$ transitions smoothly from the data distribution $p_0(x) = p_\mathrm{data}(x)$ to a tractable prior $p_T(x)$ that is close to a standard Gaussian. Two canonical choices exist:

\begin{itemize}
  \item \textbf{Variance Preserving (VP) SDE} (eq.~11 of~\cite{Song:2020sde})\textbf{:} $\mathrm{d}x = -\tfrac{1}{2}\beta(t)\,x\,\mathrm{d}t + \sqrt{\beta(t)}\,\mathrm{d}w$, where $\beta(t)$ is a noise schedule. The linear drift drives the mean toward zero while the diffusion injects noise, keeping the total variance bounded at one if the initial distribution has unit variance. DDPM \cite{Ho:2020epu} corresponds to a discrete-time approximation of this SDE.
  \item \textbf{Variance Exploding (VE) SDE} (eq.~9 of~\cite{Song:2020sde})\textbf{:} $\mathrm{d}x = \sqrt{\frac{\mathrm{d}[\sigma^2(t)]}{\mathrm{d}t}}\,\mathrm{d}w$, where $\sigma(t)$ is an increasing noise level. The drift is zero, so noise accumulates without bound; the variance explodes as $t \to \infty$. SMLD \cite{Song:2019ncsn} corresponds to a discrete-time approximation of this SDE.
\end{itemize}

In both cases, taking the number of discrete noise levels to infinity reveals DDPM and SMLD as special cases of the same SDE framework \cite{Song:2020sde}. This continuous formulation allows one to work with an arbitrary noise schedule, interpolating between and generalising both earlier approaches.

\paragraph{Score function and denoising score matching.}
The score of the noisy distribution at time $t$ is the gradient $\nabla_x \log p_t(x)$ --- a vector field pointing toward regions of higher probability density \cite{Song:2019ncsn}. Direct estimation of the data score is problematic: the standard score matching objective of Hyv\"arinen \cite{Hyvarinen:2005asm} requires computing the trace of the neural network's Jacobian, which scales poorly to high-dimensional inputs. Denoising score matching \cite{Vincent:2011dsm} avoids this entirely by observing that, for Gaussian perturbations $p_t(x|x_0) = \mathcal{N}(x;\, \mu_t(x_0),\, \sigma_t^2 I)$, the conditional score has a closed-form expression:
\begin{equation}
\nabla_x \log p_t(x | x_0) = -\frac{x - \mu_t(x_0)}{\sigma_t^2}.
\end{equation}
Using the reparameterisation $x = \mu_t(x_0) + \sigma_t \epsilon$ with $\epsilon \sim \mathcal{N}(0, I)$, training reduces to a noise-prediction objective (eq.~7 of~\cite{Song:2020sde}):
\begin{equation}
\mathcal{L}_\mathrm{DSM}(\theta) = \mathbb{E}_{t \sim \mathcal{U}[0,T],\; x_0 \sim p_\mathrm{data},\; \epsilon \sim \mathcal{N}(0,I)} \left[\lambda(t)\,\left\| s_\theta(x_t, t) + \frac{\epsilon}{\sigma_t}\right\|^2 \right],
\end{equation}
where $x_t = \mu_t(x_0) + \sigma_t \epsilon$ is the noisy sample and $s_\theta(x,t)$ is the score network. The weighting function $\lambda(t)$ balances the contribution of different noise levels to the total loss. Because the true score magnitude decreases as $1/\sigma_t$, the unweighted mean-squared-error objective receives diminishing gradient signal at high noise levels; setting $\lambda(t) \propto \sigma_t^2$ compensates for this by upweighting the contribution of large-noise timesteps and stabilises training \cite{Song:2019ncsn, Song:2020ncsnpp}.

The denoising score matching objective is not an ad hoc regression target: it arises from a principled variational argument. The fundamental goal of training a generative model is to minimise the forward Kullback--Leibler divergence $D_{\mathrm{KL}}(p_\mathrm{data} \| p_\theta)$ between the data distribution and the model distribution, which is equivalent to maximising the data log-likelihood. For diffusion models this KL divergence is intractable, but it can be bounded from above by comparing the forward and reverse Markov chains. In the discrete-time formulation of Ho et al.~\cite{Ho:2020epu}, the resulting evidence lower bound (ELBO) decomposes into a sum of KL divergences between the true forward posteriors $q(x_{t-1}|x_t, x_0)$ and the learned reverse transitions $p_\theta(x_{t-1}|x_t)$; because both distributions are Gaussian with fixed variances, each KL term reduces to a squared $L_2$ distance between their means, and reparameterising the predicted mean as a noise prediction yields the familiar denoising objective \cite{Ho:2020epu, Sohl-Dickstein:2015dhe}.

In the continuous-time SDE framework, Song et al.~\cite{Song:2020sde} showed that the same bound can be derived via the Girsanov theorem applied to the path measures of the forward and reverse-time SDEs; the resulting KL divergence between path measures reduces exactly to a time-integrated weighted score matching loss with weighting $\lambda(t) = g(t)^2$. This specific choice --- termed \emph{likelihood weighting} --- makes the continuous denoising score matching objective a strict upper bound on the negative log-likelihood of the generative model \cite{Song:2020sde}. The bound becomes tight when the learned score perfectly matches the true score at every noise level. Other choices of $\lambda(t)$, such as $\lambda(t) \propto \sigma_t^2$, sacrifice the formal likelihood bound in exchange for improved sample quality or training stability, but the underlying loss retains the same functional form.

\paragraph{Reverse-time SDE and sampling.}
Anderson \cite{Anderson:1982rtd} showed that the reverse of any diffusion process is itself a diffusion, governed by (eq.~6 of~\cite{Song:2020sde})
\begin{equation}
\mathrm{d}x = \left[f(x,t) - g(t)^2\,\nabla_x \log p_t(x)\right] \mathrm{d}t + g(t)\,\mathrm{d}\bar{w},
\end{equation}
where $\mathrm{d}t$ is now an infinitesimal negative time step and $\bar{w}$ is a Wiener process running backward from $T$ to $0$. This equation is what connects score estimation to generative modeling: once a neural network has learned to approximate $\nabla_x \log p_t(x) \approx s_\theta(x,t)$, one can generate new samples by drawing $x(T) \sim p_T \approx \mathcal{N}(0, I)$ and integrating the reverse SDE numerically from $t=T$ down to $t=0$ using a standard SDE solver such as Euler--Maruyama. Each solver step queries the score network to estimate the local direction of increasing data probability, gradually transforming unstructured noise into a sample from the learned data distribution. A useful practical consequence is that the diffusion coefficient $g(t)$ depends only on time and not on the state~$x$, making the noise \emph{additive}. In this regime the It\^o and Stratonovich interpretations of the SDE coincide, and solvers optimized for additive noise --- such as the ShARK and SEA methods --- can be used. These solvers tend to achieve higher accuracy when compared to other general It\^o or Stratonovich solvers, such as the Euler-Maruyama solver, as noted in~\cite{Arruda:2025dsbi}. GenSBI implements these solvers through the \texttt{diffrax}  library~\cite{Kidger:2022diffrax}.

\begin{figure}[t]
  \centering
  \includegraphics[width=\textwidth]{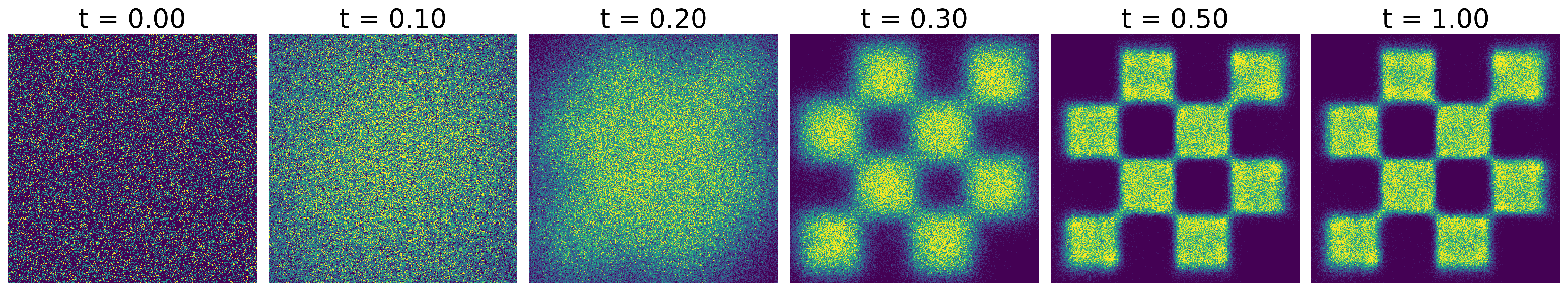}
  \caption{Unconditional density estimation with score matching (VE SDE)~\cite{Song:2020sde} on a two-dimensional checkerboard distribution. Each panel shows the sample density at a different stage of the reverse-time SDE, with time running from the noise prior ($t=0$) to the generated data distribution ($t=1$). The prior is a zero-mean Gaussian with $\sigma=15$, distributing samples broadly across the domain; spatial structure emerges gradually as the reverse diffusion denoises the samples.}
  \label{fig:SM_samples}
\end{figure}

The formulation presented here is unconditional: the score network depends only on the noisy state $x$ and the time $t$. Conditioning on additional variables, such as observed data, requires only minor modifications and is described in Section~\ref{sec:conditional_nde}. The SDE-based framework is implemented in GenSBI's \texttt{ScoreMatchingMethod}, which supports both VP and VE SDE types. Sampling uses the reverse SDE solver (\texttt{SMSDESolver}) by default; a probability flow ODE solver (\texttt{SMODESolver}) is also available for deterministic sampling and likelihood computation. Figure~\ref{fig:SM_samples} illustrates the sampling process on a two-dimensional checkerboard distribution: starting from a zero-mean Gaussian prior with $\sigma=15$ at $t=0$, the reverse-time SDE gradually denoises the samples until the checkerboard structure is recovered at $t=1$.

\subsection{Denoising Diffusion and EDM}
\label{sec:edm}

The stochastic nature of the reverse-time SDE presented in the previous section is not the only route to sample generation. Song et al.~\cite{Song:2020sde} demonstrated that for every diffusion process governed by an SDE of the form $\mathrm{d}x = f(x,t)\,\mathrm{d}t + g(t)\,\mathrm{d}w$, there exists a deterministic ordinary differential equation---the \emph{probability flow ODE}---whose trajectories share the same marginal distributions $p_t(x)$ at every time $t$ (eq.~13 of~\cite{Song:2020sde}):
\begin{equation}
  \mathrm{d}x = \left[f(x,t) - \frac{1}{2}\,g(t)^2\,\nabla_x \log p_t(x)\right]\mathrm{d}t\,.
\end{equation}
This result follows from the Fokker--Planck equation associated with the SDE. The second-order diffusion terms can be absorbed into a modified drift that depends on the score $\nabla_x \log p_t(x)$, yielding a deterministic process whose probability density evolves identically to that of the original stochastic process~\cite{Song:2020sde}. The score function is the same one estimated via denoising score matching (Section~\ref{sec:score_models}), so a single trained network supports both SDE-based and ODE-based sampling.

Replacing the stochastic solver with a deterministic ODE solver has several practical consequences. Given a fixed initial noise vector $x(T)$ drawn from the prior, the trajectory to the generated sample $x(0)$ is fully determined, which makes the generation process reproducible. Adaptive-step ODE solvers such as Runge--Kutta~45 can further reduce the number of required network evaluations compared to fixed-step SDE discretisations~\cite{Song:2020sde}. The ODE formulation also enables exact likelihood computation through the continuous change-of-variables formula (eq.~39 of~\cite{Song:2020sde}),
\begin{equation}
\log p_0(x(0)) = \log p_T(x(T)) + \int_0^T \nabla \cdot \tilde{f}_\theta(x(t),t)\,\mathrm{d}t\,,
\end{equation}
where $\tilde{f}_\theta$ denotes the ODE drift with the learned score plugged in and $\nabla \cdot$ is the divergence operator. Since computing the exact divergence scales poorly with dimensionality, the Skilling--Hutchinson trace estimator is typically used to obtain an unbiased approximation~\cite{Song:2020sde}. Exact likelihood computation is relevant for neural likelihood estimation (Section~\ref{sec:sbi}), although evaluating the integral requires solving the full ODE and remains computationally expensive in practice.

While the SDE framework of Song et al.~\cite{Song:2020sde} provided a unified theoretical picture, the relationship between design choices---noise schedules, network parameterisation, loss weighting, and solver discretisation---and their effect on sample quality was not systematically explored. Karras et al.~\cite{Karras:2022edm} addressed this gap by reformulating diffusion models through a single, generalised probability flow ODE parameterised by a noise schedule $\sigma(t)$ and a scale schedule $s(t)$ (eq.~4 of~\cite{Karras:2022edm}):
\begin{equation}
\mathrm{d}x = \left[\frac{\dot{s}(t)}{s(t)}\,x - s(t)^2\,\frac{\dot{\sigma}(t)}{\sigma(t)}\,\nabla_x \log p\!\left(\frac{x}{s(t)};\,\sigma(t)\right)\right]\mathrm{d}t\,.
\end{equation}
In this formulation, the marginal distribution at each time $t$ is a scaled and noise-corrupted version of the data distribution, and different choices of $s(t)$ and $\sigma(t)$ recover all previously proposed noise schedules as special cases. Specifically, the variance-preserving (VP) SDE~\cite{Song:2020sde} corresponds to continuous signal scaling with $s(t) = 1/\sqrt{\exp(\tfrac{1}{2}\beta_d\,t^2 + \beta_{\min}\,t)}$ and $\sigma(t) = \sqrt{\exp(\tfrac{1}{2}\beta_d\,t^2 + \beta_{\min}\,t) - 1}$, while the variance-exploding (VE) SDE corresponds to $s(t) = 1$ and $\sigma(t) = \sqrt{t}$~\cite{Karras:2022edm}. This decomposition shows that VP and VE are not distinct models but rather different schedule choices within the same framework; a comprehensive summary of the modulating coefficients for all schedule variants is given in Table~1 of Karras et al.~\cite{Karras:2022edm}.

The central practical contribution of EDM is a principled preconditioning scheme for the neural network. Rather than training a raw network $F_\theta$ to predict the noise or the score directly---both of which suffer from scale-dependent error amplification---EDM wraps $F_\theta$ in $\sigma$-dependent scaling functions that normalise its inputs and outputs (eq.~7 of~\cite{Karras:2022edm}):
\begin{equation}
D_\theta(x;\,\sigma) = c_{\mathrm{skip}}(\sigma)\,x + c_{\mathrm{out}}(\sigma)\,F_\theta\!\left(c_{\mathrm{in}}(\sigma)\,x;\; c_{\mathrm{noise}}(\sigma)\right).
\end{equation}
Each modulating function is derived from the requirement that both the effective input and target of $F_\theta$ have unit variance across all noise levels. The input scaling $c_{\mathrm{in}}(\sigma) = 1/\sqrt{\sigma^2 + \sigma_{\mathrm{data}}^2}$ normalises the noisy input, the output scaling $c_{\mathrm{out}}(\sigma) = \sigma\,\sigma_{\mathrm{data}}/\sqrt{\sigma^2 + \sigma_{\mathrm{data}}^2}$ normalises the effective training target, and the skip connection weight $c_{\mathrm{skip}}(\sigma) = \sigma_{\mathrm{data}}^2/(\sigma^2 + \sigma_{\mathrm{data}}^2)$ is chosen to minimise $c_{\mathrm{out}}(\sigma)$ and thus minimise the amplification of approximation errors made by $F_\theta$~\cite{Karras:2022edm}. The noise conditioning input $c_{\mathrm{noise}}(\sigma) = \tfrac{1}{4}\ln\sigma$ is chosen empirically. Together, these scalings ensure that the network operates in a well-conditioned regime regardless of the noise level, which leads to more stable training and higher sample quality.

The training objective minimises the expected denoising error over noise levels sampled from a log-normal distribution (eq.~8 of~\cite{Karras:2022edm}):
\begin{equation}
\mathcal{L}(\theta) = \mathbb{E}_{\sigma,\,y,\,n}\left[\lambda(\sigma)\,\|D_\theta(y + n;\,\sigma) - y\|^2\right],
\end{equation}
where $y$ is a clean data sample, $n \sim \mathcal{N}(0, \sigma^2 I)$ is the added noise, and $\lambda(\sigma) = (\sigma^2 + \sigma_{\mathrm{data}}^2)/(\sigma\,\sigma_{\mathrm{data}})^2$ is the EDM loss weighting~\cite{Karras:2022edm}. This weighting follows from the preconditioning design rather than from a variational bound. Because the denoiser output is scaled by $c_{\mathrm{out}}(\sigma)$, the effective per-sample loss seen by the raw network $F_\theta$ is $\lambda(\sigma)\,c_{\mathrm{out}}(\sigma)^2$; Karras et al.~\cite{Karras:2022edm} set $\lambda(\sigma) = 1/c_{\mathrm{out}}(\sigma)^2$ so that this effective weight equals unity at every noise level, ensuring that no range of $\sigma$ dominates the gradient signal.

The resulting weighting is a principled consequence of the unit-variance preconditioning, but it does not correspond to the likelihood weighting $\lambda(t) = g(t)^2$ that provides a variational bound on the negative log-likelihood (Section~\ref{sec:score_models}); instead, it prioritises well-conditioned optimisation across all noise scales. The training distribution $p_{\mathrm{train}}(\sigma)$ is a log-normal that concentrates sampling on intermediate noise levels where the loss can be reduced most effectively, rather than wasting capacity on extremely small or extremely large noise scales.

\begin{figure}[t]
  \centering
  \includegraphics[width=\textwidth]{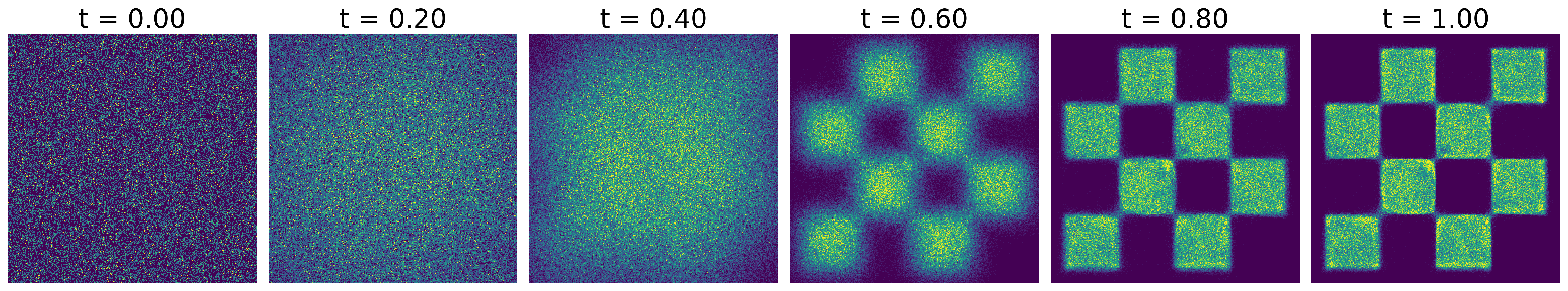}
  \caption{Same as Figure~\ref{fig:SM_samples}, but using the EDM stochastic sampler~\cite{Karras:2022edm}, which augments a second-order Heun ODE integrator with Langevin-like noise injection (``churn''), and a zero-mean Gaussian prior with $\sigma=80$. Recognisable checkerboard structure appears only in the final stages of the sampling process ($t \gtrsim 0.8$), consistent with the EDM noise schedule concentrating most of the signal recovery at low noise levels.}
  \label{fig:EDM_samples}
\end{figure}

For sampling, EDM combines a second-order Heun ODE integrator with optional stochastic noise injection. At each step, the solver evaluates the denoiser $D_\theta$ at the current noise level, computes an Euler estimate, and then applies a corrector step at the target noise level for improved accuracy (Algorithm 1 of~\cite{Karras:2022edm}). This second-order correction yields a local truncation error of $O(h^3)$ at the cost of one additional network evaluation per step. 
Beyond purely deterministic sampling, EDM introduces a stochastic variant controlled by a set of hyperparameters: $S_{\mathrm{churn}}$ governs the overall amount of noise reinjection per step, $S_{\mathrm{min}}$ and $S_{\mathrm{max}}$ define the noise-level range within which stochasticity is applied, and $S_{\mathrm{noise}}$ modulates the standard deviation of the injected noise (Algorithm 2 of~\cite{Karras:2022edm}). This controlled reinjection of stochasticity during sampling can correct errors accumulated during earlier steps, particularly at intermediate noise levels, and in practice often improves sample quality beyond what the purely deterministic solver achieves. The mechanism is conceptually related to the predictor--corrector (PC) sampling framework of Song et al.~\cite{Song:2020sde}, in which each reverse-SDE step (the predictor) is followed by one or more corrector steps of score-based MCMC --- typically annealed Langevin dynamics --- that adjust the sample to better match the target marginal $p_t(x)$ at the current noise level. EDM's stochastic churn can be viewed as a streamlined variant of this idea, where the noise injection and deterministic denoising play the roles of corrector and predictor, respectively, but are integrated into a single, tunable sampling loop rather than requiring a separate MCMC subroutine.

The EDM framework brings together network preconditioning, loss weighting, noise schedule design, and sampler construction into a single coherent recipe. This is the mathematical foundation of GenSBI's \texttt{DiffusionEDMMethod}, which supports the standard EDM schedule as well as the VP and VE schedules as special cases. GenSBI's \texttt{EDMSolver} implements the second-order Heun sampler with configurable churn parameters ($S_{\mathrm{churn}}$, $S_{\mathrm{min}}$, $S_{\mathrm{max}}$, $S_{\mathrm{noise}}$), and the solver can be swapped post-training without retraining the underlying model. Figure~\ref{fig:EDM_samples} shows the same checkerboard distribution generated via the EDM stochastic sampler~\cite{Karras:2022edm}, which augments the second-order Heun integrator with Langevin-like noise injection (churn); most of the visible structural change is concentrated in the second half of the sampling process.

\subsection{Flow Matching}
\label{sec:flow_matching}

While score-based diffusion models learn to reverse a noising process, flow matching takes an alternative route: it directly learns a velocity field $v_t(x)$ that transports samples from a simple prior distribution to the data distribution along a prescribed probability path \cite{Lipman:2023fm}. In principle, any distribution whose density is known and easy to sample from can serve as the source prior; in practice, a multivariate Gaussian with zero mean and unit covariance matrix $p_0 = \mathcal{N}(0, I)$ is the near-universal choice, since it is isotropic, easy to sample in any dimension, and pairs naturally with the affine probability paths described below. Following the flow matching convention, this section denotes the source noise as $x_0 \sim \mathcal{N}(0, I)$ and the target data as $x_1 \sim p_\mathrm{data}$, reversing the notation of sections~\ref{sec:score_models} and~\ref{sec:edm} where $x(0)$ was the data and $x(T)$ the noise. Rather than constructing a forward diffusion and then denoising, the framework specifies a time-dependent flow --- a diffeomorphism $\phi_t$ defined by the ordinary differential equation
\begin{equation}
\frac{d}{dt}\phi_t(x) = v_t(\phi_t(x)),\qquad \phi_0(x)=x,
\end{equation}
and trains a neural network to approximate $v_t$ so that the pushforward of the prior $p_0$ through $\phi_t$ matches the data distribution $p_1$ at $t=1$ (eq.~1--2 of~\cite{Lipman:2023fm}). The density $p_t$ along the flow satisfies the continuity equation
\begin{equation}
\frac{\partial}{\partial t}p_t(x) + \nabla\cdot\bigl(p_t(x)\,v_t(x)\bigr) = 0,
\end{equation}
which guarantees that probability is conserved along the trajectories (eq.~26 of~\cite{Lipman:2023fm}). At sampling time, one draws $x_0\sim p_0$ (a standard Gaussian) and integrates the learned velocity field forward from $t=0$ to $t=1$ using an ODE solver.

\paragraph{Conditional flow matching.}
The flow matching loss trains the network $v_\theta$ to approximate a target velocity field $u_t(x)$ that generates the desired probability path:
\begin{equation}
\mathcal{L}_\mathrm{FM}(\theta) = \mathbb{E}_{t\sim\mathcal{U}[0,1],\; x\sim p_t}\bigl\|v_\theta(x,t) - u_t(x)\bigr\|^2.
\end{equation}
Computing $u_t(x)$ directly is intractable, because it requires integrating over the entire data distribution. The key insight of Lipman et al.\ \cite{Lipman:2023fm} is that one can instead construct \emph{conditional} vector fields $u_t(x\mid x_1)$, each defined for a single data example $x_1$, and regress against those. The conditional flow matching (CFM) loss
\begin{equation}
\mathcal{L}_\mathrm{CFM}(\theta) = \mathbb{E}_{t,\; q(x_1),\; p_t(x\mid x_1)}\bigl\|v_\theta(x,t) - u_t(x\mid x_1)\bigr\|^2
\end{equation}
has gradients identical to those of the intractable $\mathcal{L}_\mathrm{FM}$ (Theorem~2 of~\cite{Lipman:2023fm}), so minimizing the tractable per-example objective recovers the correct marginal velocity field. This equivalence holds because the squared-error loss is a Bregman divergence, whose gradient with respect to its second argument is affine invariant --- that is, $\nabla_v D(\mathbb{E}[Y],v) = \mathbb{E}[\nabla_v D(Y,v)]$ --- which allows the conditional expectation defining the marginal velocity to be exchanged with the gradient of the loss \cite{Lipman:2023fm, Lipman:2024fmguide}. %

\paragraph{Affine probability paths and optimal transport.}
In practice, one specifies the conditional probability path through an affine transformation
\begin{equation}
\psi_t(x_0 \mid x_1) = \sigma_t\, x_0 + \alpha_t\, x_1,
\end{equation}
where $x_0\sim\mathcal{N}(0,I)$ is the source noise (\S4.8 of~\cite{Lipman:2024fmguide}). The time-dependent coefficients $\alpha_t$ and $\sigma_t$ --- together with their derivatives $\dot{\alpha}_t$ and $\dot{\sigma}_t$ --- are provided by a \emph{scheduler} and fully determine the conditional velocity field (Theorem~3 of~\cite{Lipman:2023fm}):
\begin{equation}
u_t(x\mid x_1) = \frac{\dot{\sigma}_t}{\sigma_t}\bigl(x - \alpha_t\, x_1\bigr) + \dot{\alpha}_t\, x_1.
\end{equation}

The conditional optimal transport (CondOT) scheduler \cite{Lipman:2023fm} sets $\alpha_t = t$ and $\sigma_t = 1-t$ (eq.~4.48 of~\cite{Lipman:2024fmguide}; eq.~21--22 of~\cite{Lipman:2023fm}), yielding the linear interpolant
\begin{equation}
\psi_t(x_0 \mid x_1) = (1-t)\,x_0 + t\,x_1
\end{equation}
with constant conditional velocity $u_t(x_0 \mid x_1) = x_1 - x_0$. This CondOT path corresponds to the exact solution of the dynamic optimal transport problem with quadratic cost, conditioned on a single target point. As a result, sample trajectories follow straight lines at constant speed, making the ODE easy to integrate numerically --- in the ideal case, a single Euler step suffices. In practice, because the marginal velocity field averages over many such conditional paths, $10$--$50$ solver steps are typical for accurate posterior recovery. This geometric simplicity stands in contrast to the curved trajectories of diffusion-based probability paths, where the score function varies rapidly and accurate integration requires many solver steps. This contrast is visible in Figure~\ref{fig:samples_trajectories}, which compares the trajectories followed by individual samples during generation on a two-dimensional checkerboard target: score matching with the VE SDE produces highly stochastic, tangled paths; the EDM stochastic sampler~\cite{Karras:2022edm} yields substantially straighter but still visibly curved trajectories, particularly toward the end of the integration; and flow matching with the CondOT path produces nearly rectilinear paths consistent with the constant conditional velocity $u_t(x \mid x_1) = x_1 - x_0$. 

It is important to distinguish the CondOT path (which defines a conditional optimal-transport displacement for each individual noise--data pair) from strategies that attempt to straighten trajectories at the \emph{marginal} level. Minibatch optimal transport~\cite{Tong:2023otcfm, Lipman:2024fmguide} optimises the noise-to-data assignment across an entire training batch and provably straightens marginal trajectories in the unconditional setting; however, its extension to conditional generation remains an open problem, because the OT coupling introduces a skewed conditional prior that creates a train--test mismatch. A promising remedy is the condition-aware OT coupling (C$^2$OT) of Cheng and Schwing~\cite{Cheng:2025c2ot}, which adds a condition-dependent penalty to the transport cost; this approach has not yet been validated in the SBI setting, and GenSBI therefore adopts the conservative independent coupling for the time being. Another line of work is rectified flow~\cite{Liu:2022yjp}, a distillation strategy that iteratively retrains the velocity network on its own generated trajectories to progressively straighten the flow paths, thereby reducing the number of model evaluations required at sampling time. GenSBI does not currently implement rectification, as its primary focus is posterior sample quality rather than sampling speed, and the CondOT path already provides straight conditional trajectories that are sufficient for accurate posterior recovery in the SBI setting.

With the CondOT path, the CFM loss reduces to the simple regression problem (eq.~23 of~\cite{Lipman:2023fm})
\begin{equation}
\mathcal{L}_\mathrm{CFM}(\theta) = \mathbb{E}_{t,\; q(x_1),\; p(x_0)}\bigl\|v_\theta\bigl((1-t)\,x_0 + t\,x_1,\; t\bigr) - (x_1 - x_0)\bigr\|^2,
\end{equation}
where the target is the displacement vector from noise to data, independent of $t$. Training amounts to sampling pairs $(x_0, x_1)$, interpolating to time $t$, and regressing the network output onto the constant velocity $x_1 - x_0$.

\begin{figure}[t]
  \centering
  \includegraphics[width=\textwidth]{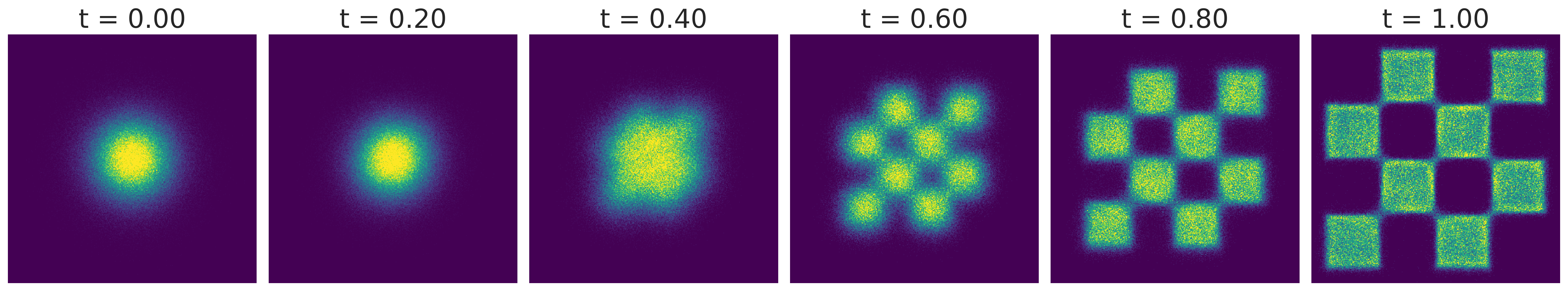}
  \caption{Same as Figures~\ref{fig:SM_samples} and~\ref{fig:EDM_samples}, but using flow matching with the conditional optimal-transport (CondOT) path~\cite{Lipman:2023fm}. The standard Gaussian prior at $t=0$ is concentrated near the origin, and the checkerboard modes emerge progressively throughout the integration, becoming clearly separated by $t=0.6$. The smoother, more gradual transition reflects the straighter probability paths prescribed by the CondOT scheduler.}
  \label{fig:FM_samples}
\end{figure}

\paragraph{Connection to the probability flow ODE.}
Flow matching and score-based diffusion are closely related. When the affine probability path in CFM is constructed to match a diffusion-style noise schedule --- for example, the VP or VE schedules of Song et al.\ \cite{Song:2020sde} --- the resulting conditional velocity field coincides exactly with the drift of the probability flow ODE described in Section~\ref{sec:edm} \cite{Lipman:2023fm}. In this sense, the CFM framework subsumes the deterministic sampling pathway of score-based models. The practical advantage is that CFM provides a direct, simulation-free regression objective for the velocity field, bypassing the need to derive and weight a denoising score matching loss through the forward SDE. Even when one restricts the probability paths to standard diffusion schedules, the flow matching objective has been found to yield more stable training than the corresponding score matching formulation \cite{Lipman:2023fm}.

The CondOT path goes further by choosing a schedule that is \emph{not} tied to any diffusion process. The resulting straight-line trajectories simplify both the loss landscape and the sampling procedure relative to diffusion-based alternatives, which is the primary motivation for adopting flow matching as the default generative method in GenSBI. The progressive refinement of the sample distribution under the CondOT scheduler is shown in Figure~\ref{fig:FM_samples} for the same checkerboard target used in Figures~\ref{fig:SM_samples} and~\ref{fig:EDM_samples}: the standard Gaussian prior, visibly concentrated near the origin at $t=0$, transitions smoothly into the multi-modal target, with well-separated modes already apparent by $t=0.6$.

\paragraph{Mass coverage and stochastic sampling.}
In practice, flow matching tends to produce mass-covering posteriors that distribute probability across the full support of the target distribution \cite{Dax:2024fmpe}. The mechanism behind this behaviour is the conditional expectation structure of the CFM objective: the optimal minimiser of the squared-error loss is the marginal velocity field $u_t(x) = \mathbb{E}[u_t(X_t \mid X_1) \mid X_t = x]$, which averages over all conditional trajectories passing through a given state \cite{Lipman:2024fmguide}. Because this expectation integrates contributions from every mode in the data distribution, the network is trained to route probability mass toward the full support rather than collapsing to a single mode. However, this is an inductive bias of the training objective, not a formal guarantee: the flow matching regression loss does not minimise or bound the forward Kullback--Leibler divergence for deterministic ODEs \cite{Lipman:2024fmguide}, and with finite network capacity or imperfect optimisation, mode-dropping can occur in principle. Under stronger regularity conditions --- specifically, when both the target and learned velocity fields possess bounded second derivatives and the base distribution is sufficiently smooth --- the MSE loss does provide an upper bound on the forward KL divergence \cite{Dax:2024fmpe}, but these smoothness conditions are not guaranteed by the training procedure and are difficult to verify in practice. The comparison with diffusion models, which offer formal mass-coverage guarantees through the variational lower bound, is discussed in Section~\ref{sec:density_estimator_choice}.

A separate practical concern is that deterministic ODE solvers may introduce a discretisation bias that consistently underestimates the variance of the target distribution, making the generated probability mass too narrow. Singh and Fischer \cite{Singh:2024ssfm} showed that one can construct a family of SDEs that share the same marginal distributions as the flow matching ODE, with the score function derived analytically from the learned velocity field rather than requiring a separately trained model. The injected stochasticity mitigates the variance-underestimation bias of deterministic integration, at the cost of a tunable bias--variance trade-off controlled by a diffusion strength parameter $\alpha$. GenSBI implements two such SDE solvers for flow matching --- \texttt{ZeroEndsSolver} and \texttt{NonSingularSolver} --- alongside the default ODE solver, giving users the option to trade sampling speed for improved mass coverage without retraining the velocity network. This formulation underlies GenSBI's \texttt{FlowMatchingMethod}, which pairs the affine path with the CondOT scheduler and trains the velocity network via the CFM objective.

\begin{figure}[t]
  \centering
  \includegraphics[width=0.3\textwidth]{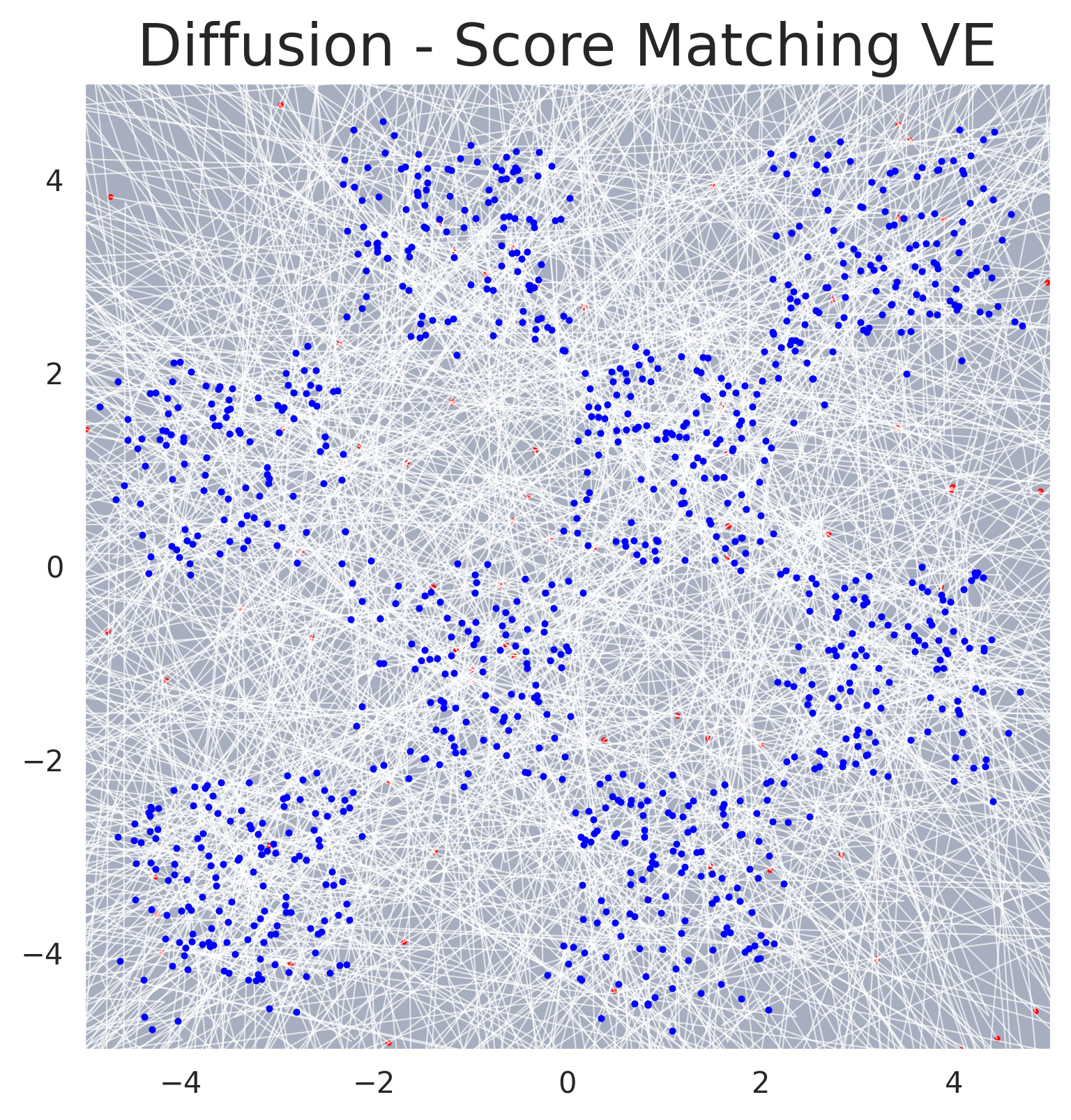}
  \includegraphics[width=0.3\textwidth]{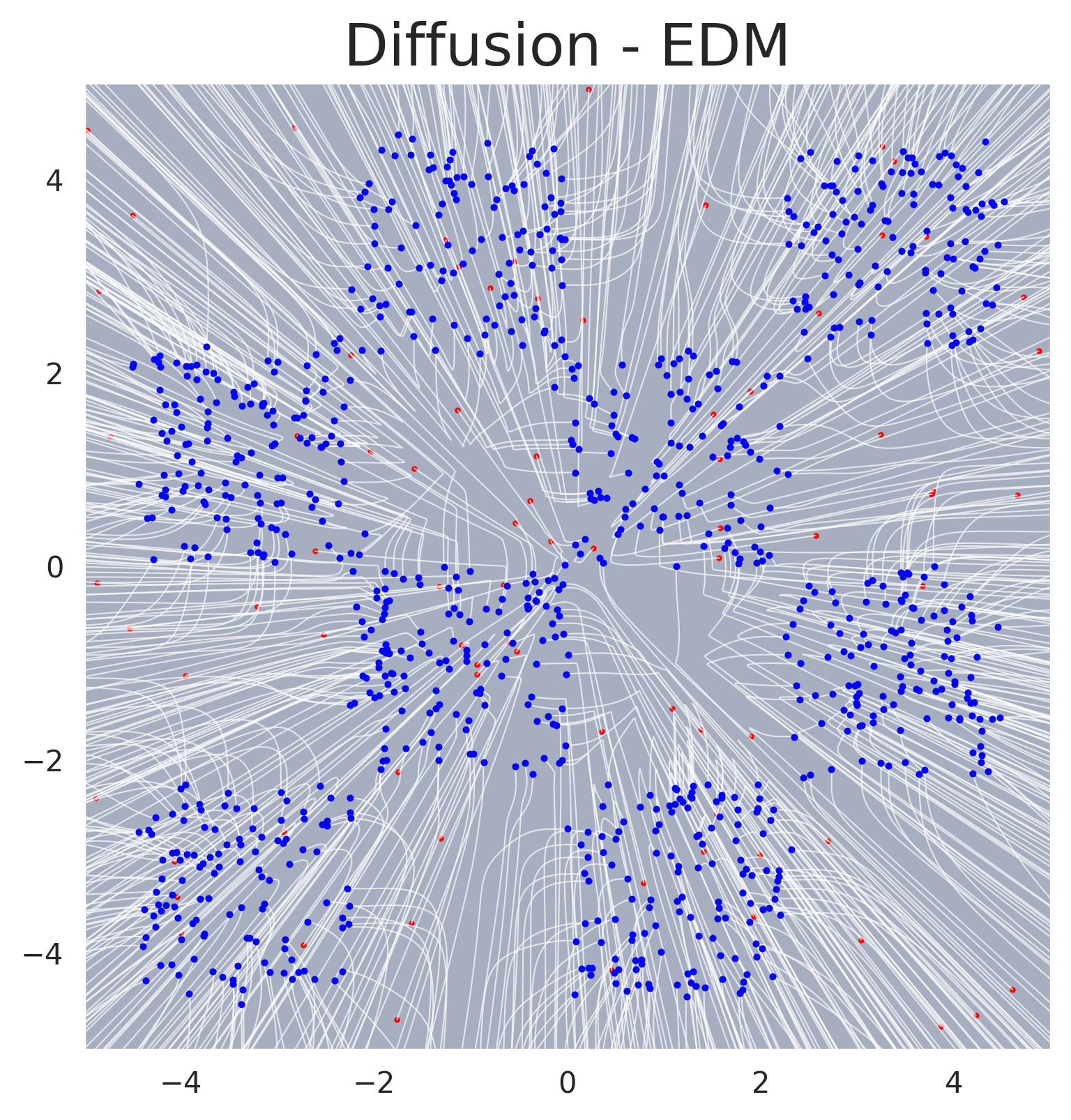}
  \includegraphics[width=0.3\textwidth]{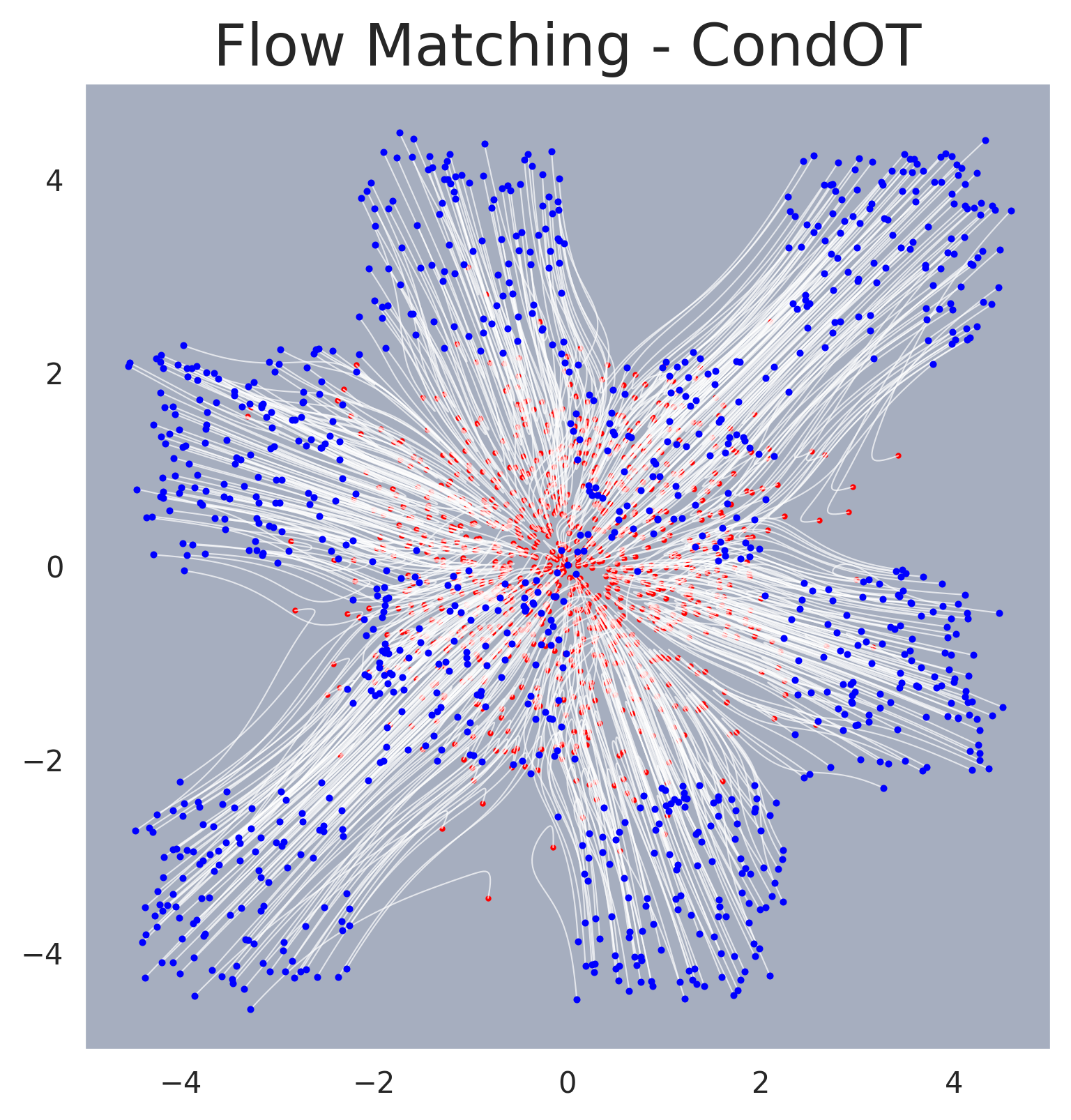}
  \caption{Comparison of individual sample trajectories during unconditional density estimation on the same two-dimensional checkerboard distribution used in Figures~\ref{fig:SM_samples}--\ref{fig:FM_samples}. Red markers show starting positions (prior samples) and blue markers show final positions (generated samples); grey lines trace the paths followed by individual samples. The prior is a zero-mean Gaussian with $\sigma=15$ for score matching, $\sigma=80$ for EDM, and $\sigma=1$ for flow matching. \textbf{Left:} Score matching with the VE SDE produces highly stochastic, tangled trajectories due to Brownian noise injected at every integration step. \textbf{Centre:} The EDM stochastic sampler~\cite{Karras:2022edm}, which augments a second-order Heun integrator with Langevin-like noise injection, yields substantially straighter paths, but noticeable curvature remains, particularly in the later stages of the integration. \textbf{Right:} Flow matching with the CondOT path produces nearly rectilinear trajectories, consistent with the straight-line conditional velocity field $u_t(x \mid x_1) = x_1 - x_0$.}
  \label{fig:samples_trajectories}
\end{figure}

\subsection{Conditional and Joint Density Estimation}
\label{sec:conditional_nde}

The three generative frameworks introduced in the preceding sections --- score matching, denoising diffusion with EDM preconditioning, and flow matching --- were presented in their unconditional form: given a dataset of samples $\{z^{(i)}\}$, the network learns to generate new samples from the same distribution. Simulation-based inference, however, requires \emph{conditional} generation: given an observation $x_\mathrm{obs}$, the goal is to draw posterior samples $\theta \sim p(\theta \mid x_\mathrm{obs})$.

Extending the unconditional formulations to conditional generation is simple in all three frameworks. The conditioning variable $x$ is supplied as an additional input to the neural network, so that the learned velocity field, score function, or denoiser becomes a function of both the noisy state and the condition. Concretely, the unconditional score $s_\phi(\theta_t, t) \approx \nabla_{\theta_t} \log p_t(\theta_t)$ is replaced by a conditional score $s_\phi(\theta_t, t, x) \approx \nabla_{\theta_t} \log p_t(\theta_t \mid x)$, and likewise for the velocity field in flow matching or the denoiser in EDM. The training objectives are unchanged except that each training step now draws a pair $(\theta, x)$ from the joint training set and feeds $x$ as an extra input to the network. For conditional flow matching, the loss becomes
\begin{equation}
\mathcal{L}_\mathrm{CFM}(\phi) = \mathbb{E}_{t,\, q(\theta_1, x),\, p_t(\theta \mid \theta_1)} \| v_\phi(\theta_t, t, x) - u_t(\theta \mid \theta_1) \|^2,
\end{equation}
where the only change relative to the unconditional case is the presence of $x$ as a network input. An analogous modification applies to the score matching loss and the EDM training objective.

This conditional formulation is the direct mathematical basis for neural posterior estimation (NPE). The network is trained on simulated pairs $\{(\theta^{(i)}, x^{(i)})\}$ drawn from the joint distribution $p(\theta, x) = p(x \mid \theta)\, p(\theta)$, where $p(\theta)$ is the prior and $p(x \mid \theta)$ is the simulator. After training, the model generates samples $\theta \sim q_\phi(\theta \mid x_\mathrm{obs})$ for any new observation $x_\mathrm{obs}$ by running the reverse SDE, probability flow ODE, or flow ODE with $x_\mathrm{obs}$ held fixed. The inference cost is independent of the simulator: the expensive forward simulations are performed once during dataset generation, and the trained model can be applied to as many observations as needed without additional simulations. This is the amortization property discussed in Section~\ref{sec:nde_sbi}.

How the conditioning information enters the network depends on the architecture. Three mechanisms are commonly used in transformer-based models \cite{Peebles:2023dit, Rombach:2022ldm}. In \emph{cross-attention}, the embedding of $x$ is projected into a separate key--value sequence, and the model's intermediate representations attend to it through standard multi-head attention. This allows the network to selectively extract relevant features from the condition at each layer. In \emph{concatenation}, the embedded condition is appended to the noisy state as additional input tokens and processed jointly through self-attention. This is the simplest approach and works well when the condition and target have comparable dimensionality. In \emph{adaptive layer normalization} (adaLN), the conditioning vector modulates the layer normalization parameters: the scale and shift applied after normalization are regressed from the condition rather than learned as fixed parameters \cite{Peebles:2023dit}. The adaLN-Zero variant additionally regresses a per-block scaling factor initialized to zero, so that each transformer block starts as the identity function and learns to incorporate the condition gradually during training. In GenSBI, Flux1 uses cross-attention to inject the condition through double-stream transformer blocks, while SimFormer and Flux1Joint rely on concatenation with node-level embeddings (Section~\ref{sec:nn_architectures}). Both Flux1 and Flux1Joint employ adaLN-Zero modulation for timestep conditioning.

Beyond conditional generation, the same frameworks support \emph{joint} density estimation. Instead of learning the conditional $p(\theta \mid x)$, a joint model learns the full distribution $p(\theta, x)$ and recovers any conditional at inference time by fixing the appropriate variables. In GenSBI, this is implemented through a binary \texttt{condition\_mask} $m \in \{0,1\}^{d_\theta + d_x}$, where $m_i = 1$ marks an observed (conditioned) coordinate and $m_i = 0$ an unobserved coordinate to be generated \cite{Gloeckler:2024simformer}. During training, the mask is randomized, exposing the model to all possible conditioning patterns. Concretely, the mask enters in two places. First, the network input is constructed by replacing the observed coordinates with their clean values:
\begin{equation}
  z_t^{m} = (1 - m) \odot z_t + m \odot z_0,
\end{equation}
where $z_t$ is the noisy state and $z_0$ is the clean data. Second, the learned dynamics are constrained to act only on unobserved dimensions. For flow matching, this amounts to zeroing the velocity field on observed coordinates:
\begin{equation}
  \frac{dz_t}{dt} = (1 - m) \odot v_\phi(z_t^{m},\, t,\, m),
\end{equation}
so that $z_t$ evolves only along the unobserved dimensions while the observed values remain fixed \cite{Gloeckler:2024simformer}. Correspondingly, the training loss computes the regression error only on unobserved coordinates: the residual between the predicted and target velocity is multiplied element-wise by $(1-m)$ before squaring, so no gradient signal arises from observed dimensions. The same masking strategy applies to score-based diffusion, where $(1-m)$ multiplies the score residual \cite{Gloeckler:2024simformer}. At inference time, the mask is set to fix $x = x_\mathrm{obs}$ and generate $\theta$, yielding posterior samples; alternatively, fixing $\theta$ and generating $x$ yields likelihood samples. From a conceptual and implementation standpoint, enabling this flexibility requires no architectural changes and no separate training procedure beyond randomising the mask --- the same training loop handles all masking patterns. The added generality may, however, require longer training, more expressive networks, or larger simulation budgets, since the model must learn to approximate many different conditionals and marginals rather than a single one. SimFormer and Flux1Joint support this joint mode in GenSBI. The masked conditional flow matching formulation described here was derived independently in GenSBI from the score-based formulation of Gloeckler et al.~\cite{Gloeckler:2024simformer}. 
Notably, a recent work (OneFlowSBI~\cite{Nautiyal:2026oneflowsbi}) converged independently at the same implementation, confirming that this is the natural extension of SimFormer's masking strategy to flow matching.

GenSBI also supports \emph{unconditional} density estimation, in which neither parameters nor observations play a distinguished role. In this mode, the model learns to sample from an arbitrary target distribution $p(z)$ without any conditioning input. This makes GenSBI usable as a general-purpose neural density estimation library, extending its applicability beyond the SBI context to any problem that requires learning and sampling from complex distributions.

\section{GenSBI: Software Design and Features}
\label{sec:software}

The previous two sections established the theoretical foundations of simulation-based inference and the generative models that serve as its computational backbone. We now turn to GenSBI itself: a JAX-native library that implements these ideas in a modular, extensible software framework. GenSBI provides three generative methods --- flow matching, denoising diffusion (EDM), and score matching --- paired with three transformer-based neural network architectures --- Flux1, SimFormer, and Flux1Joint --- under a unified interface that allows users to combine any generative method with a compatible architecture through lightweight configuration changes. Flux1 is designed exclusively for conditional density estimation, while SimFormer and Flux1Joint support conditional, joint, and unconditional modes through a configurable condition mask. The library targets scientists and machine learning practitioners who need to perform neural posterior estimation (or, more generally, neural density estimation) within the JAX ecosystem, and who value the ability to experiment with different generative formulations and network designs without rewriting their training and inference pipelines.

This section presents the library's design and features, mapped to the theory developed in Section~\ref{sec:generative_models}. Section~\ref{sec:architecture} describes the overall software architecture: the strategy pattern that decouples generative methods from inference pipelines, the pipeline hierarchy that handles conditional, joint, and unconditional workflows, and the model wrapper layer that adapts each neural network's native interface to the calling convention expected by the pipeline. Section~\ref{sec:nn_architectures} introduces the three transformer-based architectures. Section~\ref{sec:solvers} describes the solver system. Section~\ref{sec:recipes} presents the high-level recipes API with a minimal code example. Section~\ref{sec:diagnostics} covers the built-in posterior calibration diagnostics. Finally, Section~\ref{sec:jax_integration} discusses GenSBI's integration with the broader JAX ecosystem.

\subsection{Architecture Overview}
\label{sec:architecture}

\begin{figure}[t]
  \centering
  \includegraphics[width=\textwidth]{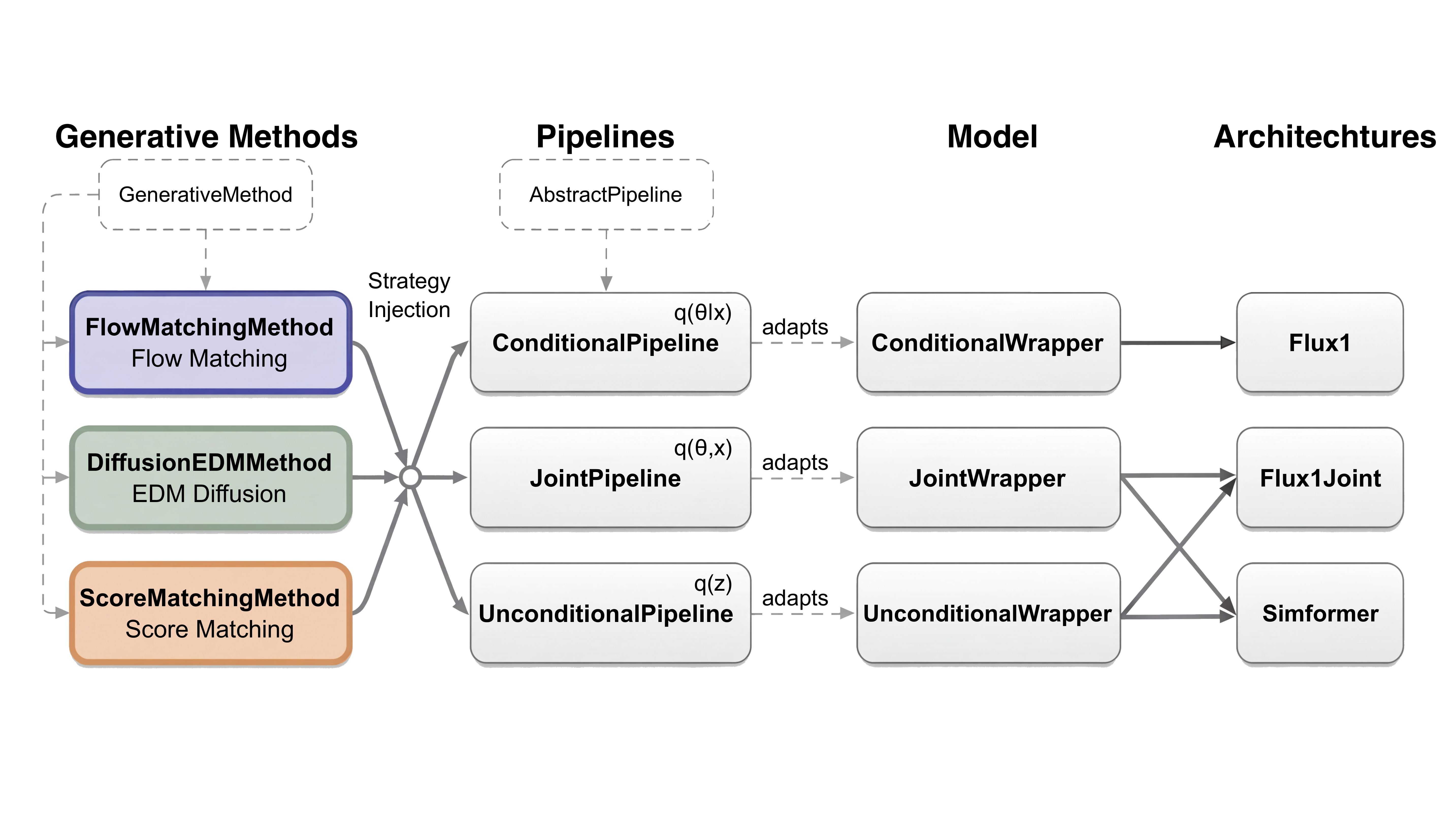}
  \caption{GenSBI's three-axis architecture. Generative methods (left) inject into pipelines (centre) via the strategy pattern; each pipeline delegates to a model wrapper (right) that adapts the neural network's interface. Any method can be combined with any pipeline; SimFormer and Flux1Joint work with all three pipeline types, while Flux1 is restricted to the conditional pipeline.}
  \label{fig:architecture_overview}
\end{figure}

GenSBI is organised around a central design principle: the mathematical formulation of the generative process (flow matching, EDM diffusion, or score matching) should be decoupled from both the neural network architecture and the inference mode (conditional, joint, or unconditional). This separation is achieved through a layered architecture comprising three components, illustrated in Figure~\ref{fig:architecture_overview}: a \emph{generative method} that encapsulates the mathematical framework, a \emph{pipeline} that manages the training and sampling workflow for a given inference mode, and a \emph{model wrapper} that adapts the neural network's calling convention to match the pipeline's requirements. Each component corresponds to a largely independent axis of variation, and the software composes them so that all valid combinations are available without code duplication.

\paragraph{Generative methods: the strategy pattern.}
At the core of the architecture is an abstract \texttt{GenerativeMethod} base class that defines the interface every generative formulation must satisfy. Three concrete implementations correspond to the three frameworks introduced in Section~\ref{sec:generative_models}:
\begin{itemize}
  \item \texttt{ScoreMatchingMethod} --- implements score-based generative modeling via VP and VE stochastic differential equations (Section~\ref{sec:score_models}).
  \item \texttt{DiffusionEDMMethod} --- implements the EDM denoising diffusion framework with configurable noise schedules (EDM, VP, VE) and the preconditioning scheme of Karras et al.~\cite{Karras:2022edm} (Section~\ref{sec:edm}).
  \item \texttt{FlowMatchingMethod} --- implements optimal-transport conditional flow matching with affine probability paths and the CondOT scheduler (Section~\ref{sec:flow_matching}).
\end{itemize}
Each method encapsulates five responsibilities: \emph{path construction} (defining the probability path or noise schedule), \emph{loss creation} (building the training objective), \emph{batch preparation} (sampling noise and time for each training step), \emph{solver construction} (instantiating the sampler for inference), and \emph{initial sample generation} (drawing from the prior distribution). By consolidating all method-specific logic within the \texttt{GenerativeMethod} object, the rest of the library --- the pipeline, the neural network, the solver --- remains agnostic to the choice of generative formulation. Switching from flow matching to EDM diffusion, for instance, requires changing a single constructor argument; the training loop, loss computation, sampling interface, and checkpoint management all remain unchanged.

\paragraph{Pipelines: inference mode abstraction.}
The second axis of variation, shown in the centre column of Figure~\ref{fig:architecture_overview}, is the inference mode. GenSBI provides three pipeline classes, each composing with any \texttt{GenerativeMethod} via strategy injection:
\begin{itemize}
  \item \texttt{ConditionalPipeline} --- for conditional density estimation. In the SBI context, this is the primary pipeline for neural posterior estimation: the model learns $q_\phi(\theta \mid x)$ and generates posterior samples given an observation $x_\mathrm{obs}$. This pipeline can also be used for neural likelihood estimation and to train emulators.
  \item \texttt{JointPipeline} --- for joint density estimation. The model learns $q_\phi(\theta, x)$ and can be conditioned on any subset of variables at inference time, enabling both posterior, likelihood and evidence queries from a single trained model.
  \item \texttt{UnconditionalPipeline} --- for unconditional density estimation. The model learns an arbitrary target distribution $q_\phi(z)$ with no conditioning input, making GenSBI usable as a general-purpose neural density estimation library beyond the SBI context.
\end{itemize}
All three pipelines inherit from a common \texttt{AbstractPipeline} base class that provides the training loop, optimizer construction (AdamW with warmup and cosine decay), exponential moving average (EMA) of model parameters, checkpoint management via Orbax, and early stopping based on a validation loss ratio. The pipeline receives the generative method at construction time and delegates all method-specific operations --- batch preparation, loss computation, solver instantiation, and initial noise sampling --- to the method object. This composition means that the nine possible combinations of three methods and three inference modes are all available without code duplication: each pipeline works with any method, and each method works with any pipeline.

\begin{figure}[t]
  \centering
  \includegraphics[width=\textwidth]{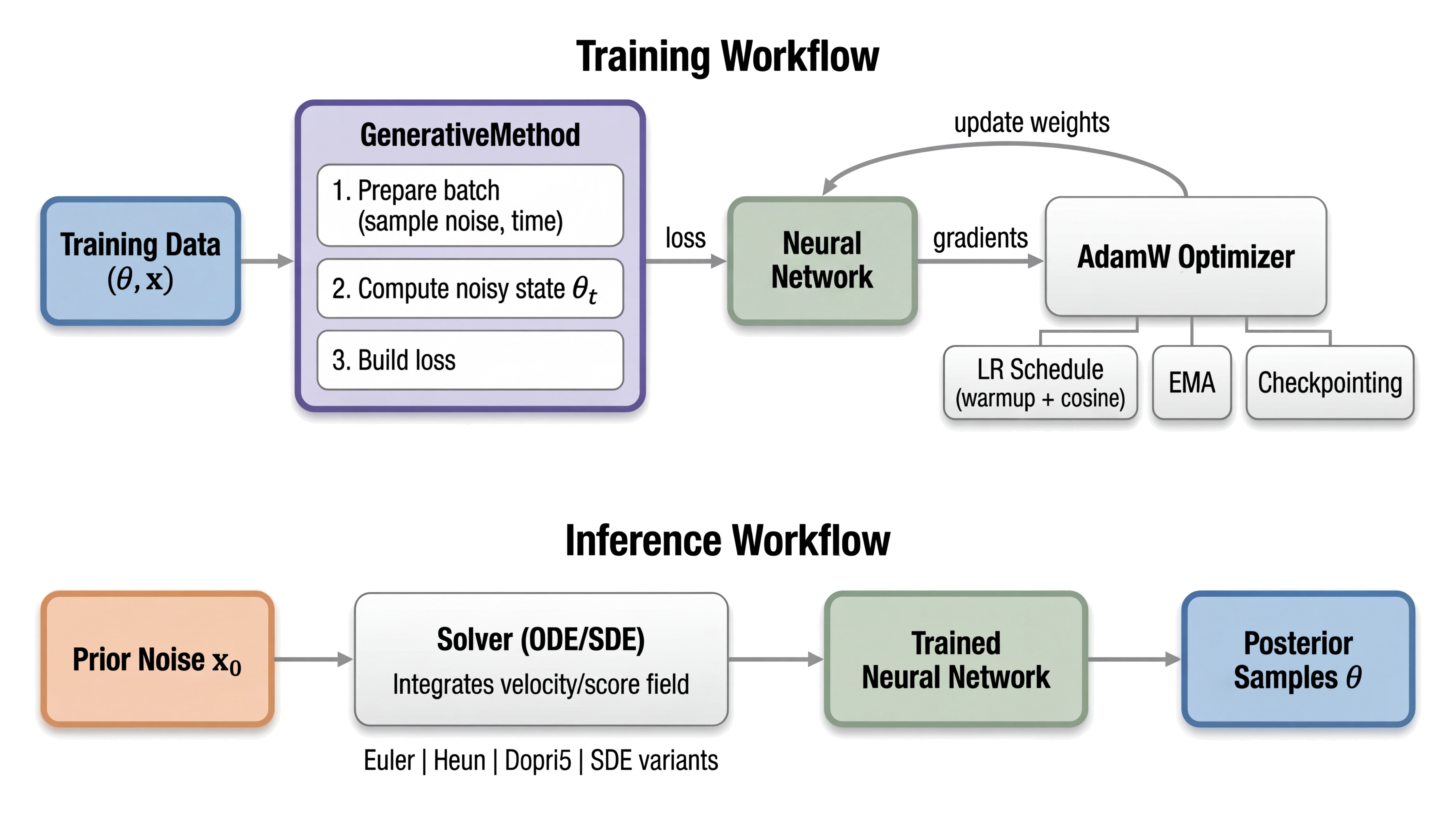}
  \caption{Training and inference workflows. During training (top), the generative method prepares noisy batches and builds the loss; the pipeline handles optimisation, EMA, and checkpointing. During inference (bottom), a solver integrates the learned velocity or score field from prior noise to posterior samples.}
  \label{fig:pipeline_composition}
\end{figure}

Figure~\ref{fig:pipeline_composition} details how these components interact at runtime. During training, the pipeline delegates batch preparation and loss construction to the generative method, then passes the resulting objective to the optimizer. At inference time, the pipeline constructs a solver from the method and integrates the learned field to produce samples.

\begin{figure}[t]
  \centering
  \includegraphics[width=0.85\textwidth]{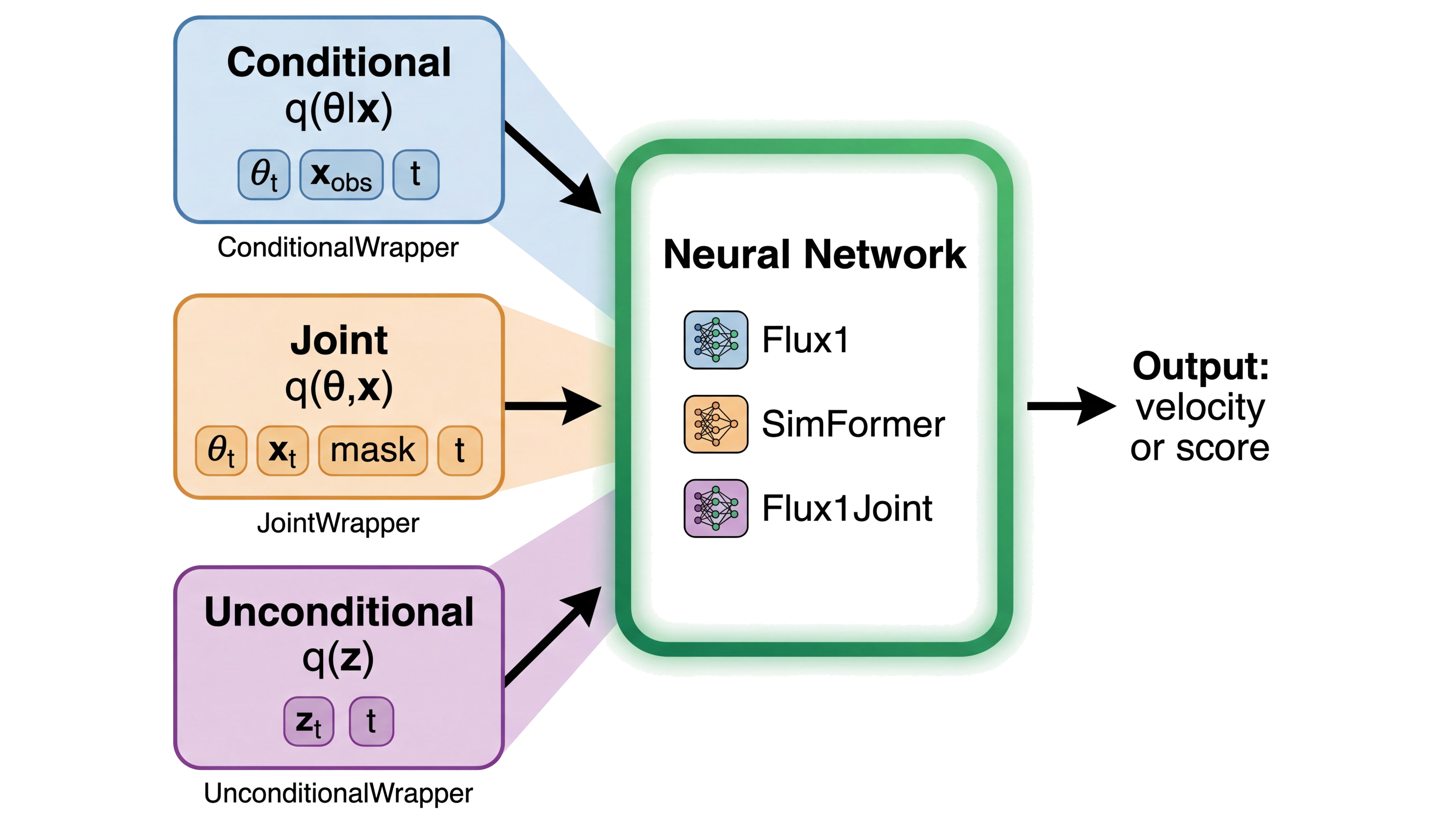}
  \caption{The model wrapper adapter pattern. Each wrapper translates the pipeline's unified calling convention into the model's native interface: \texttt{ConditionalWrapper} forwards separate observation and conditioning streams; \texttt{JointWrapper} concatenates them into a joint sequence with a condition mask; \texttt{UnconditionalWrapper} passes only the noisy state. The pipeline determines the inference mode; the wrapper determines how the model is called.}
  \label{fig:wrapper_pattern}
\end{figure}

\paragraph{Model wrappers: adapting the backbone.}
The third component, visualised in Figure~\ref{fig:wrapper_pattern}, is a thin adapter layer that translates each neural network's native call signature into the common interface that the pipeline invokes. Because the three architectures accept different inputs --- Flux1 expects two separate token streams for observations and conditioning, while SimFormer and Flux1Joint expect a single joint sequence with a condition mask --- the pipelines would otherwise need architecture-specific logic. Wrappers eliminate this coupling: each pipeline type uses a single, fixed wrapper class, and the wrapper handles whatever translation is needed.
\begin{itemize}
  \item \texttt{ConditionalWrapper} --- used by \texttt{ConditionalPipeline}. Receives separate observation and conditioning tensors from the pipeline, expands their dimensions, and forwards them to the model's two-stream interface.
  \item \texttt{JointWrapper} --- used by \texttt{JointPipeline}. Passes the joint vector $z = (\theta, x)$ together with a binary \texttt{condition\_mask} that indicates which tokens are observed and which are to be inferred, and returns the model's prediction for the unconditioned dimensions.
  \item \texttt{UnconditionalWrapper} --- used by \texttt{UnconditionalPipeline}. Runs a joint model in unconditional mode by setting no conditioning, so that all dimensions are inferred jointly.
\end{itemize}
This adapter layer ensures that the pipeline and solver code see a uniform callable interface regardless of the underlying architecture, preserving the decoupling established by the other two axes.

\paragraph{Putting it together.}
Returning to the full picture in Figure~\ref{fig:architecture_overview}, the overall design can be summarised as a three-axis factorisation: \emph{which generative method} (flow matching, EDM, or score matching), \emph{which inference mode} (conditional, joint, or unconditional), and \emph{which neural network} (Flux1, SimFormer, Flux1Joint, or a user-provided model). The method and pipeline axes are fully independent, and the software composes them through the strategy pattern and inheritance, respectively. The architecture axis, however, is partially constrained: SimFormer and Flux1Joint work with all three pipelines
\footnote{Although the SimFormer and Flux1Joint models can in principle be used with the \texttt{ConditionalPipeline}, we recommend the \texttt{JointPipeline} instead: conditional density estimation is recovered by fixing the condition mask to \texttt{posterior} or \texttt{likelihood} mode during both training and sampling.}
, while Flux1 is restricted to the conditional pipeline owing to its two-stream design. Users who wish to extend GenSBI --- for example, by implementing a new generative formulation --- need only subclass \texttt{GenerativeMethod} and implement the five abstract methods; the existing pipelines, wrappers, and diagnostics will work without modification.

\subsection{Neural Network Architectures}
\label{sec:nn_architectures}

The choice of neural network architecture is central to any density estimation framework, and it is here that GenSBI departs most visibly from existing SBI libraries. Traditional NPE implementations parameterise the posterior through normalizing flows --- typically masked autoregressive flows (MAFs)~\cite{Papamakarios:2017tec} or neural spline flows (NSFs)~\cite{Durkan:2019nsq} --- in which the network defines an explicit bijection that maps a simple base distribution to the target posterior via the change-of-variables formula. Because the transformation must be analytically invertible with a tractable Jacobian, the design space of admissible architectures is severely constrained --- a price paid for the single-pass exact likelihoods discussed in Section~\ref{sec:density_estimator_choice}. The flow matching and diffusion formulations adopted in GenSBI (Section~\ref{sec:generative_models}) lift this restriction entirely: the network is trained to regress a time-dependent velocity field $v_\theta(x, t)$ or score function $\nabla_x \log p_t(x)$, and samples are obtained by numerically integrating the corresponding ODE or SDE. Since the training objective no longer requires invertibility, any sufficiently expressive function approximator can serve as the backbone. Transformers are a natural choice, given their well-documented scaling properties~\cite{Peebles:2023dit} and their ability to process heterogeneous token sequences through attention.

GenSBI provides three transformer-based architectures, each targeting a different use case. SimFormer is adapted from the ``All-in-one'' model of Gloeckler et al.~\cite{Gloeckler:2024simformer} and is designed for joint density estimation. Flux1 is derived from Flux.1 by Black Forest Labs~\cite{BlackForestLabs:2025flux}, originally an image generation model, and repurposed here as a conditional density estimator for scientific data. Flux1Joint is a new architecture introduced in GenSBI that combines design elements from both SimFormer and Flux1 to enable joint estimation with modern transformer blocks.

\begin{figure}[t]
  \centering
  \includegraphics[width=\textwidth]{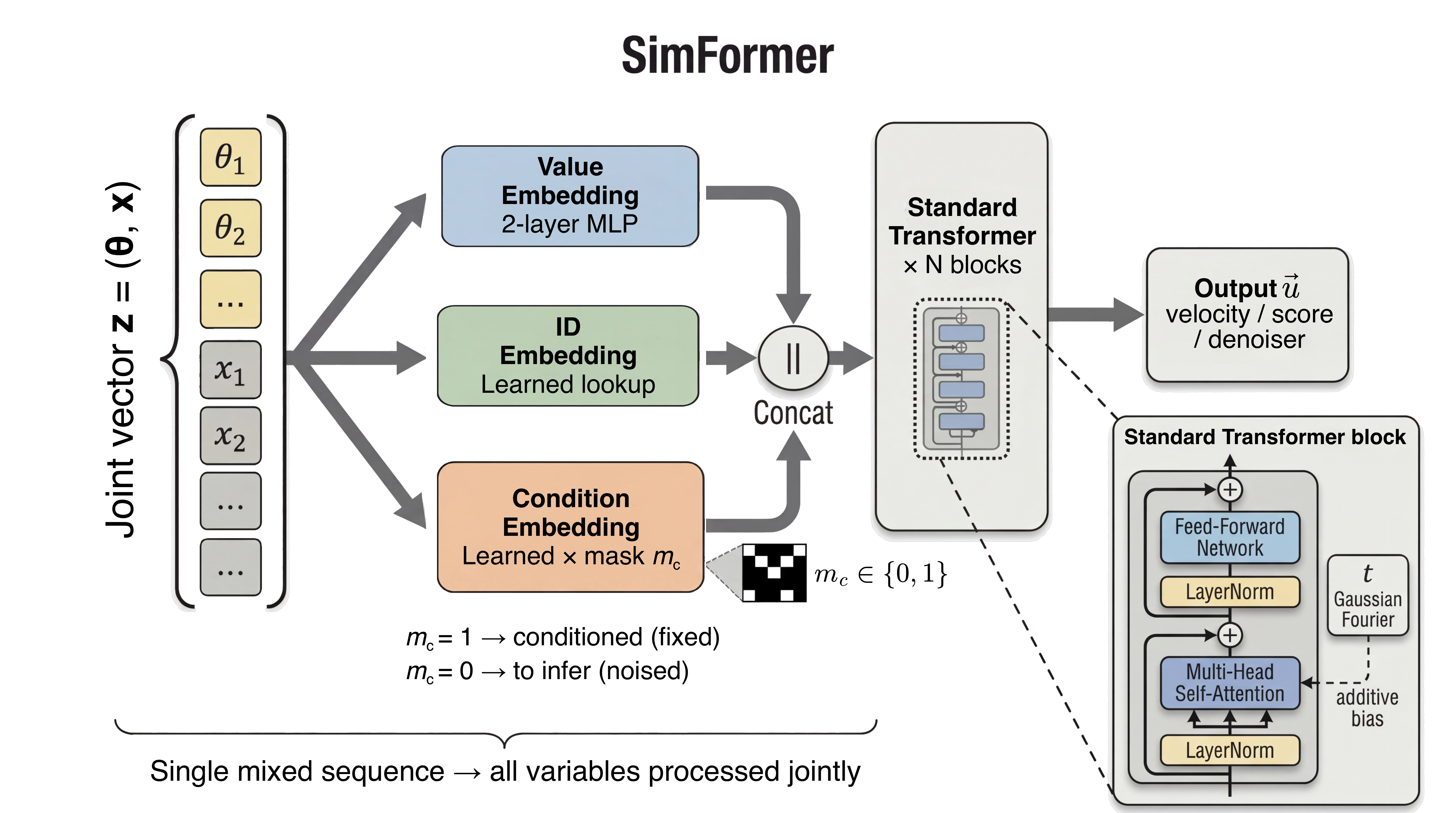}
  \caption{SimFormer architecture. Each scalar component of the joint vector $z = (\theta, x)$ is treated as an individual token. Three per-token embeddings --- value, ID, and condition --- are concatenated and processed by a standard transformer stack with multi-head self-attention and feed-forward blocks. The condition embedding, modulated by a binary mask $m_C$, enables flexible conditioning at inference time. The timestep $t$ is injected via a Gaussian Fourier embedding as an additive bias.}
  \label{fig:simformer_arch}
\end{figure}

\paragraph{SimFormer.}
The SimFormer architecture \cite{Gloeckler:2024simformer}, illustrated in Figure~\ref{fig:simformer_arch}, was originally introduced as a diffusion-based transformer for ``all-in-one'' posterior and likelihood estimation. Its defining feature is that it operates on a single token sequence representing the full joint vector $z = (\theta, x)$, rather than processing parameters and observations through separate streams. Each scalar component of $z$ is treated as an individual token, and the model processes three per-token embeddings: a \emph{value embedding} that encodes the numerical content through a two-layer MLP, an \emph{ID embedding} that identifies which variable each token corresponds to (a learned embedding indexed by position), and a \emph{condition embedding} that indicates whether a token is observed or to be inferred. These three embeddings are concatenated to form the input to a standard transformer stack consisting of alternating multi-head self-attention and feed-forward blocks with skip connections. The diffusion timestep $t$ is encoded via a learnable Gaussian Fourier embedding and injected into each dense block as an additive bias after the context projection.

The condition embedding is what enables SimFormer's flexible inference modes. During training, a binary mask $m_C \in \{0, 1\}^d$ is drawn randomly for each sample in the batch, indicating which tokens are conditioned ($m_C = 1$) and which are to be inferred ($m_C = 0$). This mask acts on the model in two ways. First, it modulates the condition embedding: a learnable vector is multiplied element-wise by $m_C$ and concatenated with the value and ID embeddings, so the transformer receives an explicit signal distinguishing conditioned from unconditioned tokens. Second, conditioned tokens are not noised --- they retain their clean values throughout the diffusion or flow process --- and the training loss is computed only on the unconditioned tokens, so the network learns to predict the target field exclusively where inference is needed. At test time, one simply fixes $m_C$ to the desired conditioning pattern: setting $m_C = 1$ on observation tokens and $m_C = 0$ on parameter tokens recovers the posterior $p(\theta | x_\mathrm{obs})$. Because the model has been trained on random masking patterns, it can handle posterior inference, likelihood queries, joint sampling, and the computation of arbitrary marginals without retraining.

In GenSBI, the SimFormer implementation extends the original diffusion-only formulation to support flow matching as well. The network architecture itself is agnostic to the generative method; it simply maps $(z_t, t, m_C) \mapsto \vec{u}$, where $\vec{u}$ is the velocity, score, or denoiser output depending on the method. This separation is enforced through GenSBI's strategy pattern: the generative method determines the training objective and the interpretation of $\vec{u}$, while the neural network remains unchanged.

\begin{figure}[t]
  \centering
  \includegraphics[width=\textwidth]{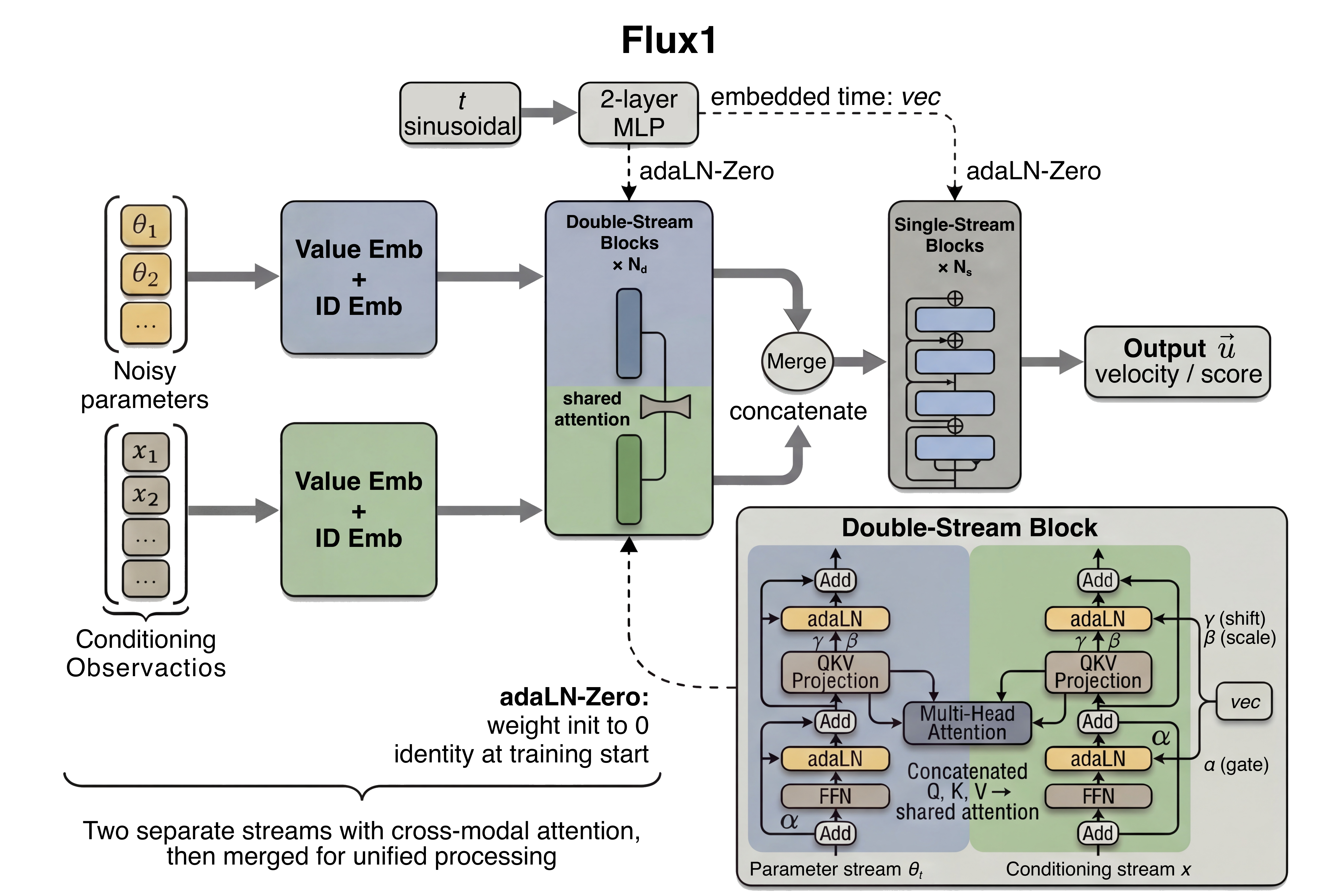}
  \caption{Flux1 architecture. Two separate token streams --- an observation stream (noisy parameters $\theta_t$) and a conditioning stream (observed data $x$) --- are processed through double-stream blocks with shared attention and independent residual paths, followed by single-stream blocks over the merged sequence. Both block types use adaLN-Zero modulation, where the timestep $t$ controls per-dimension shift, scale, and gate parameters.}
  \label{fig:flux1_arch}
\end{figure}

\paragraph{Flux1.}
The Flux1 architecture, shown in Figure~\ref{fig:flux1_arch}, is adapted from the Flux.1 model developed by Black Forest Labs for high-resolution text-to-image generation. Where SimFormer processes parameters and observations as a single mixed sequence, Flux1 maintains two separate token streams: an \emph{observation stream} carrying the noisy parameters $\theta_t$ that are being denoised, and a \emph{conditioning stream} carrying the observed data $x$. This separation mirrors the standard conditional NPE setup and is natural when the roles of $\theta$ (to be inferred) and $x$ (given) are fixed at training time.

The architecture processes these two streams through a two-stage pipeline. The first stage consists of \emph{double-stream blocks}, in which the observation and conditioning streams maintain independent layer normalization, QKV projections, and feed-forward networks, but share attention: the queries, keys, and values from both streams are concatenated before computing the attention matrix, so that each stream attends to the full joint context while preserving separate residual paths. The second stage consists of \emph{single-stream blocks}, where the two streams are concatenated into a single sequence and processed through a unified attention and feed-forward path. This two-stage design allows the model to first build modality-specific representations with cross-modal attention and then refine them through full self-attention over the merged sequence.

Both block types use adaptive layer normalization with zero initialization (adaLN-Zero), following the Diffusion Transformer (DiT) design of Peebles and Xie~\cite{Peebles:2023dit}. The diffusion or flow timestep $t$ is encoded as a sinusoidal embedding and projected through a two-layer MLP to produce a conditioning vector, termed $vec$. Each transformer block applies a modulation layer that transforms $vec$ into per-dimension shift, scale, and gate parameters: the shift and scale are applied after layer normalization, and the gate multiplies the block's output before it enters the residual connection. The modulation weights are initialized to zero, which makes each block act as an identity function at the start of training and yields more stable optimization dynamics \cite{Peebles:2023dit}.

A distinctive feature of GenSBI's Flux1 implementation is its flexible token identification system, which allows the same transformer architecture to process unstructured parameter vectors, one-dimensional sequences, and two-dimensional images by changing only the ID embedding strategy and the input reshaping. Because the model is designed for general scientific data rather than pixel grids, it supports three families of encoding strategies, each suited to different data semantics. Learned absolute embeddings assign a unique trainable vector to each token position, making them well suited to unstructured parameter vectors where positions carry no intrinsic ordering. Sinusoidal positional encodings inject absolute position information through fixed sine and cosine functions of varying frequency; they are preferable when the location of a feature within a sequence is physically meaningful --- for instance, a peak appearing at $t = 2\,\mathrm{s}$ in a time series carries different information from the same peak at $t = 5\,\mathrm{s}$. Rotary positional embeddings (RoPE) encode relative distances between tokens rather than absolute positions, which is advantageous when the spacing of features matters more than where they occur --- for example, in a star field image where the relative separations between sources constrain cluster properties independently of where the cluster falls on the detector. Both sinusoidal and RoPE encodings are available in one-dimensional and two-dimensional variants: the 1D versions operate on a scalar position index suited to sequential data, while the 2D versions operate on a grid coordinate pair $(i, j)$ suited to image-structured data. Each strategy can be applied independently to the observation and conditioning streams, and can be combined with the value embeddings by summation or concatenation --- the latter being preferable for small models where summing positional and value signals into a low-dimensional space could create interference.

For two-dimensional data such as images, GenSBI adopts a vision-transformer-style patchification~\cite{Dosovitskiy:2020vit, BlackForestLabs:2025flux}: the input is reshaped into a sequence of patches ($2 \times 2$ by default), each becoming a single token whose feature vector concatenates the pixel values within the patch. Each token is then assigned a three-component ID of the form $(\mathrm{semantic\_id},\, i,\, j)$, where $\mathrm{semantic\_id}$ identifies the token group (e.g.\ distinguishing observation from conditioning tokens when both are present) and $(i, j)$ specifies the patch position on the spatial grid. This scheme preserves the two-dimensional structure of the data through the attention mechanism and enables the use of 2D RoPE embeddings, so that the transformer can exploit spatial correlations between neighbouring patches --- an approach that mirrors the standard practice in latent diffusion models for image generation~\cite{Rombach:2022ldm}.

\begin{figure}[t]
  \centering
  \includegraphics[width=\textwidth]{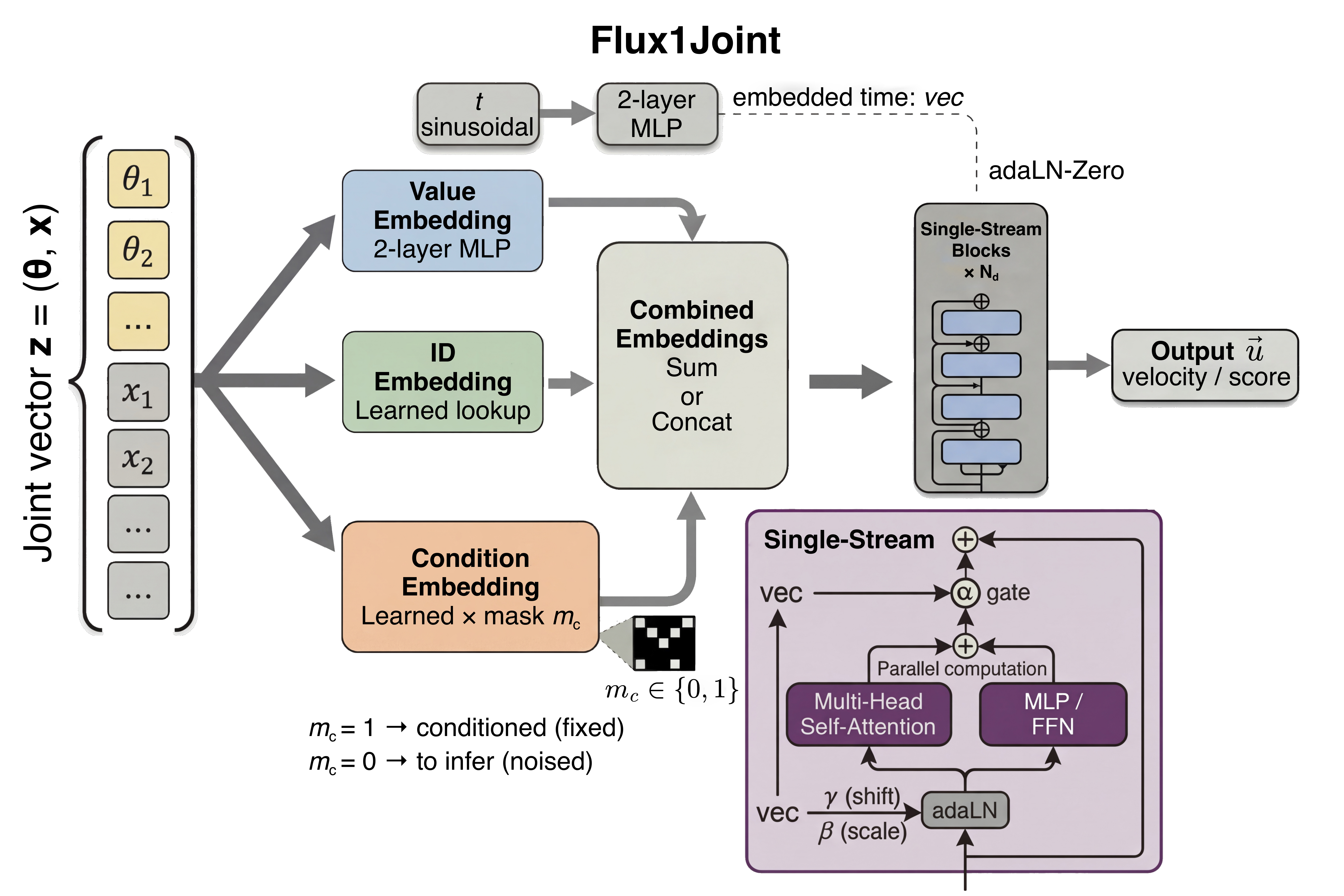}
  \caption{Flux1Joint architecture. Like SimFormer, the model processes a single joint token sequence $z = (\theta, x)$ with per-token value, ID, and condition embeddings. Unlike SimFormer, the transformer stack uses Flux1's single-stream blocks with adaLN-Zero modulation and parallel attention-MLP computation, combining SimFormer's flexible conditioning mechanism with Flux1's more expressive block design.}
  \label{fig:flux1joint_arch}
\end{figure}

\paragraph{Flux1Joint.}
Flux1Joint, illustrated in Figure~\ref{fig:flux1joint_arch}, is a new architecture introduced in GenSBI that combines the single-stream processing and adaLN-Zero modulation of Flux1's transformer blocks with SimFormer's condition mask mechanism and ID embeddings. The motivation is to obtain an architecture that supports joint density estimation and post-training conditioning --- as SimFormer does --- while using the more expressive transformer blocks from the Flux1 family.

Concretely, Flux1Joint processes a single token sequence representing the joint vector $z = (\theta, x)$, as SimFormer does, but replaces SimFormer's standard attention-plus-feed-forward blocks with Flux1's single-stream blocks that include adaLN-Zero gating and parallel linear layers. The model accepts a condition mask and ID embeddings in the same manner as SimFormer: per-token value, ID, and condition embeddings are combined (by summation or concatenation) and fed through the single-stream transformer stack, with the timestep injected via the adaLN-Zero modulation. Flux1Joint omits Flux1's double-stream blocks entirely, since the joint-modeling paradigm does not distinguish between parameter and observation streams at the architectural level --- the condition mask handles this distinction dynamically.
Flux1Joint thus retains SimFormer's post-training conditioning flexibility while benefiting from the more expressive Flux1 block design.

A concurrent and independent implementation of masked conditional flow matching for joint SBI is OneFlowSBI~\cite{Nautiyal:2026oneflowsbi}, which parameterises the velocity field with a residual MLP whose blocks combine layer normalisation, adaLN-zero modulation, and feedforward layers --- functionally equivalent to the feedforward portion of Flux1Joint's single-stream blocks, but without self-attention. By dispensing with attention, OneFlowSBI obtains a lighter model; self-attention, however, lets each token interact with every other token in the joint sequence, improving expressiveness on tasks with complex inter-variable correlations --- a pattern visible in the SLCP benchmark (Section~\ref{sec:benchmarks}). Both approaches can incorporate domain-specific encoders for structured data, but in GenSBI a user-provided embedding feeds into an unmodified transformer backbone, whereas OneFlowSBI requires redesigning the core network per modality (see Appendices C.2--C.3 of~\cite{Nautiyal:2026oneflowsbi}). More broadly, our transformer architectures achieves competitive SBIBM performance with a near-uniform model and training configurations (Appendix~\ref{app:training_configs}), while feedforward architectures generally require careful per-task tuning (see Table 2 of~\cite{Nautiyal:2026oneflowsbi}), a trade-off for their faster training times. Notably, \texttt{OneFlowSBI} does not release its source code, which limits reproducibility and practical adoption by the community.

\paragraph{Model wrappers.}
Because our three architectures have different native call signatures, each is paired with a lightweight model wrapper (Section~\ref{sec:architecture}) that presents the uniform interface the pipeline expects. The wrapper is transparent to the rest of the library: once wrapped, any model --- built-in or user-provided --- can be used interchangeably within a given pipeline type.

\paragraph{Custom models and embedding networks.}
The three transformer architectures described above are designed to serve as strong, general-purpose defaults for a wide range of inference problems. When the input data has specialised structure --- such as time series, 2D images, or graph-valued observations --- users can extend the framework at two levels. The most common scenario is to embed the conditioning data with a domain-specific encoder: the user defines a custom network (for example, a 1D or 2D convolutional neural network) that maps the raw observation into a compact latent representation. These learned representations are then passed directly to one of the built-in transformer models as conditioning tokens, so the user benefits from the full transformer backbone without modifying it. Alternatively, users who require full control over the model architecture can implement an entirely custom velocity field or score network --- for instance, a simple multi-layer perceptron for low-dimensional density estimation --- and plug it into the pipeline through the appropriate model wrapper. The pipeline imposes no constraints on the model's internal structure; it only requires that the wrapped model maps a noisy input, a timestep, and (for conditional pipelines) conditioning tokens to an output of matching shape. End-to-end examples for both use cases --- including CNN-based embedding of gravitational wave time series and strong gravitational lensing images, as well as a custom MLP for unconditional density estimation --- are provided in the package documentation and in the \texttt{GenSBI-examples} repository~\cite{GenSBI_examples}.

\subsection{Solvers}
\label{sec:solvers}

A trained generative model defines a velocity field, score function, or denoiser, but it does not by itself specify how to \emph{numerically integrate} the corresponding differential equation at inference time. GenSBI separates the \emph{solver} --- the numerical integrator (e.g.\ Euler, Heun, Dopri5) --- from the \emph{model} and the \emph{generative method}. Within a given generative method, the user can swap solvers freely without retraining. All solvers inherit from a common abstract base class that exposes a unified \texttt{sample} interface and, where mathematically supported, a \texttt{log\_prob} interface for exact density evaluation.

\paragraph{Flow matching solvers.}
The default solver for flow matching is \texttt{FMODESolver}, which integrates the learned velocity field $v_\phi(\theta_t, t, x)$ from $t=0$ (noise) to $t=1$ (data) using the \texttt{diffrax}  library~\cite{Kidger:2022diffrax}. Both fixed-step methods (Euler, Heun, midpoint) and adaptive-step methods (Dopri5 with PID step-size control) are available. For problems where the ODE trajectories are nearly straight --- as expected from optimal-transport conditional flow matching --- an Euler solver with 100--200 steps typically suffices; adaptive solvers become useful when the velocity field varies sharply in certain time regions, trading a modest computational overhead for guaranteed error control.

The \texttt{FMODESolver} also supports exact log-probability computation via the continuous change-of-variables formula. Given a sample $\theta_1$ at the data end, the solver integrates the velocity field and its divergence backward from $t=1$ to $t=0$, accumulating the log-determinant of the Jacobian along the trajectory. The total log-probability is $\log q_\phi(\theta_1 \mid x) = \log p_0(\theta_0) + \int_1^0 \nabla_\theta \cdot v_\phi(\theta_t, t, x)\, dt$, where $p_0$ is the known prior (source) density. The divergence can be computed exactly via automatic differentiation, scaling as $\mathcal{O}(d)$ network evaluations per step, or estimated stochastically using the Hutchinson trace estimator with Rademacher probe vectors~\cite{Hutchinson:1989trace}, which requires only one additional evaluation but introduces variance. This capability underpins potential neural likelihood estimation (NLE) workflows, though as discussed in Section~\ref{sec:sbi}, the per-evaluation cost of solving an ODE currently makes it impractical for MCMC-based inference.

Beyond deterministic sampling, GenSBI provides two stochastic SDE solvers for flow matching --- \texttt{ZeroEndsSolver} and \texttt{NonSingularSolver} --- both derived from the framework of Singh and Fischer~\cite{Singh:2024ssfm}. These solvers augment the learned velocity field with a score-dependent drift correction and a time-dependent diffusion term, converting the deterministic ODE into an SDE that shares the same marginal distributions at every time step. The two variants differ in how the diffusion coefficient behaves at the integration boundaries: the \texttt{ZeroEndsSolver} uses $\tilde{g}(t) = \alpha \sqrt{t(1-t)}$, which vanishes at both endpoints and ensures exact boundary conditions; the \texttt{NonSingularSolver} uses $\tilde{g}(t) = \alpha \sqrt{1-t}$, which avoids the singularity at $t=0$ and may be preferable when integration starts from a small but nonzero $t = \varepsilon$. In both cases, the perturbation strength is controlled by a single hyperparameter $\alpha$, and the score function needed for the drift correction is obtained analytically from the velocity field using the identity relating conditional velocity and score under the affine Gaussian probability path~\cite{Singh:2024ssfm}. Adding stochasticity can improve sample diversity and distribution coverage at the cost of requiring a random key and per-sample integration via \texttt{jax.vmap} over independent Brownian paths.

\paragraph{Score matching solvers.}
For score matching models, GenSBI provides two complementary solvers. The default is \texttt{SMSDESolver}, which integrates the reverse-time SDE from $t=T$ toward $t=\varepsilon$ using Euler--Maruyama discretization. This requires both the learned score $s_\phi(\theta_t, t, x) \approx \nabla_\theta \log p_t(\theta_t \mid x)$ and the forward-process coefficients $f(\theta, t)$ and $g(t)$, which are provided by the SDE scheduler (VP or VE). The deterministic counterpart, \texttt{SMODESolver}, follows the probability flow ODE that shares the same marginal distributions as the reverse SDE but produces deterministic trajectories, enabling reproducible sampling and, in principle, exact log-probability evaluation%
\footnote{While the PF ODE is supported and theoretically grounded, flow matching is generally preferred if exact log-probability evaluation is needed. Although the PF ODE shares the same marginal distributions as the reverse SDE in the exact-score limit~\cite{Song:2020sde}, minimising the denoising score matching objective controls the KL divergence of the \emph{stochastic} generative model but is insufficient, in general, to bound the likelihood of the deterministic ODE~\cite{albergo2025stochasticinterpolantsunifyingframework}. Intuitively, the SDE's diffusion term acts as an implicit Langevin corrector that drives samples back toward the target marginal at each time step, actively compensating for score estimation errors~\cite{Karras:2022edm}; the deterministic ODE lacks this self-correcting mechanism and can suffer from collapse errors when the learned score is imperfect.}.
The \texttt{SMODESolver} supports fixed-step (Euler, Heun) and adaptive-step (Dopri5) integration, mirroring the interface of the flow matching \texttt{FMODESolver}.

\paragraph{EDM diffusion solvers.}
The \texttt{EDMSolver} implements the sampling algorithm of Karras et al.~\cite{Karras:2022edm}, operating in $\sigma$-space (noise level) rather than a conventional time variable. Three noise schedules are available --- EDM, VP, and VE --- and can be selected independently for training and sampling. The solver steps through a decreasing schedule $\sigma_\mathrm{max} \to 0$, applying the EDM-preconditioned denoiser at each step; a second-order Heun corrector is used by default, averaging an initial Euler prediction with a second denoiser evaluation at the predicted point to reduce truncation error.

Four stochastic churn parameters --- $S_\mathrm{churn}$, $S_\mathrm{min}$, $S_\mathrm{max}$, and $S_\mathrm{noise}$ --- control optional noise injection during sampling~\cite{Karras:2022edm}. With $S_\mathrm{churn} = 0$ (the default), sampling is fully deterministic. Setting $S_\mathrm{churn} > 0$ injects Gaussian noise at intermediate steps where $\sigma \in [S_\mathrm{min},\, S_\mathrm{max}]$, which can improve sample quality for complex, multimodal distributions by mitigating the accumulation of discretisation error. The optimal churn parameters are problem-dependent; Karras et al.~\cite{Karras:2022edm} provide detailed tuning guidance in their Appendix~E. Since the churn parameters affect only the sampling procedure, they can be adjusted post-training without retraining the model.

\paragraph{Solver--method compatibility and practical guidance.}
The key design principle is that solvers are interchangeable \emph{within} a generative method: any flow matching solver can be used with any model trained via \texttt{FlowMatchingMethod}, and likewise for the score matching and EDM families. The choice of solver affects sampling speed, sample quality, and whether log-probability evaluation is available, but never requires retraining. Table~\ref{tab:solvers} summarises the full solver inventory, listing the available numerical integrators for each solver class, with the recommended default marked by a dagger. In practice, we recommend starting with the default solver for each method and exploring alternatives only when sample quality is insufficient. In particular, it is useful to use the stochastic variants of flow matching and diffusion EDM when the resulting distributions are under-dispersed. 

\begin{table}[t]
\centering
\caption{Solver inventory in GenSBI. Each generative method provides one or more solver classes; within a given method, solvers are interchangeable without retraining. The \emph{Numerical integrators} column lists the discretisation schemes available for each solver, with the recommended default marked by a dagger~($\dagger$). ``Stochastic'' indicates whether the solver injects noise during sampling; for \texttt{EDMSolver} stochasticity is configurable via the churn parameters $S_\mathrm{churn}$, $S_\mathrm{min}$, $S_\mathrm{max}$, $S_\mathrm{noise}$~\cite{Karras:2022edm}. ``Log-prob'' indicates support for exact log-probability evaluation via the continuous change-of-variables formula.}
\label{tab:solvers}
\small
\begin{tabular}{@{}llllcc@{}}
\toprule
\textbf{Solver class} & \textbf{DE type} & \textbf{Numerical integrators} & \textbf{Stochastic} & \textbf{Log-prob} \\
\midrule
\multicolumn{5}{@{}l}{\textbf{Flow Matching} (\texttt{FlowMatchingMethod})} \\[2pt]
\texttt{FMODESolver}       & ODE & Euler$^\dagger$, Heun, Midpoint, Dopri5 & --- & \checkmark \\
\texttt{ZeroEndsSolver}    & SDE & Euler$^\dagger$, Heun, SEA, ShARK       & \checkmark & --- \\
\texttt{NonSingularSolver} & SDE & Euler$^\dagger$, Heun, SEA, ShARK       & \checkmark & --- \\
\midrule
\multicolumn{5}{@{}l}{\textbf{Score Matching} (\texttt{ScoreMatchingMethod}; VP or VE SDE)} \\[2pt]
\texttt{SMSDESolver}       & Reverse SDE & Euler$^\dagger$, Heun, SEA, ShARK & \checkmark & ---  \\
\texttt{SMODESolver}       & PF-ODE      & Euler$^\dagger$, Heun, Dopri5     & ---  & \checkmark$^{1}$ \\
\midrule
\multicolumn{5}{@{}l}{\textbf{EDM Diffusion} (\texttt{DiffusionEDMMethod}; EDM, VP, or VE schedule)} \\[2pt]
\texttt{EDMSolver}         & ODE$^2$ & Euler, Heun$^\dagger$ (2nd-order) & cfg. & --- \\
\bottomrule
\end{tabular}
\vspace{2pt}
\par\noindent{\footnotesize $^1$\,Log-probability via the PF-ODE change-of-variables formula is supported, but can be imprecise.}
\par\noindent{\footnotesize $^2$\,EDM uses an ODE solver with optional noise injection during sampling (crunch).}
\end{table}

\subsection{High-Level Recipes API}
\label{sec:recipes}

While the modular architecture described in the previous sections gives users full control over every component --- generative method, neural network, solver, and path --- assembling these pieces correctly requires familiarity with how they interact. To lower the barrier to entry, GenSBI provides a \emph{recipes} layer: pre-configured pipelines that bundle a specific architecture with a specific generative method and expose a single, self-contained object for the entire training and inference workflow.

The recipes follow the pipeline class hierarchy introduced in Section~\ref{sec:architecture}. The three intermediate pipelines --- \texttt{ConditionalPipeline}, \texttt{JointPipeline}, and \texttt{UnconditionalPipeline} --- each accept an arbitrary \texttt{GenerativeMethod} and model, and handle the mode-specific concerns (batch structure, wrapping, and sampler construction) while inheriting the shared training infrastructure (optimizer, EMA, checkpointing, early stopping) from \texttt{AbstractPipeline}.

On top of this hierarchy sit the pre-configured recipe pipelines, formed by combining the supported architectures with the three generative methods. Because Flux1 is restricted to conditional density estimation, its recipe pipelines (e.g.\ \texttt{Flux1FlowPipeline}, \texttt{Flux1DiffusionPipeline}, \texttt{Flux1SMPipeline}) instantiate a Flux1 model inside a \texttt{ConditionalPipeline}. SimFormer and Flux1Joint recipes (e.g.\ \texttt{SimformerDiffusionPipeline}, \texttt{Flux1JointFlowPipeline}) instantiate their models inside a \texttt{JointPipeline}, which supports conditional, joint, and unconditional inference through the condition mask mechanism described in Section~\ref{sec:architecture}. All recipe classes supply sensible default parameters for both the architecture (number of heads, depth, embedding dimensions) and the training loop (learning rate, number of steps, EMA decay rate), so that users can instantiate a working pipeline by specifying only the datasets, the parameter dimensionality, and the observation dimensionality.
All defaults can be overridden by passing custom \texttt{params} or \texttt{training\_config} dictionaries, or by using YAML configuration files via the \texttt{init\_pipeline\_from\_config} class method.

Listing~\ref{lst:flux1_flow} sketches a typical end-to-end workflow using the \texttt{Flux1FlowPipeline}. Starting from simulated parameter--observation pairs loaded through Grain\footnote{\url{https://github.com/google/grain}} data loaders, the user instantiates the pipeline with custom architecture parameters, launches training with a single call to \texttt{pipeline.train()}, and draws posterior samples via \texttt{pipeline.sample()}. The listing is illustrative pseudocode intended to convey the high-level API; complete, ready-to-run tutorials covering data preparation, training configuration, and post-processing are available in the package documentation at \href{https://gensbi.com}{gensbi.com}.

\begin{figure}[t]
\caption{Illustrative pseudocode for a \texttt{Flux1FlowPipeline} workflow: data simulation, training, and posterior sampling. Complete, ready-to-run tutorials are available at \href{https://gensbi.com}{gensbi.com}.}
\label{lst:flux1_flow}
\begin{lstlisting}[language=Python]
# Minimal illustrative example; for complete,
# ready-to-run tutorials see https://gensbi.com
from gensbi.recipes import Flux1FlowPipeline
from gensbi.models import Flux1Params

# 1. Simulate training data
train_data = simulator(jax.random.PRNGKey(0), 100_000)
val_data   = simulator(jax.random.PRNGKey(1), 2_000)

# 2. Build efficient data loaders with Grain
train_ds = (grain.MapDataset.source(train_data)
    .shuffle(42).repeat().to_iter_dataset()
    .batch(256))
val_ds = (grain.MapDataset.source(val_data)
    .shuffle(42).repeat().to_iter_dataset()
    .batch(256))

# 3. Configure model and training
params = Flux1Params(
    in_channels=1, context_in_dim=1,
    num_heads=2, depth=4, depth_single_blocks=8,
    dim_obs=3, dim_cond=3, rngs=nnx.Rngs(42))

pipeline = Flux1FlowPipeline(
    train_ds, val_ds, dim_obs=3, dim_cond=3,
    params=params)

# 4. Train and sample
pipeline.train(nnx.Rngs(42))
samples = pipeline.sample(
    jax.random.PRNGKey(0), x_obs, nsamples=100_000)
\end{lstlisting}
\end{figure}

The solver used at inference time can be changed post-training without retraining the model. In the example above, the default deterministic ODE solver is used; the user can instead pass a stochastic solver such as the \texttt{ZeroEndsSolver} to obtain SDE-based samples. This flexibility, discussed in detail in Section~\ref{sec:solvers}, allows practitioners to explore the speed--quality--stochasticity tradeoff without additional training cost.

The recipes layer also provides explicit extension points. Users who require a custom architecture can bypass the pre-configured pipelines and instantiate an intermediate pipeline directly, passing their own model alongside any \texttt{GenerativeMethod}; the pipeline handles all method-specific logic, so the custom model only needs to implement a standard forward pass. Similarly, subclassing \texttt{AbstractPipeline} allows users to override individual components (optimizer schedule, training loop, checkpoint strategy) while inheriting the rest. The recipes thus serve as convenient defaults rather than rigid constraints.

\subsection{Diagnostics and Validation}
\label{sec:diagnostics}

For scientific applications of simulation-based inference, a well-calibrated posterior is not merely desirable --- it is a prerequisite. An approximate posterior that systematically under-covers or over-covers the true parameter values can lead to erroneous physical conclusions, regardless of how expressive the underlying generative model may be. Yet none of the training objectives discussed in Section~\ref{sec:generative_models} directly enforce calibration: a model that minimizes the flow matching loss or denoising score matching loss to zero would recover the true posterior, but in practice, finite data, limited network capacity, and discretisation errors in the ODE/SDE solver all introduce approximation gaps. External diagnostics are therefore essential to verify whether a trained model can be trusted for downstream scientific analysis. GenSBI integrates four complementary calibration diagnostics as first-class components of the library, adopting an interface similar to the \texttt{sbi} library \cite{BoeltsDeistler_sbi_2025}.

\paragraph{Simulation-Based Calibration (SBC).}
SBC~\cite{Talts:2018sbc} exploits a self-consistency property of Bayesian inference: if parameters $\theta$ are drawn from the prior and observations $x$ from the simulator conditioned on $\theta$, then the rank of the true $\theta$ within a set of posterior samples drawn from $q_\phi(\theta \mid x)$ must be uniformly distributed. Deviations from uniformity reveal specific failure modes --- a U-shaped rank histogram indicates overconfidence (under-dispersion), an inverted-U shape signals underconfidence (over-dispersion), and skewness indicates systematic bias. SBC is a \emph{necessary} condition for posterior correctness: a well-calibrated posterior will pass SBC, but an uninformative posterior (e.g., one that simply returns the prior) can also pass, so SBC should always be supplemented with tests that assess the posterior's informativeness. GenSBI provides \texttt{run\_sbc}, \texttt{sbc\_rank\_plot}, and \texttt{check\_sbc} functions that compute ranks, produce diagnostic histograms, and perform quantitative uniformity tests (Kolmogorov--Smirnov) alongside a data-averaged posterior sanity check.

\paragraph{TARP (Tests of Accuracy with Random Points).}
TARP~\cite{Lemos:2023tarp} provides a complementary global diagnostic based on expected coverage probabilities. For each simulated pair $(\theta, x)$, TARP computes the empirical fraction of posterior samples that fall closer to a random reference point than the true parameter value, and aggregates these fractions across many pairs to construct an expected coverage probability (ECP) curve as a function of the nominal credible level $\alpha$. A well-calibrated posterior produces an ECP curve that lies on the diagonal; curves above the diagonal indicate underconfidence, and curves below indicate overconfidence. Unlike SBC, TARP operates directly on distances in parameter space and does not require computing marginal ranks, making it sensitive to correlations between parameters. GenSBI's implementation extends the standard TARP diagnostic with Jeffreys confidence intervals on the ECP curve, providing a principled quantification of the statistical uncertainty in the coverage estimate.

\paragraph{LC2ST (Local Classifier Two-Sample Test).}
While SBC and TARP assess calibration \emph{globally} --- averaged over the prior predictive distribution --- they cannot guarantee correctness at any specific observation. LC2ST~\cite{Linhart:2023lc2st} addresses this limitation by providing a \emph{local} diagnostic. A binary classifier is trained on a calibration dataset to distinguish between samples drawn from the true joint $p(\theta, x)$ and samples constructed by pairing approximate posterior draws $q_\phi(\theta \mid x)$ with the corresponding observations. At test time, the classifier is evaluated on posterior samples generated for a specific observation $x_\mathrm{obs}$: if it can reliably distinguish approximate from true samples, the null hypothesis of posterior accuracy is rejected. LC2ST does not require access to samples from the true posterior, making it applicable to problems where reference posteriors are unavailable. GenSBI provides the \texttt{LC2ST} class and \texttt{plot\_lc2st} function for training the classifier and visualising the results.

\paragraph{Marginal Coverage.}
As a lightweight complement to the tests above, GenSBI includes empirical marginal coverage diagnostics. For each parameter dimension $d$ and a grid of nominal credible levels $\alpha$, the marginal coverage test computes the fraction of simulated pairs for which the true parameter $\theta^{(d)}$ falls within the $\alpha$-credible interval of the approximate marginal posterior. The resulting coverage curve should follow the diagonal; systematic deviations indicate marginal miscalibration. GenSBI supports both histogram-based and kernel density estimation (KDE) methods for constructing the marginal credible intervals via \texttt{compute\_marginal\_coverage} and \texttt{plot\_marginal\_coverage}.

These four diagnostics test different and complementary properties: SBC and TARP provide global calibration checks averaged over the prior, differing in whether they operate on marginal ranks or distances in parameter space; LC2ST offers observation-specific correctness assessment; and marginal coverage provides a per-dimension summary. In practice, we recommend running all available diagnostics and combining them with posterior predictive checks to build a comprehensive picture of the model's reliability. The SBC, TARP, and LC2ST implementations mirror those of the \texttt{sbi} library~\cite{BoeltsDeistler_sbi_2025}, while the marginal coverage diagnostic and the Jeffreys confidence intervals on the TARP ECP curve follow the \texttt{swyft} library~\cite{Miller2022}; all are reimplemented natively in JAX for seamless integration with the rest of GenSBI.

\subsection{JAX Ecosystem Integration}
\label{sec:jax_integration}

A core design principle of GenSBI is to build on well-established JAX ecosystem libraries where possible, rather than reimplement functionality that already exists in mature, independently maintained packages. Many existing implementations of flow matching and diffusion models for SBI bundle custom numerical solvers, training utilities, and distribution primitives within the library itself. GenSBI takes a different approach: the library's contribution is to integrate and unify these external tools into a coherent SBI framework, so that each component --- neural network construction, numerical integration, probabilistic prior specification, and model serialisation --- is handled by specialised software that the broader JAX community actively maintains and improves. The practical consequence for users is that GenSBI interoperates with standard JAX tools they may already use in other projects, and that the entire pipeline, from training through sampling, is compatible with JAX's program transformations out of the box.

All neural network modules in GenSBI --- the three transformer architectures described in Section~\ref{sec:nn_architectures}, their embedding layers, and the model wrappers --- are implemented as \texttt{Flax nnx} modules~\cite{Flax:2024}. Because \texttt{nnx} modules participate natively in JAX's functional transformation system, GenSBI's training steps, loss functions, and sampling routines are compiled end-to-end via \texttt{jit}, and batched evaluation over multiple observations or posterior samples uses \texttt{vmap}. The same compiled code executes on CPUs, GPUs, and TPUs without modification; switching between accelerators requires only changing the JAX backend configuration, not the GenSBI code itself. For numerical integration, the flow matching and score matching solvers delegate ODE and SDE integration to \texttt{diffrax} ~\cite{Kidger:2022diffrax}, as described in Section~\ref{sec:solvers}, providing access to both fixed-step and adaptive-step methods through a common, differentiable, and JIT-compiled interface.

Prior distributions are specified using NumPyro's distribution API~\cite{Phan:2019numpyro}. This is a deliberate design decision: rather than defining a custom distribution interface, GenSBI accepts any \texttt{numpyro} \texttt{Distribution} object as the source distribution for the generative process. Users can employ the full range of NumPyro distributions --- multivariate normal, truncated, mixture, or any user-defined distribution --- and these priors are JIT-compatible, supporting both \texttt{sample} and \texttt{log\_prob} within compiled pipelines. The latter is required for exact likelihood computation via the continuous change-of-variables formula (Section~\ref{sec:solvers}). Model checkpointing is handled by \texttt{Orbax}~\cite{orbax2024}, which provides standardised serialisation of model parameters, optimiser state, and EMA weights.

Multi-device model parallelism --- sharding parameters across multiple accelerators --- is not yet implemented, but it is planned for a future GenSBI release. JAX's sharding primitives and \texttt{Flax nnx}'s SPMD support make this a natural extension of the existing architecture, requiring no changes to users' model or training code.

\section{Benchmark Results}
\label{sec:benchmarks}

The previous section described GenSBI's software architecture, neural network models, solvers, and built-in diagnostics. We now validate the library empirically on a suite of inference tasks drawn from the simulation-based inference literature. The goals of this section are threefold: first, to demonstrate that GenSBI produces well-calibrated posteriors across a range of problem dimensionalities and posterior geometries; second, to systematically compare the library's architectures (Flux1, Flux1Joint), generative methods (flow matching, score matching, denoising diffusion (EDM)), and simulation budgets, placing the results in the context of existing SBI methods; and third, to illustrate the library's applicability to scientific problems where the observations are structured data --- time series and images --- rather than low-dimensional summary statistics.

Section~\ref{sec:benchmark_tasks} introduces the seven benchmark tasks. Section~\ref{sec:experimental_setup} describes the experimental setup, including the choice of architecture, generative method, and training configuration. Section~\ref{sec:posterior_quality} presents the posterior quality results in three stages: we first establish ceiling performance with a generous simulation budget, then systematically compare all model variants as a function of simulation budget, and finally benchmark GenSBI against existing methods from the literature. Section~\ref{sec:calibration} reports the calibration diagnostics. Finally, Section~\ref{sec:advanced_applications} demonstrates GenSBI on two scientific applications --- gravitational wave parameter estimation and strong gravitational lensing --- where the observations are high-dimensional structured data and no reference posterior is available.

\subsection{Benchmark Tasks}
\label{sec:benchmark_tasks}

We evaluate GenSBI on seven inference tasks spanning a range of parameter dimensionalities, observation structures, and posterior geometries. Five tasks are drawn from the Simulation-Based Inference Benchmark (SBIBM)~\cite{Lueckmann:2021sbibm}, which provides analytically computed or MCMC-derived reference posteriors for quantitative evaluation. Two additional tasks --- gravitational wave inference and strong gravitational lensing --- involve high-dimensional, structured observations for which no reference posterior is available; these serve as demonstrations of GenSBI's applicability to realistic scientific problems. The datasets used to train and evaluate all benchmark tasks are publicly available on HuggingFace\footnote{\url{https://huggingface.co/datasets/aurelio-amerio/SBI-benchmarks}}, and the \texttt{gensbi-examples} subpackage~\cite{GenSBI_examples} provides utilities for downloading and preprocessing them. Table~\ref{tab:benchmark_tasks} summarises the task specifications.

\begin{table}[t]
\centering
\caption{Summary of benchmark tasks. The SBIBM tasks provide reference posteriors for quantitative evaluation via the C2ST metric. The advanced tasks lack reference posteriors and are validated through calibration diagnostics alone.}
\label{tab:benchmark_tasks}
\small
\begin{tabular}{@{}llccl@{}}
\toprule
\textbf{Task} & \textbf{Source} & $\dim(\theta)$ & $\dim(x)$ & \textbf{Reference posterior} \\
\midrule
Two Moons & SBIBM & 2 & 2 & \checkmark \\
Gaussian Linear & SBIBM & 10 & 10 & \checkmark \\
Gaussian Mixture & SBIBM & 2 & 2 & \checkmark \\
SLCP & SBIBM & 5 & 8 & \checkmark \\
Bernoulli GLM & SBIBM & 10 & 10 & \checkmark \\
\midrule
Gravitational Waves & \cite{Hermans:2019ioj} & 2 & $2 \times 8192$ & --- \\
Strong Lensing & \cite{Miller2022} & 2 & $64 \times 64$ & --- \\
\bottomrule
\end{tabular}
\end{table}

\paragraph{Two Moons.}
The Two Moons task~\cite{Lueckmann:2021sbibm} maps a two-dimensional parameter $\theta \in \mathbb{R}^2$ to a two-dimensional observation $x \in \mathbb{R}^2$ through a simulator that generates samples along two crescent-shaped regions. The posterior is strongly non-Gaussian, exhibiting both global bimodality and complex local curvature within each mode. This makes it a standard test for whether a density estimator can capture multimodal target distributions with distinct geometric structure.

\paragraph{Gaussian Linear.}
The Gaussian Linear task~\cite{Lueckmann:2021sbibm} is a ten-dimensional linear model where the parameter vector $\theta \in \mathbb{R}^{10}$ serves as the mean of a multivariate Gaussian likelihood with a fixed covariance matrix. The posterior is analytically Gaussian, making this task a sanity check for correctness in moderate dimensionality. It verifies that the density estimator can handle the scaling to higher dimensions without introducing systematic biases.

\paragraph{Gaussian Mixture.}
The Gaussian Mixture task~\cite{Lueckmann:2021sbibm} generates two-dimensional observations from a mixture of two Gaussians that share a common mean $\theta \in \mathbb{R}^2$ but have different covariance scales --- one broader ($I$) and one narrower ($0.1\,I$). This configuration creates a posterior with multiple spatial length scales, testing the ability of the density estimator to resolve both coarse and fine structure simultaneously. The task is a classical benchmark from the Approximate Bayesian Computation (ABC) literature.

\paragraph{SLCP (Simple Likelihood Complex Posterior).}
The SLCP task~\cite{Lueckmann:2021sbibm} generates eight-dimensional observations from a Gaussian likelihood whose mean and covariance are highly nonlinear functions of the five-dimensional parameter vector $\theta \in \mathbb{R}^5$. Despite the simple likelihood form, the resulting posterior is intricate, featuring four symmetrical modes and sharp, nonlinear boundaries. SLCP is widely regarded as one of the most challenging SBIBM tasks and tests the expressive capacity of the density estimator under complex posterior geometry.

\paragraph{Bernoulli GLM.}
The Bernoulli GLM task~\cite{Lueckmann:2021sbibm} is a ten-dimensional generalised linear model that produces discrete Bernoulli observations. The prior is a multivariate Gaussian with a structured covariance matrix that penalises second-order differences in the parameter vector, enforcing smooth, correlated relationships between dimensions. We use the sufficient-statistics variant ($x \in \mathbb{R}^{10}$), which compresses the raw $100$-dimensional binary observation sequence into its sufficient summary. This task tests inference with non-factorizable priors and discrete data generation processes.

\paragraph{Gravitational waves.}
The gravitational wave task follows the setup of Hermans et al.~\cite{Hermans:2019ioj}; it simulates the merger of two black holes using a simplified version of the \texttt{pycbc} waveform generator~\cite{alex_nitz_2024_pycbc}. The goal is to infer the two component masses $\theta = (m_1, m_2) \in \mathbb{R}^2$ from the observed strain time series recorded by two detectors (LIGO Hanford and Livingston), yielding an observation of shape $x \in \mathbb{R}^{2 \times 8192}$. The high dimensionality of the observation ($2 \times 8192$ time-domain samples) necessitates the use of a learned embedding network --- specifically, a 1D convolutional autoencoder --- to compress the raw strain data into a compact latent representation before passing it to the inference model (Section~\ref{sec:nn_architectures}). No reference posterior is available for this task; validation relies on calibration diagnostics. We note that this is a deliberately simplified setup: only the two component masses are varied, whereas a full compact-binary coalescence analysis involves 15 or more parameters~\cite{Christensen:2022bxb}. The task is therefore intended as a proof-of-concept demonstration of GenSBI's ability to handle high-dimensional time-series conditioning, not as a competitive gravitational wave analysis pipeline.

\paragraph{Strong gravitational lensing.}
The strong lensing task is a toy gravitational lensing problem modelled after the image analysis tutorial in the \textsc{swyft} documentation~\cite{Miller2022}. A simple simulator generates $64 \times 64$ pixel images of lensing ring systems parametrised by the ring position and radius, $\theta \in \mathbb{R}^2$, and the goal is to infer these parameters from the image. As with the gravitational wave task, a learned embedding --- here a 2D convolutional autoencoder --- compresses the image into a latent space for the transformer-based inference model; no reference posterior is available. Unlike realistic strong lensing, where a background galaxy is distorted by the gravitational potential of a foreground mass into arcs or partial rings whose morphology depends on the source-lens alignment~\cite{Treu:2010uj}, the simulator generates idealised ring images with additive Gaussian (unstructured) noise and random superimposed lines (structured noise). The task therefore serves as a controlled testbed for image-conditioned posterior estimation rather than a realistic lensing analysis.

\subsection{Experimental Setup}
\label{sec:experimental_setup}

The choice of model architecture for each task class reflects the structural properties of the inference problem. The five SBIBM tasks are unstructured inference problems: the parameters $\theta$ and observations $x$ are low-dimensional vectors with no inherent conditional asymmetry, so both the joint distribution $p(\theta, x)$ and the conditional $p(\theta \mid x)$ are equally natural modelling targets. For these tasks, we train both the \textbf{Flux1} architecture (conditional density estimation via the \texttt{ConditionalPipeline}) and the \textbf{Flux1Joint} architecture (joint density estimation via the \texttt{JointPipeline}; see Section~\ref{sec:architecture}), using all three generative methods implemented in GenSBI: flow matching, score matching, and EDM. This yields a systematic comparison across two architectures and three generative formulations.

The two advanced tasks --- gravitational waves and strong lensing --- have a different character. The observations are high-dimensional, structured data (time series and images, respectively) that are naturally given as conditioning inputs, and the parameters to be inferred are low-dimensional physical quantities. These problems present themselves as conditional density estimation tasks, and joint modelling over the raw observation space would be neither practical nor meaningful. For these tasks, we use the \textbf{Flux1} architecture with \textbf{flow matching}, training via the \texttt{ConditionalPipeline}. Both tasks employ a learned embedding network to compress the high-dimensional observations into compact latent representations: a 1D convolutional encoder for the gravitational wave strain data and a 2D convolutional encoder for the lensing images. The encoder is trained jointly with the transformer-based density estimator in an end-to-end fashion, allowing the learned representation to adapt to the requirements of the downstream inference task. The two tasks also illustrate the flexibility of GenSBI's ID embedding system (Section~\ref{sec:nn_architectures}): the gravitational wave encoder produces a one-dimensional latent sequence that is paired with 1D sinusoidal positional encodings, while the lensing encoder outputs a spatially structured latent map that is patchified into tokens and paired with 2D RoPE embeddings, preserving spatial correlations between encoded patches in a conceptually similar way to latent diffusion models for image generation~\cite{Rombach:2022ldm}.

\paragraph{Simulation budgets.}
The SBIBM tasks are evaluated at two budget levels, each serving a distinct purpose. First, we train a set of models with a generous budget of $10^6$ parameter--observation pairs to verify that GenSBI can achieve reasonable posterior recovery when data is abundant; for these models we also compute full calibration diagnostics (Section~\ref{sec:calibration}). Second, we perform a systematic scan across three simulation budgets --- $10^4$, $3 \times 10^4$, and $10^5$ --- for the full battery of model variants (2 architectures $\times$ 3 generative methods). The upper limit of $10^5$ matches the budget range used in the SimFormer benchmark~\cite{Gloeckler:2024simformer}, enabling direct comparison with previous studies; training at larger budgets would preclude such comparisons and provide little additional insight. For the advanced tasks, we use a budget of $10^5$ simulations, reflecting a scenario where the observations are more complex (high-dimensional time series and images) and where larger simulation budgets may not always be available in practice.

\paragraph{Training details.}
All models are trained with the AdamW optimiser using a cosine learning rate schedule with linear warmup, and we maintain an exponential moving average (EMA) of the model parameters throughout training. Early stopping based on the validation loss ratio is enabled for all experiments.
We note that flow and diffusion generative models adopting transformers as their backbone greatly benefit from longer training schedules, even when the training loss has apparently converged~\cite{Peebles:2023dit, Geng:2025meanflow}. For this reason, we train our models for $(5\text{--}10) \times 10^5$ steps by default, unless the validation loss diverges --- which typically signals an excessively high learning rate or insufficient model capacity rather than over-training.

The complete training and architecture configurations for every combination of task, model, generative method, and simulation budget are reported in Appendix~\ref{app:training_configs} and the \texttt{GenSBI-examples} repository~\cite{GenSBI_examples}. All experiments were run on NVIDIA Tesla V100/A100 GPUs on the Artemisa computing cluster at IFIC (University of Valencia)\footnote{\url{https://artemisa.ific.uv.es}}.

\paragraph{Computational cost.}
Table~\ref{tab:computational_benchmarks} reports wall-clock training speed, training time, and posterior sampling time for all six model variants on the Two Moons task, measured on a single NVIDIA Tesla V100 GPU. 
All models were trained for $50{,}000$ steps with a batch size of 256. Within a fixed architecture, the three generative methods achieve essentially identical training throughput (e.g.\ $\sim$4.5--4.7\,it/s for Flux1, $\sim$11.5\,it/s for Flux1Joint\footnote{The absolute speed difference between the two architectures in this table is an artifact of our specific configurations (e.g., our \texttt{Flux1Joint} setup has fewer layers); the relevant comparison here is the relative consistency of training speeds across generative methods within a given architecture.}), 
confirming that the choice of generative formulation has no impact on training cost. Sampling speed, in contrast, varies substantially across methods because each formulation requires a different number of solver steps to produce high-quality samples: EDM uses an 18-step second-order Heun solver, flow matching uses a 100-step ODE solver, and score matching requires $\sim$$1{,}000$ SDE solver steps, following the recommendations of the respective original papers~\cite{Karras:2022edm, Lipman:2023fm, Song:2020sde}. Despite these differences, drawing $10{,}000$ posterior samples takes at most $\sim$42\,s in the slowest configuration. We further note that all benchmark models in this section can also be trained on a consumer-grade NVIDIA RTX 4070 GPU (12\,GB VRAM) with a batch size of 256, demonstrating that GenSBI does not require high-end cluster hardware for practical use.

\begin{table}[t]
\centering
\caption{Computational benchmarks for the Two Moons task on an NVIDIA Tesla V100 GPU. All models were trained with a batch size of 256 for $50{,}000$ steps. Sampling time reports the wall-clock time to draw $10{,}000$ posterior samples using each method's default solver and recommended step count.}
\label{tab:computational_benchmarks}
\small
\begin{tabular}{@{}llcccc@{}}
\toprule
\textbf{Architecture} & \textbf{Method} & \textbf{Training speed} & \textbf{Training time} & \textbf{Solver steps} & \textbf{Sampling time} \\
 & & (it/s) & (50k steps) & & ($10^4$ samples) \\
\midrule
\multirow{3}{*}{Flux1}
 & Flow Matching   & 4.58  & $\sim$3.0\,h & 100  & 6.9\,s \\
 & Score Matching  & 4.50  & $\sim$3.1\,h & 1000 & 24.1\,s \\
 & EDM             & 4.73  & $\sim$2.9\,h & 18   & 8.3\,s \\
\midrule
\multirow{3}{*}{Flux1Joint}
 & Flow Matching   & 11.56 & $\sim$1.2\,h & 100  & 6.4\,s \\
 & Score Matching  & 11.63 & $\sim$1.2\,h & 1000 & 42.3\,s \\
 & EDM             & 11.51 & $\sim$1.2\,h & 18   & 5.3\,s \\
\bottomrule
\end{tabular}
\end{table}

\subsection{Posterior Quality}
\label{sec:posterior_quality}

We assess the quality of the learned posteriors through visual inspection and the classifier two-sample test (C2ST)~\cite{Lopez-Paz:2017c2st}. A C2ST score of $0.5$ indicates that a binary classifier cannot distinguish samples from the learned posterior and the reference posterior --- corresponding to a perfect match --- while a score approaching $1.0$ indicates a large discrepancy. The SBIBM framework provides, for each task, 10 test observations together with pre-computed reference posterior samples obtained via analytical calculation or MCMC runs~\cite{Lueckmann:2021sbibm}. For each of these observations, we generate posterior samples from the trained model and compute the C2ST against the corresponding reference posterior.

\begin{figure}[t]
    \centering
    \includegraphics[width=0.7\columnwidth]{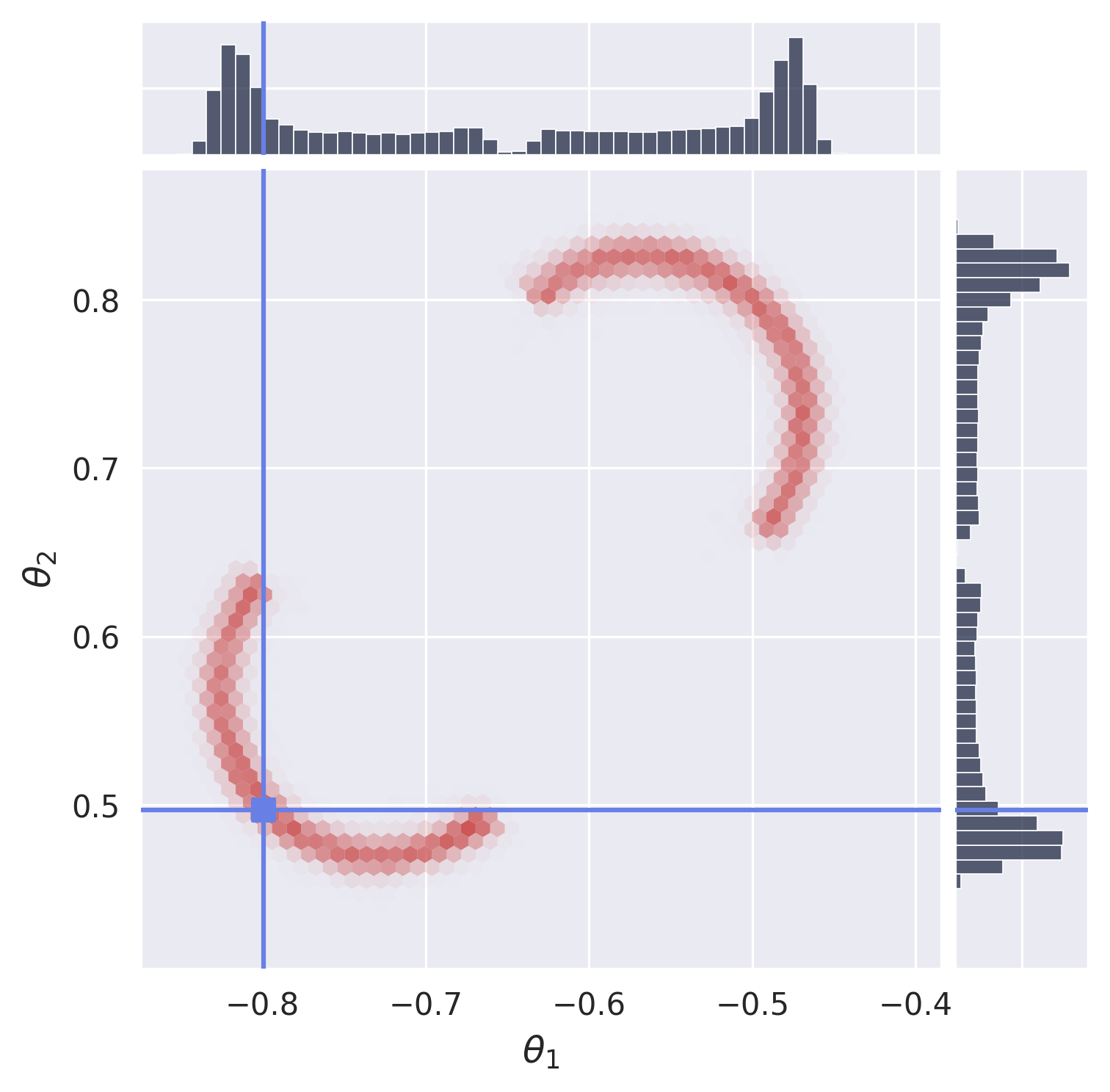}
    \caption{Marginal posterior distributions for the Two Moons task (Flux1Joint, flow matching, EMA). The bimodal crescent structure is well resolved. Marginal plots for the remaining SBIBM tasks are shown in Appendix~\ref{app:sbibm_results}.}
    \label{fig:marginals_two_moons}
\end{figure}

Figure~\ref{fig:marginals_two_moons} shows the marginal posterior distributions for the Two Moons task as a representative example; the learned posterior accurately resolves both crescent-shaped modes and their local curvature. Marginal posterior plots for the remaining four SBIBM tasks are reported in Appendix~\ref{app:sbibm_results} and show similarly close agreement with the reference posteriors across all tasks.

\paragraph{Ceiling performance.}
We first verify that, given sufficient training data, GenSBI can achieve reasonable posterior recovery across all benchmark tasks. Table~\ref{tab:c2st_summary} reports the C2ST accuracy, averaged over 10 test observations, for each SBIBM task trained with a generous simulation budget of $10^6$ parameter--observation pairs using Flux1Joint with flow matching. Both Gaussian tasks achieve scores very close to $0.5$ ($\lesssim 0.51$), confirming correct posterior recovery on these baseline problems. Two Moons achieves $0.504 \pm 0.010$, indicating accurate capture of the bimodal structure. Bernoulli GLM scores $0.557 \pm 0.020$, reflecting the moderate difficulty of inferring the correlated 10-dimensional posterior. SLCP is the most challenging task, and Flux1Joint achieves $0.549 \pm 0.019$ --- a substantial improvement over the standard Flux1 architecture ($0.689 \pm 0.047$). This result is consistent with the findings of Gloeckler et al.~\cite{Gloeckler:2024simformer}, who observe that training on the full joint distribution is particularly beneficial for likelihood-dominated tasks --- problems where the likelihood is simple but the posterior exhibits complex, multimodal geometry. By learning the joint $p(\theta, x)$ rather than the conditional $p(\theta \mid x)$ alone, the model captures the structure of both the likelihood and the prior, which facilitates more accurate posterior recovery on tasks like SLCP. These $10^6$-budget models also serve as the basis for the calibration analysis in Section~\ref{sec:calibration}.

\begin{table}[t]
\centering
\caption{C2ST accuracy (mean $\pm$ std.\ over 10 test observations) for each SBIBM task using Flux1Joint with flow matching, EMA parameters, and a simulation budget of $10^6$. These results demonstrate that, with a generous training budget, the model achieves reasonable posterior recovery across all tasks, including those with complex multimodal geometry.}
\label{tab:c2st_summary}
\small
\begin{tabular}{@{}lc@{}}
\toprule
\textbf{Task} & \textbf{C2ST (Flux1Joint)} \\
\midrule
Gaussian Linear & $0.507 \pm 0.004$ \\
Gaussian Mixture & $0.501 \pm 0.004$ \\
Two Moons & $0.504 \pm 0.010$ \\
SLCP & $0.549 \pm 0.019$ \\
Bernoulli GLM & $0.557 \pm 0.020$ \\
\bottomrule
\end{tabular}
\end{table}

\begin{figure}[t]
    \centering
    \includegraphics[width=\columnwidth]{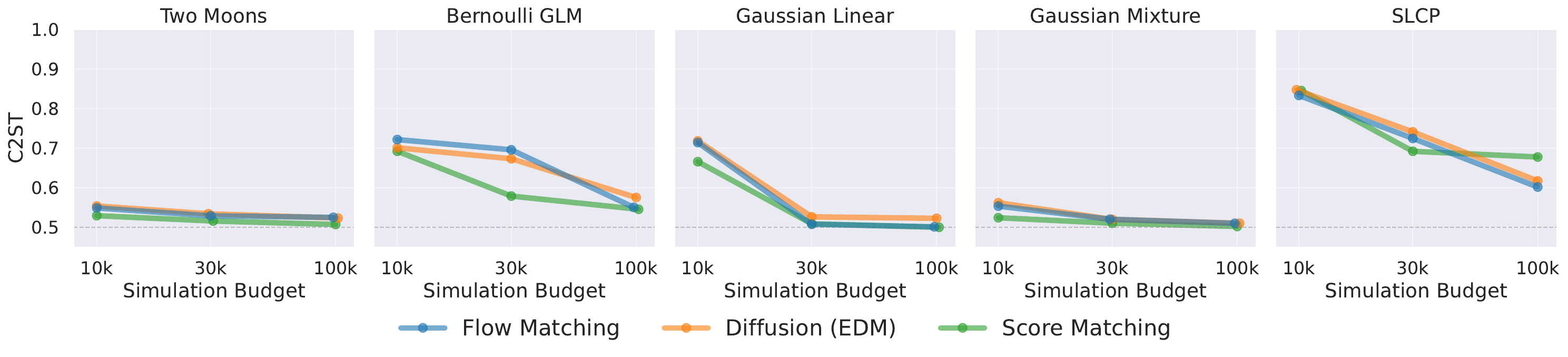}
    \caption{Best C2ST accuracy as a function of simulation budget for the Flux1 architecture across five SBIBM tasks and three generative methods (flow matching, score matching, EDM). Lower is better; $0.5$ indicates a perfect posterior match.}
    \label{fig:c2st_budget_flux1}
\end{figure}

\begin{figure}[t]
    \centering
    \includegraphics[width=\columnwidth]{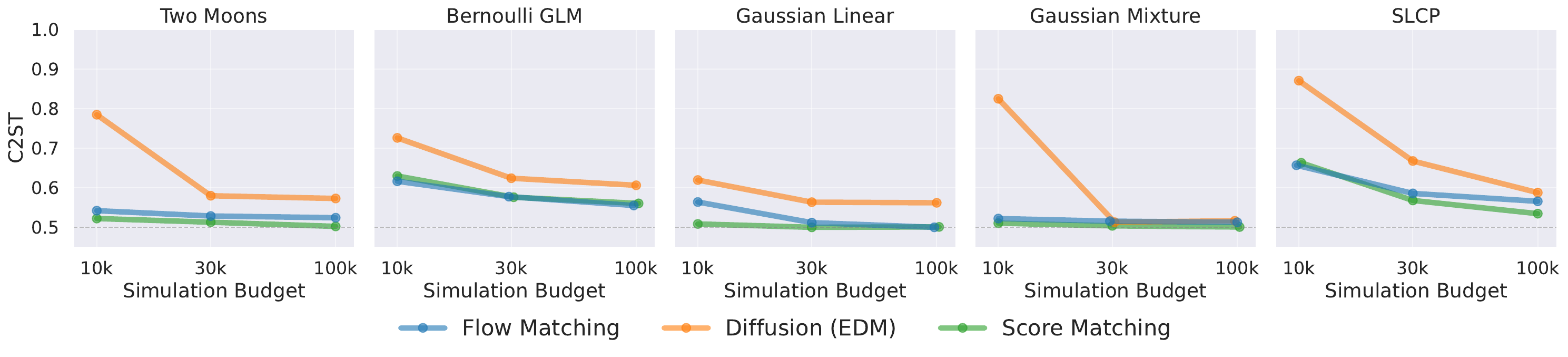}
    \caption{Best C2ST accuracy as a function of simulation budget for the Flux1Joint architecture across five SBIBM tasks and three generative methods. Flux1Joint achieves stronger performance than Flux1 (Figure~\ref{fig:c2st_budget_flux1}), particularly on the SLCP task.}
    \label{fig:c2st_budget_flux1joint}
\end{figure}

\paragraph{Method and architecture comparison.}
Having established that GenSBI can achieve reasonable performance given sufficient data, we now systematically compare all model variants across simulation budgets. Figures~\ref{fig:c2st_budget_flux1} and~\ref{fig:c2st_budget_flux1joint} show the best C2ST score achieved at each budget level ($10^4$, $3 \times 10^4$, $10^5$) for flow matching, score matching, and EDM, for the Flux1 and Flux1Joint architectures respectively.

Several trends emerge. All three generative methods converge towards similar C2ST scores as the simulation budget increases, empirically validating the interchangeability of the three formulations implemented in GenSBI. On simpler tasks (Two Moons, Gaussian Mixture, Gaussian Linear), all methods achieve near-optimal scores ($\lesssim 0.52$) already at $3 \times 10^4$ samples. On harder tasks (SLCP, Bernoulli GLM), flow matching and score matching tend to reach good performance faster than the EDM formulation, which often requires a larger simulation budget to converge to the same C2ST score. Comparing the two architectures, for tasks with unstructured data, the best Flux1Joint models consistently achieve better or comparable C2ST scores relative to Flux1, with the largest gains on SLCP. 
This suggests that some tasks may benefit from training a joint density estimation model, even if the end goal is conditional density estimation. Furthermore, this success demonstrates that the joint model can accurately recover complex, multi-modal posterior geometries despite the broader capacity demands of learning the full joint distribution.
The EDM formulation presents a notable exception to this architectural trend: given the same number of samples, Flux1 tends to perform better than Flux1Joint when trained with EDM. This is likely a consequence of EDM's slower scaling with sample size coupled with the fact that learning the joint distribution $p(\theta, x)$ is inherently more challenging than learning the conditional $p(\theta \mid x)$.
At the smallest training budget of $10^4$ simulations, this weakness becomes especially pronounced for Flux1Joint, with C2ST scores reaching $0.825$ on Gaussian Mixture and $0.871$ on SLCP -- values well above the $\sim 0.5$ optimum and substantially worse than those obtained by FM or SM under identical conditions. We hypothesise that the preconditioning and noise schedule proposed by~\cite{Karras:2022edm}, originally tuned for high-dimensional image distributions where loss mass concentrates at intermediate noise levels $\sigma$, may not transfer optimally to the low-dimensional joint densities encountered in SBI. A more detailed investigation of alternative schedules, loss weightings, and step counts is deferred to Section~\ref{sec:limitations}.
The only case where Flux1Joint maintains an edge over Flux1 within the EDM framework is the SLCP dataset, where joint modelling provides a substantial overall advantage. The full numerical values are reported in Appendix~\ref{app:c2st_budget_tables}.

\begin{figure}[t]
    \centering
    \includegraphics[width=\columnwidth]{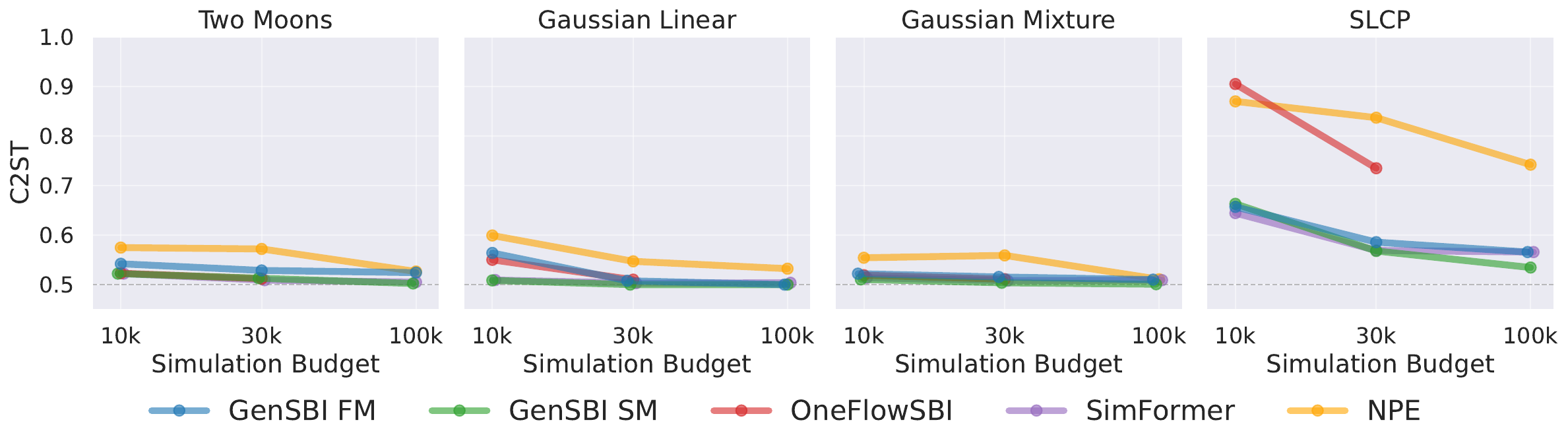}
    \caption{C2ST accuracy as a function of simulation budget, comparing the best GenSBI model (flow matching and score matching with Flux1/Flux1Joint) against three baselines from the literature: OneFlowSBI~\cite{Nautiyal:2026oneflowsbi}, SimFormer~\cite{Gloeckler:2024simformer}, and NPE~\cite{BoeltsDeistler_sbi_2025}. Lower is better; $0.5$ indicates a perfect posterior match. OneFlowSBI results are available only at $10^4$ and $3 \times 10^4$ budgets. GenSBI achieves competitive or state-of-the-art performance across all tasks, matching or surpassing all baselines at $10^5$ simulations.}
    \label{fig:c2st_comparison}
\end{figure}

\paragraph{Comparison with the literature.}
To place GenSBI's performance in the context of existing methods, Figure~\ref{fig:c2st_comparison} compares the best C2ST scores achieved by GenSBI (selecting the best-performing combination of generative method and architecture at each budget) against three baselines from the literature: OneFlowSBI~\cite{Nautiyal:2026oneflowsbi}, SimFormer~\cite{Gloeckler:2024simformer}, and NPE as implemented in the \texttt{sbi} library~\cite{BoeltsDeistler_sbi_2025}. The baseline C2ST values are taken directly from the respective publications; in particular, OneFlowSBI results are available only for budgets of $10^4$ and $3 \times 10^4$, as no $10^5$-budget data is reported in their paper. A full per-model breakdown at the $3 \times 10^4$ budget is provided in Appendix~\ref{app:c2st_comparison_table}.

GenSBI achieves competitive performance across all tasks and simulation budgets. At the largest budget ($10^5$ simulations), GenSBI matches or surpasses all baselines on every task for which comparison data is available: it achieves C2ST scores very close to $0.5$ on Gaussian Linear ($0.500$) and Gaussian Mixture ($0.501$), closely matches SimFormer on Two Moons ($0.502$ vs.\ $0.505$), and outperforms all competitors on SLCP ($0.534$ vs.\ $0.566$ for SimFormer and $0.742$ for NPE). These results consolidate flow matching and score matching as effective generative frameworks for simulation-based inference and demonstrate that modern transformer architectures --- originally developed for image generation --- can be successfully adapted for posterior density estimation, achieving competitive or state-of-the-art performance. Notably, GenSBI obtains these results with a nearly uniform training configuration across all tasks (Appendix~\ref{app:training_configs}), in contrast to approaches such as OneFlowSBI, which require extensive per-task hyperparameter optimisation --- varying the model architecture, learning rate, batch size, and time-sampling schedule across benchmark problems~\cite{Nautiyal:2026oneflowsbi}.

\subsection{Calibration Diagnostics}
\label{sec:calibration}

The C2ST metric provides a useful quantitative summary of posterior quality: a score near $0.5$ confirms that samples from the learned posterior are statistically indistinguishable from the reference. However, C2ST is a scalar diagnostic --- when it deviates from $0.5$, it signals that a discrepancy exists but reveals nothing about its nature. It cannot distinguish, for example, a posterior that is over-dispersed from one that is under-dispersed, shifted, or misshapen. To move beyond detection and toward diagnosis, we complement the C2ST analysis with the TARP diagnostic (Section~\ref{sec:diagnostics}), which checks whether the posterior's credible regions achieve the expected frequentist coverage across the prior predictive distribution. We compute full calibration diagnostics for the $10^6$-budget models presented in the previous section; for the systematic budget scan, the C2ST metric alone provides a sufficient and well-understood measure of posterior quality for comparing model variants.

\begin{figure}[t]
    \centering
    \includegraphics[width=0.9\columnwidth]{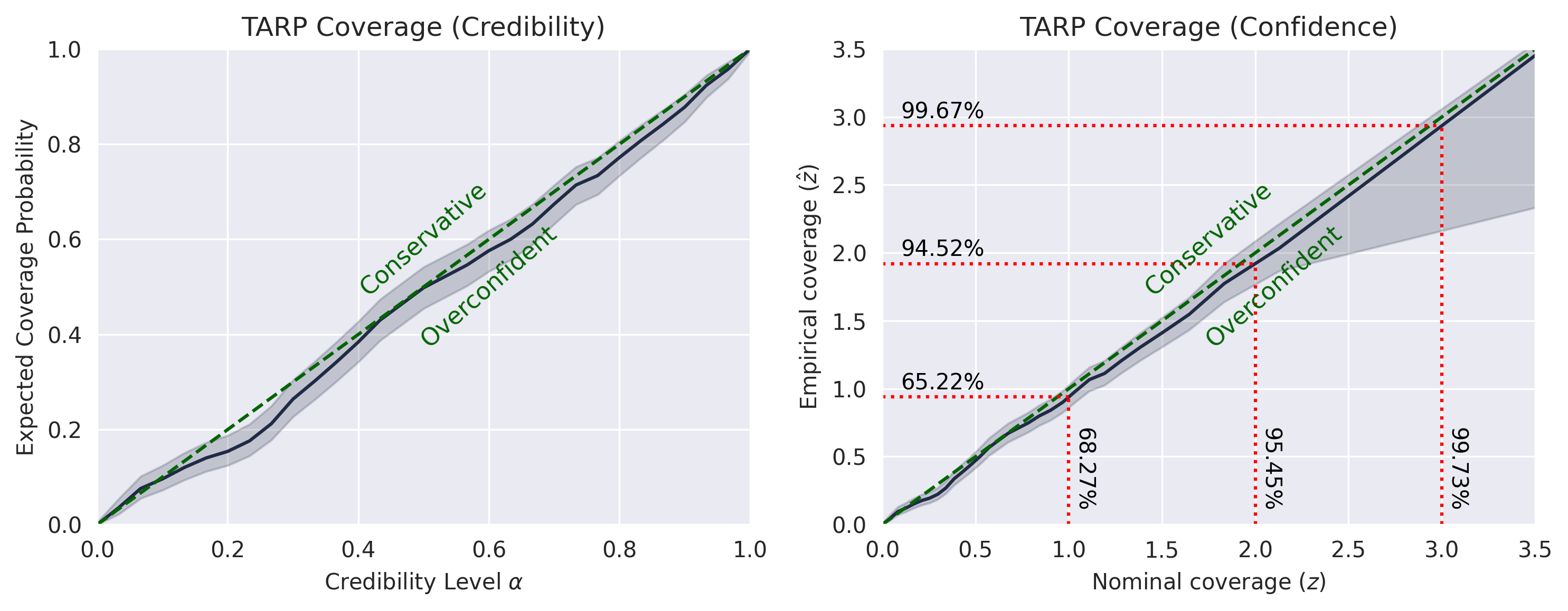}
    \caption{TARP expected coverage probability curve for the Two Moons task (Flux1Joint, flow matching, EMA). The curve lies on the diagonal (dashed line), indicating well-calibrated posteriors. Shaded band shows the Jeffreys 95\% confidence interval. TARP plots for the remaining SBIBM tasks are shown in Appendix~\ref{app:sbibm_results}.}
    \label{fig:tarp_two_moons}
\end{figure}

A well-calibrated posterior produces an ECP curve that lies on the diagonal; curves above the diagonal indicate conservative (over-covering) posteriors, and curves below indicate overconfident (under-covering) posteriors. Figure~\ref{fig:tarp_two_moons} shows the TARP coverage curve for the Two Moons task as a representative SBIBM result: the curve falls on the diagonal and within the Jeffreys confidence intervals, confirming well-calibrated posterior estimation. The TARP curves for the remaining four SBIBM tasks, reported in Appendix~\ref{app:sbibm_results}, show equally good calibration across all problem dimensionalities and posterior geometries.

For brevity, we report only the TARP diagnostic in this section. The \texttt{GenSBI-examples} repository~\cite{GenSBI_examples} provides a complete diagnostic suite for all tasks and model variants, including SBC rank histograms, LC2ST diagnostics, and marginal coverage plots, offering a full picture of model performance beyond what can be presented here.

\subsection{Advanced Applications}
\label{sec:advanced_applications}

The SBIBM tasks provide controlled, quantitative evaluation of posterior quality using reference posteriors, but they involve low-dimensional observations and relatively simple data generation processes. To demonstrate that GenSBI can handle inference problems with high-dimensional, structured observations, we present two additional applications: parameter estimation from simulated gravitational wave signals and parameter recovery from simulated strong gravitational lensing images. These tasks feature more complex observation structures --- time series and images, respectively --- representative of scenarios where the simulation cost or complexity may limit the available training budget. Both tasks are trained with $10^5$ simulations, both require learned embedding networks to compress the raw observations into compact latent representations, and neither has a tractable reference posterior --- validation relies entirely on calibration diagnostics.

\begin{figure}[t]
    \centering
    \resizebox{0.99\textwidth}{!}{%
    \hspace{-3mm}
    \raisebox{-0.5\height}{\includegraphics[width=0.40\linewidth]{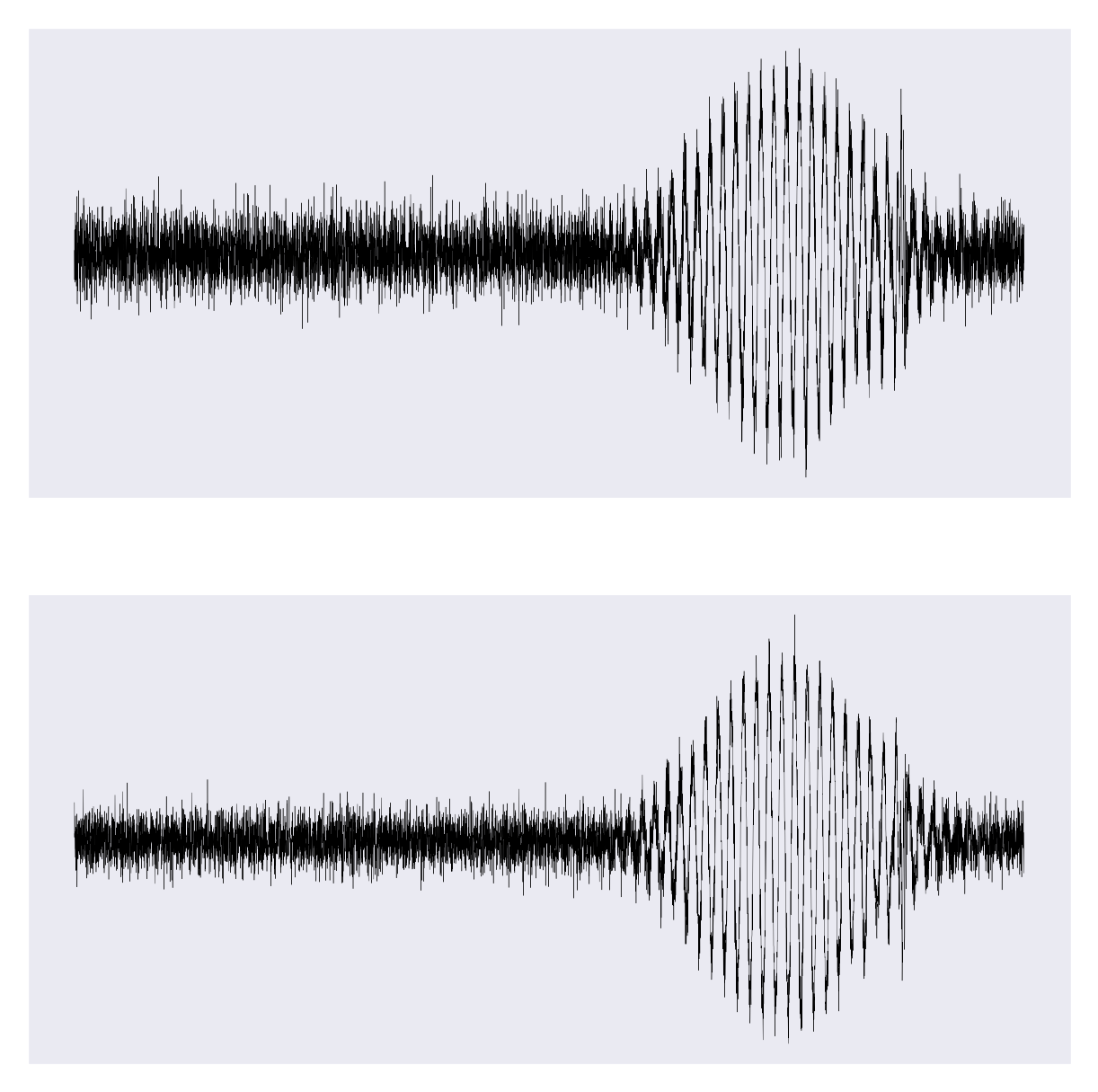}}
    \hspace{-3mm}
    \raisebox{-0.5\height}{\includegraphics[width=0.60\linewidth]{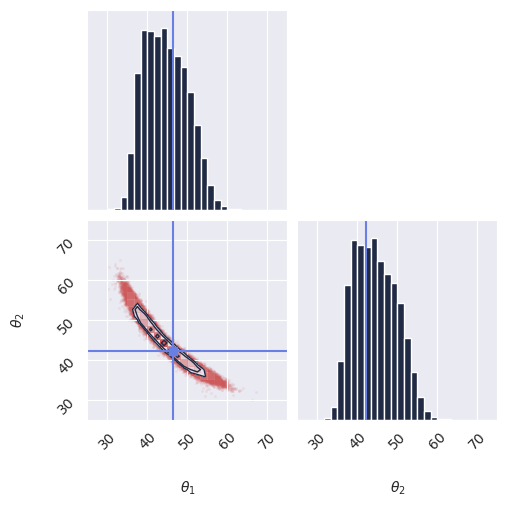}}
    }
    \caption{Gravitational wave parameter estimation task (Flux1, flow matching, EMA).
    \panel{Left}: example simulated strain time series from the two LIGO detectors (Hanford and Livingston).
    \panel{Right}: posterior marginal distributions for the component masses $(m_1, m_2)$ of the merging binary; the true parameter values are indicated by the crosshairs.}
    \label{fig:gw_results}
\end{figure}

\begin{figure}[t]
    \centering
    \includegraphics[width=0.9\columnwidth]{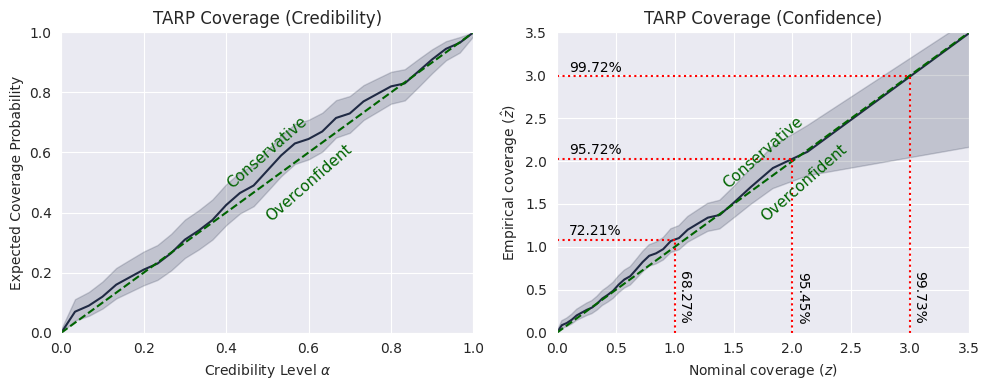}
    \caption{TARP expected coverage probability curve for the gravitational wave task (Flux1, flow matching, EMA). The curve lies close to the diagonal, confirming well-calibrated posteriors.}
    \label{fig:tarp_gw}
\end{figure}

\paragraph{Gravitational wave parameter estimation.}
Figure~\ref{fig:gw_results} shows the posterior for the component masses of a simulated binary black hole merger, inferred from strain time-series data recorded by two LIGO detectors. The Flux1 model, conditioned on the latent representation produced by a 1D convolutional encoder, recovers a posterior that correctly contains the true parameter values and exhibits the expected correlation structure between the two masses. The TARP diagnostic (Figure~\ref{fig:tarp_gw}) confirms that the posterior is well calibrated across the prior predictive distribution.

\begin{figure}[t]
    \centering
    \resizebox{0.99\textwidth}{!}{%
    \hspace{-3mm}
    \raisebox{-0.5\height}{\includegraphics[width=0.40\linewidth]{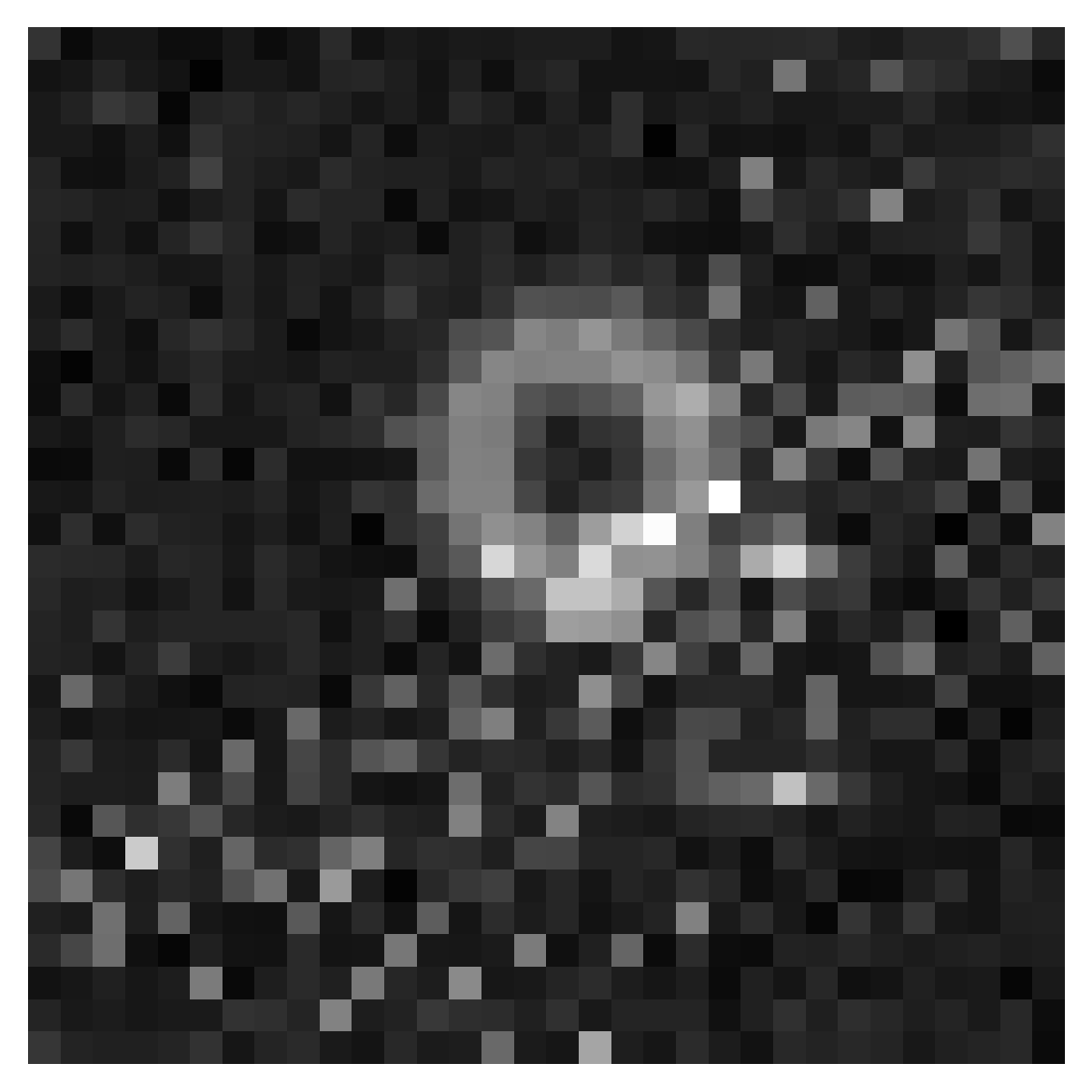}}
    \hspace{-3mm}
    \raisebox{-0.5\height}{\includegraphics[width=0.60\linewidth]{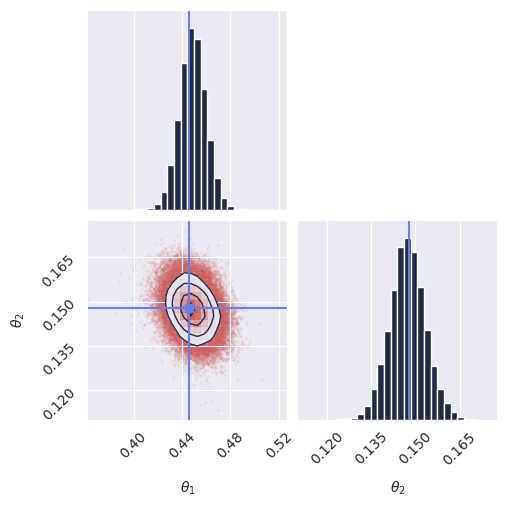}}
    }
    \caption{Strong gravitational lensing task (Flux1, flow matching, EMA).
    \panel{Left}: example simulated $64 \times 64$ lensing image.
    \panel{Right}: posterior marginal distributions for the ring position and radius; the true parameter values are indicated by the crosshairs.}
    \label{fig:lensing_results}
\end{figure}

\begin{figure}[t]
    \centering
    \includegraphics[width=0.9\columnwidth]{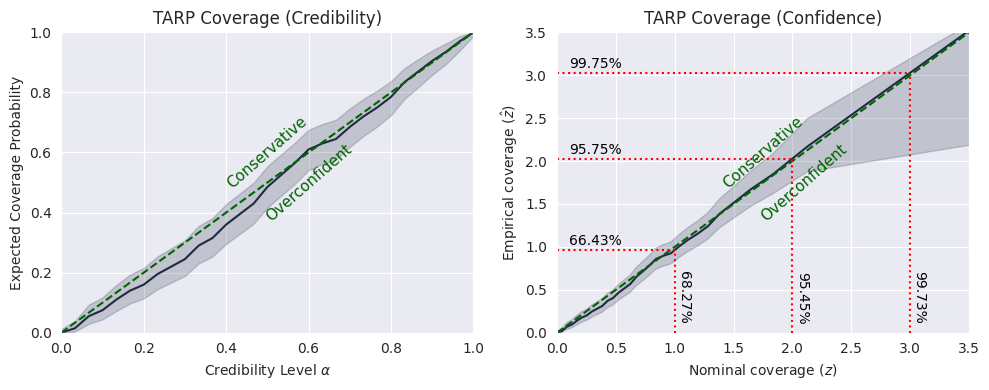}
    \caption{TARP expected coverage probability curve for the strong lensing task (Flux1, flow matching, EMA). The curve lies close to the diagonal, confirming well-calibrated posteriors.}
    \label{fig:tarp_lensing}
\end{figure}

\paragraph{Strong gravitational lensing.}
Figure~\ref{fig:lensing_results} shows the posterior for the ring position and radius inferred from a simulated $64 \times 64$ image of a toy gravitational lensing system (Section~\ref{sec:benchmark_tasks}). The 2D convolutional encoder compresses the image into a lower-dimensional latent map that retains spatial structure; this map is then patchified into tokens and processed by the Flux1 conditioning stream with 2D RoPE embeddings (Section~\ref{sec:nn_architectures}), so that the transformer can exploit spatial correlations between neighbouring patches. The resulting posteriors are consistent with the true parameter values, and the TARP diagnostic (Figure~\ref{fig:tarp_lensing}) shows good calibration, confirming that the model generalises to image-conditioned inference tasks. Together with the gravitational wave example, this demonstrates that GenSBI's architecture --- particularly the separation between the embedding network and the transformer-based density estimator, combined with the flexible ID embedding system --- supports a unified workflow for applying neural posterior estimation to diverse scientific data types.

\section{Related Software}
\label{sec:related}

The SBI software landscape is dominated by PyTorch-based toolkits, reflecting PyTorch's maturity and broad adoption in the machine learning community. GenSBI complements this ecosystem by bringing flow matching and diffusion-based NPE to JAX, where a growing number of scientific workflows --- differentiable simulators, probabilistic programming, and hardware-accelerated sampling --- already run natively.

\subsection{Established SBI toolkits}
\label{sec:established_toolkits}

The \texttt{sbi} library~\cite{BoeltsDeistler_sbi_2025} is arguably the most comprehensive toolkit available for simulation-based inference. Built in PyTorch, it provides a unified interface for neural posterior estimation (NPE), neural likelihood estimation (NLE), and neural ratio estimation (NRE), with normalizing-flow backbones --- masked autoregressive flows~\cite{Papamakarios:2017tec} and neural spline flows~\cite{Durkan:2019nsq} --- as the primary density estimators. Recent versions added flow matching and diffusion models as density estimators and integrated the \texttt{lampe} library~\cite{Rozet:2025lampe} for composable normalizing-flow architectures. The breadth of this method coverage --- spanning all three inference strategies, multiple density estimator families, and built-in calibration diagnostics --- makes \texttt{sbi} the standard reference for SBI research.

A complementary approach is truncated marginal neural ratio estimation (TMNRE), implemented by the \texttt{swyft} library~\cite{Miller2022}. Rather than fitting the full posterior, \texttt{swyft} trains an amortised likelihood-to-evidence ratio estimator that targets user-specified marginals, avoiding the cost of learning the joint posterior when only low-dimensional projections are of interest. The library has been widely adopted in astrophysics and cosmology, with applications ranging from gravitational wave parameter estimation to CMB inference and strong lensing substructure detection~\cite{Miller2022}.

\texttt{BayesFlow}~\cite{Radev:2020bayesflow} implements amortised Bayesian workflows using invertible neural networks, built on Keras for backend-agnostic deployment across PyTorch, JAX, and TensorFlow. A distinctive feature is its built-in summary statistics network, which learns to compress high-dimensional or variable-length observations into fixed-size representations jointly with the inference network. \texttt{BayesFlow} targets a different generative backbone and inference niche than GenSBI, but shares the goal of making modern neural density estimation accessible to domain scientists.

These toolkits are excellent, production-grade software, and the choice of PyTorch is a well-motivated design decision with clear advantages in community size and ecosystem breadth. GenSBI complements them along two axes. First, by focusing exclusively on flow and diffusion models, GenSBI can explore the design space of these methods more thoroughly than a general-purpose toolkit: it offers three generative formulations, multiple solver families that can be swapped post-training, and transformer-based architectures purpose-built for scientific data, all under a modular design in which the generative method, architecture, and inference mode can each be varied independently. This narrower scope allows a more comprehensive and flexible implementation of the methods it supports. Second, GenSBI aims to provide a solid SBI option in the JAX ecosystem, where a growing number of scientific workflows --- differentiable simulators, probabilistic programming, hardware-accelerated sampling --- already run natively. For researchers working in that ecosystem, GenSBI offers a way to do modern NPE without switching frameworks.

\subsection{Design inspirations}
\label{sec:design_lineage}

GenSBI's design draws on two threads: architectural inspiration from recent SBI models, and algorithmic foundations from the generative modelling literature.

The most direct architectural influence is SimFormer~\cite{Gloeckler:2024simformer}, which introduced the ``all-in-one'' approach to SBI: a transformer-based diffusion model trained with random masking on the joint distribution, enabling posterior, likelihood, and joint queries from a single model. GenSBI adopts a similar model architecture (Section~\ref{sec:nn_architectures}) and draws inspiration from the joint estimation paradigm, but reimplements the generative methods from scratch to support flow matching, score matching, and denoising diffusion (EDM). The SimFormer codebase is tied to a single generative formulation; GenSBI decouples the architecture from the generative method so that both can be varied independently. Replacing SimFormer's score-based diffusion backbone with a flow matching objective is a natural extension of the framework, and two independent implementations pursued it concurrently: \texttt{OneFlowSBI}~\cite{Nautiyal:2026oneflowsbi} and GenSBI's \texttt{Flux1Joint}, both released in January 2026. The core methodology --- masked conditional flow matching over the concatenated joint space --- is the same in both; a detailed comparison of their architectural and methodological differences is given in Section~\ref{sec:nn_architectures}. 
A key distinction at the software level is that \texttt{OneFlowSBI} does not open-source its codebase, and its implementation is tightly coupled to a specific generative formulation and network architecture; GenSBI's strategy-pattern design, by contrast, allows any generative method to be composed with any compatible architecture through a single configuration change.

On the algorithmic side, GenSBI's three generative formulations build on the theoretical frameworks introduced by Meta's \texttt{flow\_matching} library~\cite{Lipman:2024fmguide} for optimal-transport conditional flow matching, NVIDIA's EDM repository~\cite{Karras:2022edm} for diffusion training and sampling, and Yang Song's \texttt{score\_sde}~\cite{Song:2020sde} for score-based generative modelling via stochastic differential equations. GenSBI does not call, wrap, or bundle any of these codebases; all algorithms were reimplemented from scratch in JAX, consulting the original implementations to verify algorithmic correctness. At a design-philosophy level, the pipeline-based architecture of the HuggingFace \texttt{diffusers} library~\cite{vonPlaten:2022diffusers} also informed GenSBI's API design.

\subsection{The JAX ecosystem}
\label{sec:jax_ecosystem}

Beyond the SBI-specific tools discussed above, GenSBI sits within a maturing JAX ecosystem for scientific computing and Bayesian inference. The only other SBI-specific JAX package, \texttt{sbijax}~\cite{Dirmeier:2024sbijax}, covers a limited subset of inference methods. The broader ecosystem, however, provides substantial infrastructure: a scientist can now assemble an entire inference pipeline --- differentiable simulator, neural density estimation, and MCMC refinement --- without leaving a single framework. \texttt{diffrax} ~\cite{Kidger:2022diffrax} provides GPU-accelerated ODE and SDE solvers that GenSBI uses directly for the numerical integration underlying all three generative methods. \texttt{BlackJAX}~\cite{Cabezas:2024blackjax} provides composable MCMC and sequential Monte Carlo samplers, and \texttt{NumPyro}~\cite{Phan:2019numpyro} offers a probabilistic programming interface; both are widely adopted in the physics and statistics communities. Scientists who already use these tools can now add NPE via GenSBI to their workflow without switching frameworks.

More directly relevant to GenSBI's technical scope are the emerging JAX libraries for normalizing flows and optimal transport. \texttt{bijx}~\cite{Gerdes2025Bijx} provides bijections and normalizing flows built on \texttt{Flax}   NNX, sharing the same neural network framework that GenSBI uses, and \texttt{ott-jax}~\cite{Cuturi:2022ottjax} implements optimal transport solvers including Sinkhorn divergences and linear programs. Both provide numerical primitives that GenSBI could leverage for specialised operations in the future, without reimplementing them internally.

\subsection{Feature comparison}
\label{sec:feature_comparison}

\definecolor{TableGray}{gray}{0.92}
\newcolumntype{g}{>{\columncolor{TableGray}}c}

\begin{table}[!t]
\centering
\caption{Feature comparison of simulation-based inference libraries. Columns report the underlying framework, supported inference strategies, density estimator families, and distinguishing architectural capabilities. The $\sim$ entry for \texttt{sbi} indicates that NJE is reachable through specific configurations rather than exposed as a first-class strategy. The $\sim$ entry for GenSBI indicates that NLE is supported in principle via PF-ODE-based exact log-probabilities, but each query requires a full ODE solve and is therefore impractical as an MCMC inner loop (see Section~\ref{sec:limitations}).}
\label{tab:comparison}
\small
\setlength{\aboverulesep}{0pt}
\setlength{\belowrulesep}{0pt}
\setlength{\extrarowheight}{.75ex}
\begin{tabular}{@{}lcccg}
\toprule
\textbf{Feature} & \texttt{sbi} & \texttt{BayesFlow} & \texttt{swyft} & \textbf{GenSBI} \\
\midrule
Framework         & PyTorch & Keras & PyTorch & \textbf{JAX} \\
NPE               & \checkmark & \checkmark & ---    & \checkmark \\
NLE               & \checkmark & ---        & ---    & $\sim$     \\
NJE               & $\sim$     & \checkmark & ---    & \checkmark \\
NRE               & \checkmark & ---        & \checkmark & ---   \\
Flow matching     & \checkmark & ---        & ---    & \checkmark \\
Score matching    & \checkmark & ---        & ---    & \checkmark \\
EDM diffusion     & ---        & ---        & ---    & \checkmark \\
Normalizing flows & \checkmark & \checkmark & ---    & ---        \\
Transformer archs & $\sim$     & ---        & ---    & \checkmark \\
Calibration checks & \checkmark & \checkmark & \checkmark & \checkmark \\
\bottomrule
\end{tabular}
\end{table}

Table~\ref{tab:comparison} summarises these differences. Together, the tools discussed in this section cover most of the SBI workflow: \texttt{sbi} provides the broadest method coverage in PyTorch, \texttt{swyft} targets efficient marginal inference, and \texttt{BayesFlow} brings amortised workflows with summary networks. GenSBI contributes the piece that was missing from the JAX side of this landscape --- flow matching, score matching, and denoising diffusion (EDM) paired with transformer architectures, under a modular design that lets each component be varied independently.

\subsection{Current limitations}
\label{sec:limitations}

GenSBI is a young library. We discuss its current limitations below and outline how we plan to address them in future releases.

\paragraph{Inference strategies and density-estimator families.}
Neural ratio estimation is not implemented in the current release; users who need amortised likelihood-to-evidence ratios or truncated marginal variants should use \texttt{sbi} or \texttt{swyft}, which cover these methods thoroughly. Normalizing flows are also absent from the initial public release. This is a scoping choice, not a judgement on the method: normalizing flows remain the best option for workflows that require single-pass log-probability evaluation, including neural likelihood estimation with MCMC inner loops, Bayesian evidence computation, and any pipeline in which the density is queried far more often than it is sampled. A future release will add normalizing-flow backbones to GenSBI.

\paragraph{Exact log-probability and the role of normalizing flows.}
Flow matching and diffusion models can evaluate exact log-probabilities via the probability-flow ODE, and this is how neural likelihood estimation works in GenSBI today. Each density query, however, requires a full ODE solve. This cost is acceptable for one-off evaluations or Bayesian model comparison with a modest number of likelihood calls, but it becomes impractical for MCMC, where every step in the chain pays the ODE cost. This is the distinction marked by the $\sim$ entry for GenSBI's NLE support in Table~\ref{tab:comparison}. Normalizing flows avoid this bottleneck through single-pass evaluation; the planned NF release will address the density-evaluation-dominated use case directly.

\paragraph{Benchmark dimensionality.}
The SBIBM tasks used in Section~\ref{sec:posterior_quality} cap at $\dim(\theta) = 10$ (Gaussian Linear and Bernoulli GLM), and the two advanced applications target low-dimensional parameter spaces ($\dim(\theta) = 2$ for both the gravitational-wave and strong-lensing examples). Flow matching with transformer backbones has been applied to much larger parameter spaces in the literature~\cite{Dax:2024fmpe}, but this scaling has not yet been tested inside GenSBI at $\dim(\theta) \gtrsim 10$. The \texttt{GenSBI-examples} companion repository~\cite{GenSBI_examples} covers a literature-standard set of tasks; higher-dimensional benchmarks are planned (Section~\ref{sec:conclusion}).

\paragraph{EDM at small simulation budgets.}
As noted in Section~\ref{sec:posterior_quality}, the EDM formulation underperforms flow matching and score matching at $10^4$ simulations when paired with the Flux1Joint architecture: C2ST scores reach $0.825$ on Gaussian Mixture and $0.871$ on SLCP, well above FM or SM values under identical conditions. We suspect the EDM preconditioning and noise schedule from~\cite{Karras:2022edm} were designed for high-dimensional image distributions, where most of the loss mass concentrates at intermediate noise levels $\sigma$, and may not transfer well to the low-dimensional joint densities in SBI. A systematic study of alternative noise schedules, loss weightings, and their interaction with joint-distribution training is left to future work. EDM remains competitive at larger budgets and offers one of the most aggressive step-count reductions at sampling time, so the investigation is warranted.

\paragraph{Structured, field-level outputs.}
The current GenSBI pipelines target unstructured parameter vectors: the model learns whatever correlations exist between parameter components, but the architecture does not encode prior knowledge about the structure of the output space. For outputs with a known structure (images, dense spatial fields, time series), specialised architectural choices can exploit that structure directly: RoPE-based spatial embeddings or the joint input/output embedding strategy of the Flux Kontext family~\cite{BlackForestLabs:2025flux} are two examples. First-class support for structured outputs, with these components available as configuration options, is on the roadmap.

\paragraph{Library maturity.}
GenSBI has a small developer base compared to established toolkits such as \texttt{sbi}. The API may change between minor versions, documentation is still growing, and the user community is small. We view these as the normal conditions of an initial public release and expect them to improve as adoption grows. Contributions, bug reports, and use-case feedback through the project repository are welcome; the modular, strategy-pattern design is intended to make adding new generative methods, architectures, or diagnostics straightforward.

\section{Conclusions}
\label{sec:conclusion}

We have presented GenSBI, an open-source, JAX-native library for simulation-based inference built on flow matching and diffusion models. Systematic benchmarking on all five considered SBIBM tasks shows that the framework achieves competitive or state-of-the-art posterior quality: at a simulation budget of $10^5$, the best model configurations reach C2ST scores of $0.500$--$0.502$ on the Gaussian and Two Moons tasks and $0.534$ on SLCP, outperforming both SimFormer ($0.566$) and NPE ($0.742$) on this challenging multimodal problem. Notably, all three generative formulations --- flow matching, score matching, and EDM --- converge to comparable accuracy as the simulation budget grows, empirically validating their interchangeability within the unified interface. These results are obtained with a nearly uniform training configuration, without extensive per-task hyperparameter tuning. Beyond the standard benchmarks, the framework produces well-calibrated posteriors on two example scientific applications involving high-dimensional structured observations --- gravitational wave strain time series and strong gravitational lensing images --- demonstrating that the modular separation of embedding network and transformer-based density estimator generalises to the kind of inference problems encountered in practice.

The empirical performance reported above is enabled by a library designed around composability. GenSBI implements flow matching, score matching, and denoising diffusion (EDM) as interchangeable density estimation strategies, paired with three transformer-based architectures: SimFormer, Flux1, and a novel Flux1Joint that combines Flux1's adaLN-Zero modulation with SimFormer's condition mask mechanism for joint density estimation. A strategy-pattern interface decouples the generative method from the neural backbone and the inference mode, allowing users to swap any component --- or introduce entirely new ones --- without modifying the rest of the pipeline. Because the library is built entirely in JAX on top of Flax, diffrax, NumPyro, and Orbax, it fills a gap in the current SBI software ecosystem: no other JAX-native library currently provides flow matching and diffusion-based neural posterior estimation with modern transformer architectures and integrated calibration diagnostics.

Several directions for future work follow naturally from the current release, and a fuller discussion of current limitations is provided in Section~\ref{sec:limitations}. A natural first extension is the addition of normalizing-flow backbones, which would enable single-pass log-probability evaluation for MCMC-coupled workflows and Bayesian model comparison --- use cases where flow matching and diffusion models incur substantial computational overhead due to ODE-based density evaluation. A complementary direction is broadening the benchmark suite to higher-dimensional parameter spaces, where the scaling behaviour of flow matching relative to normalizing flows can be characterized empirically within a single, controlled implementation. Additional neural network architectures --- including lightweight alternatives to transformers --- would further expand the range of accessible use cases. Although the applications presented here are drawn from physics, the framework is domain-agnostic: it applies wherever a stochastic simulator defines an implicit likelihood, from epidemiological modelling and tumour dynamics to computational neuroscience. By filling the current gap for a JAX-native SBI library and offering flow matching and diffusion models within a single composable interface, GenSBI provides a practical entry point for researchers whose workflows already rely on the JAX ecosystem. The code and documentation are publicly available at \url{https://github.com/aurelio-amerio/GenSBI}.

\begin{ack}
I thank 
Adrian Bayer, 
Carolina Cuesta-Lazaro, 
Androniki Dimitriou, 
Konstantin Leyde, 
Julia Linhart, 
Dmitry Malyshev,
Elena Pinetti, 
Francisco Villaescusa-Navarro,
and Bryan Zaldivar for
helpful conversations.
AA is grateful for the hospitality of the Center for Computational Astrophysics (Flatiron Institute, New York) where part of this work was carried out. 
This research was supported by the Generalitat Valenciana through the ``GenT program'',
ref.: CIDEGENT/2020/055 (PI: B.~Zaldivar).
The author acknowledges the computer resources at Artemisa and the
technical support provided by the Instituto de F\'isica Corpuscular, IFIC (CSIC-UV).
Artemisa is co-funded by the European Union through the 2014--2020 ERDF Operative
Programme of Comunitat Valenciana, project IDIFEDER/2018/048.
\end{ack}

\section*{AI Usage Disclosure}
The author acknowledges the use of large language models (LLMs) as writing aids during the preparation of this manuscript.
LLMs were used exclusively for drafting, editing, and formatting tasks;
they were not involved in the design of experiments, the development of methods,
or the analysis of results.
All scientific content, claims, and conclusions are the sole
responsibility of the human authors, who have thoroughly reviewed and approved
the final text.

\bibliographystyle{JHEP}
\bibliography{biblio}

\conditionalpagebreak
\appendix

\section{Additional Benchmarks and Configurations}
\label{app:additional_benchmarks}

This appendix presents additional benchmark results and training configurations.

\subsection{SBIBM Results}
\label{app:sbibm_results}

\begin{figure}[t]
    \centering
    \includegraphics[width=\columnwidth]{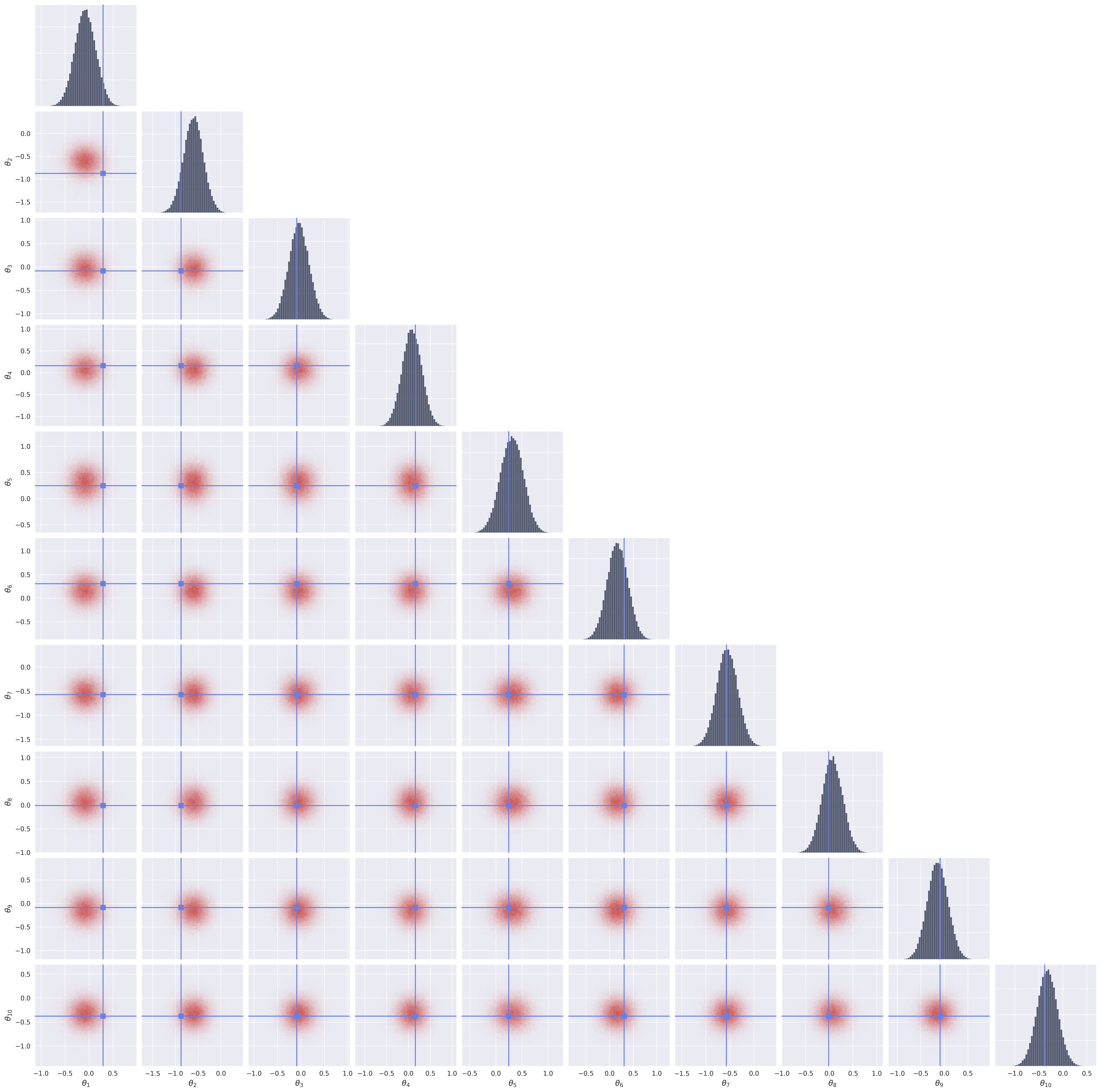}
    \caption{Marginal posterior distributions for the Gaussian Linear task (Flux1Joint, flow matching, EMA). The 10-dimensional Gaussian posterior is accurately recovered.}
    \label{fig:marginals_gaussian_linear}
\end{figure}

\begin{figure}[t]
    \centering
    \includegraphics[width=\columnwidth]{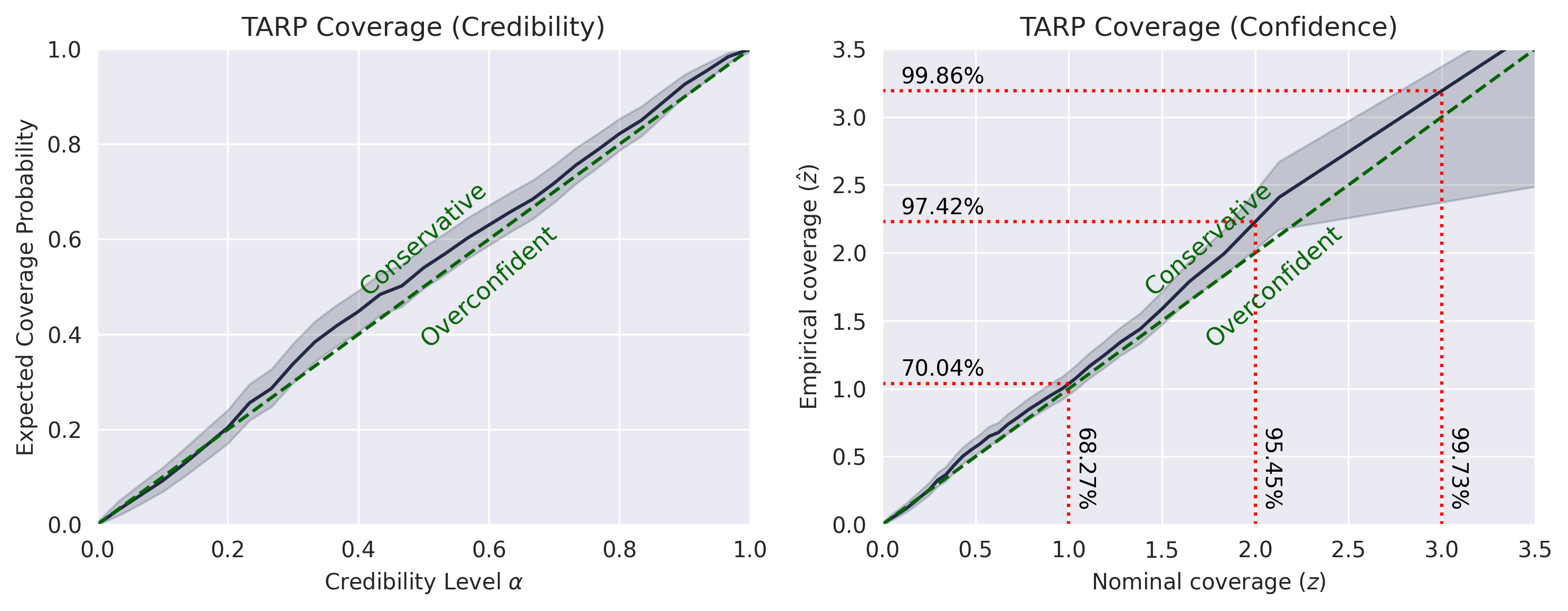}
    \caption{TARP expected coverage probability curve for the Gaussian Linear task.}
    \label{fig:tarp_gaussian_linear}
\end{figure}

\begin{figure}[t]
    \centering
    \includegraphics[width=0.5\columnwidth]{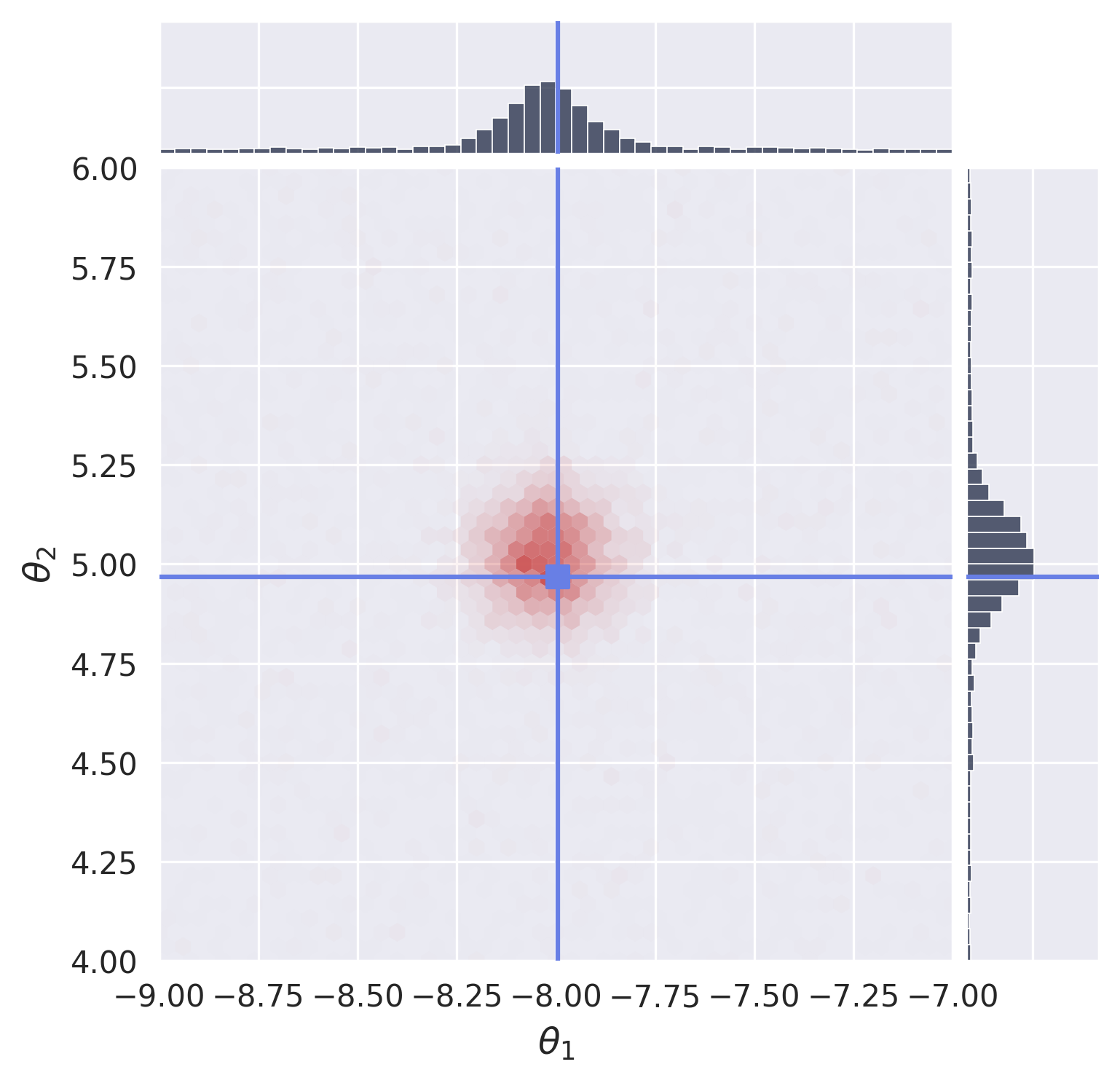}
    \caption{Marginal posterior distributions for the Gaussian Mixture task (Flux1Joint, flow matching, EMA).}
    \label{fig:marginals_gaussian_mixture}
\end{figure}

\begin{figure}[t]
    \centering
    \includegraphics[width=\columnwidth]{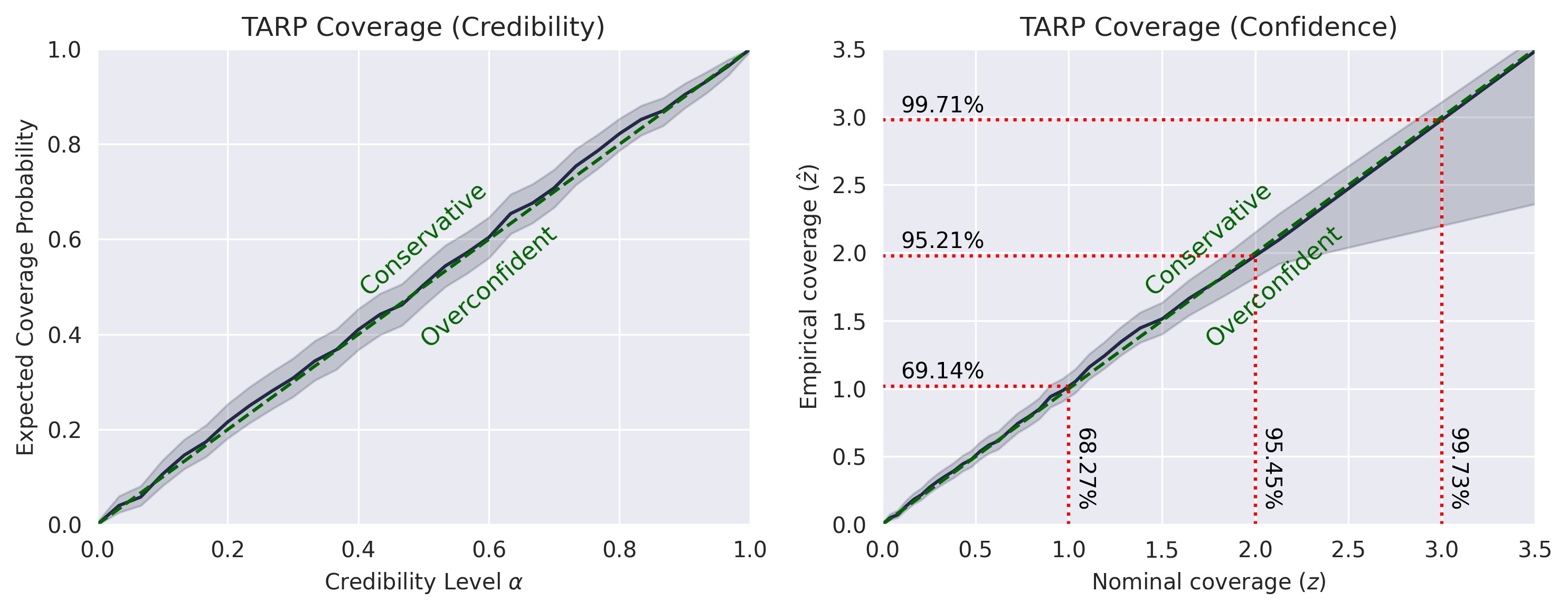}
    \caption{TARP expected coverage probability curve for the Gaussian Mixture task.}
    \label{fig:tarp_gaussian_mixture}
\end{figure}

\begin{figure}[t]
    \centering
    \includegraphics[width=\columnwidth]{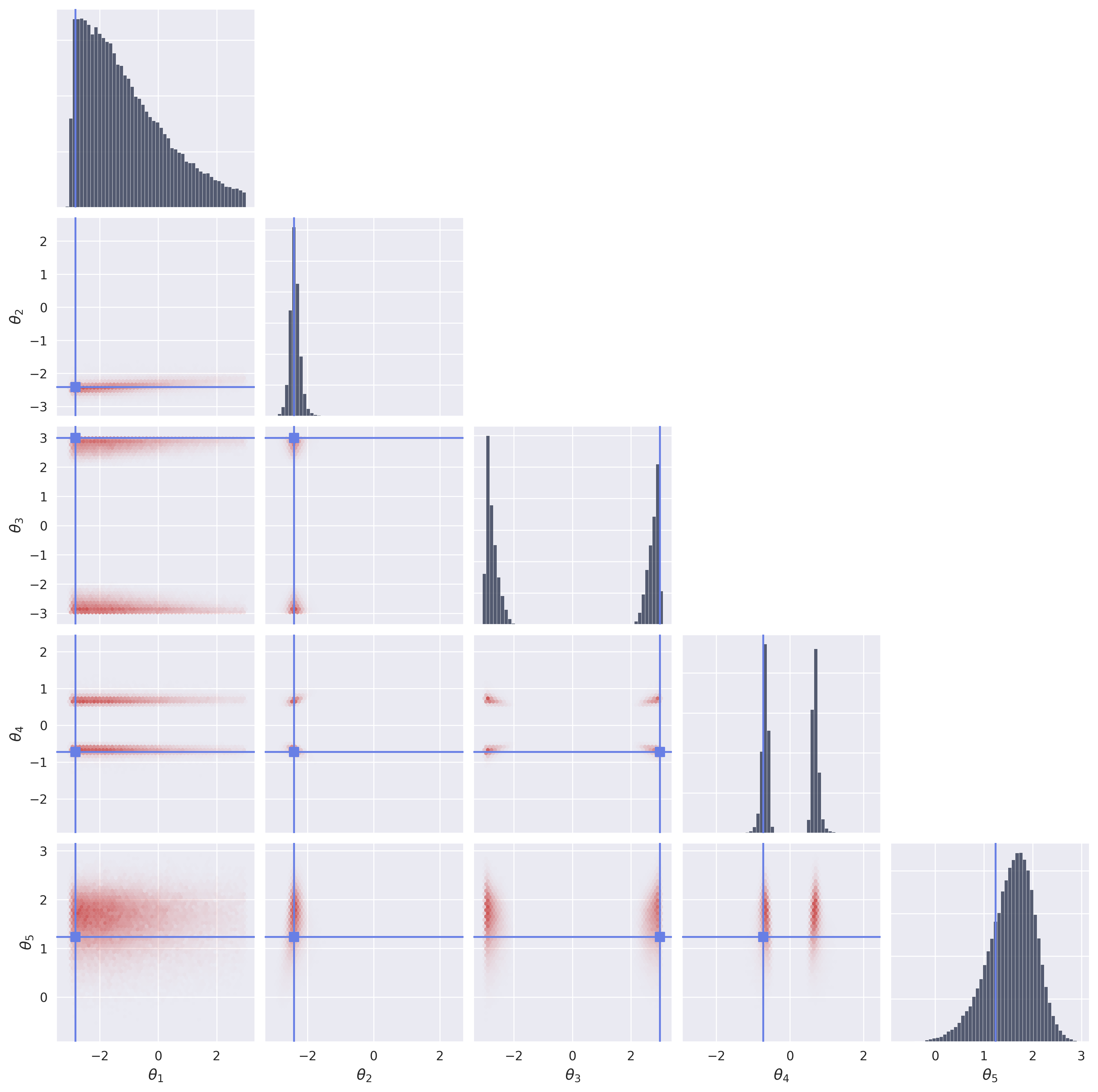}
    \caption{Marginal posterior distributions for the SLCP task (Flux1Joint, flow matching, EMA). The four-mode structure and sharp boundaries are captured.}
    \label{fig:marginals_slcp}
\end{figure}

\begin{figure}[t]
    \centering
    \includegraphics[width=\columnwidth]{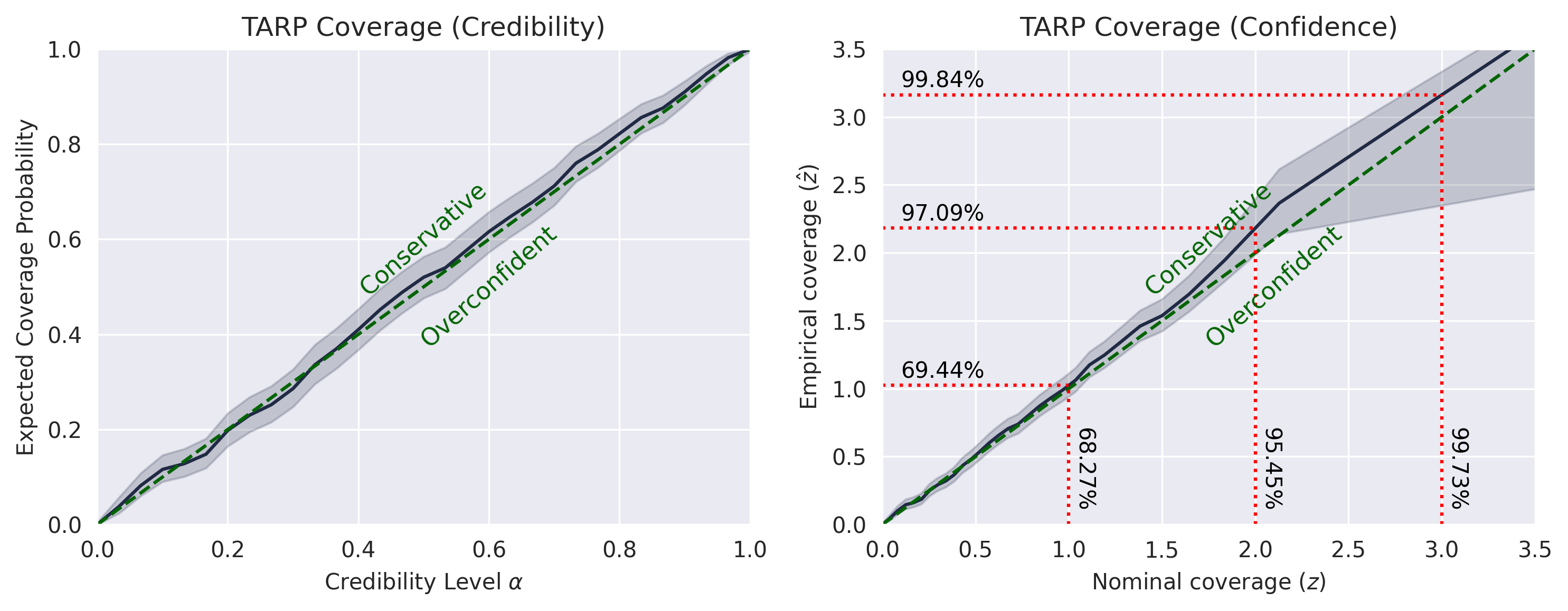}
    \caption{TARP expected coverage probability curve for the SLCP task.}
    \label{fig:tarp_slcp}
\end{figure}

\begin{figure}[t]
    \centering
    \includegraphics[width=\columnwidth]{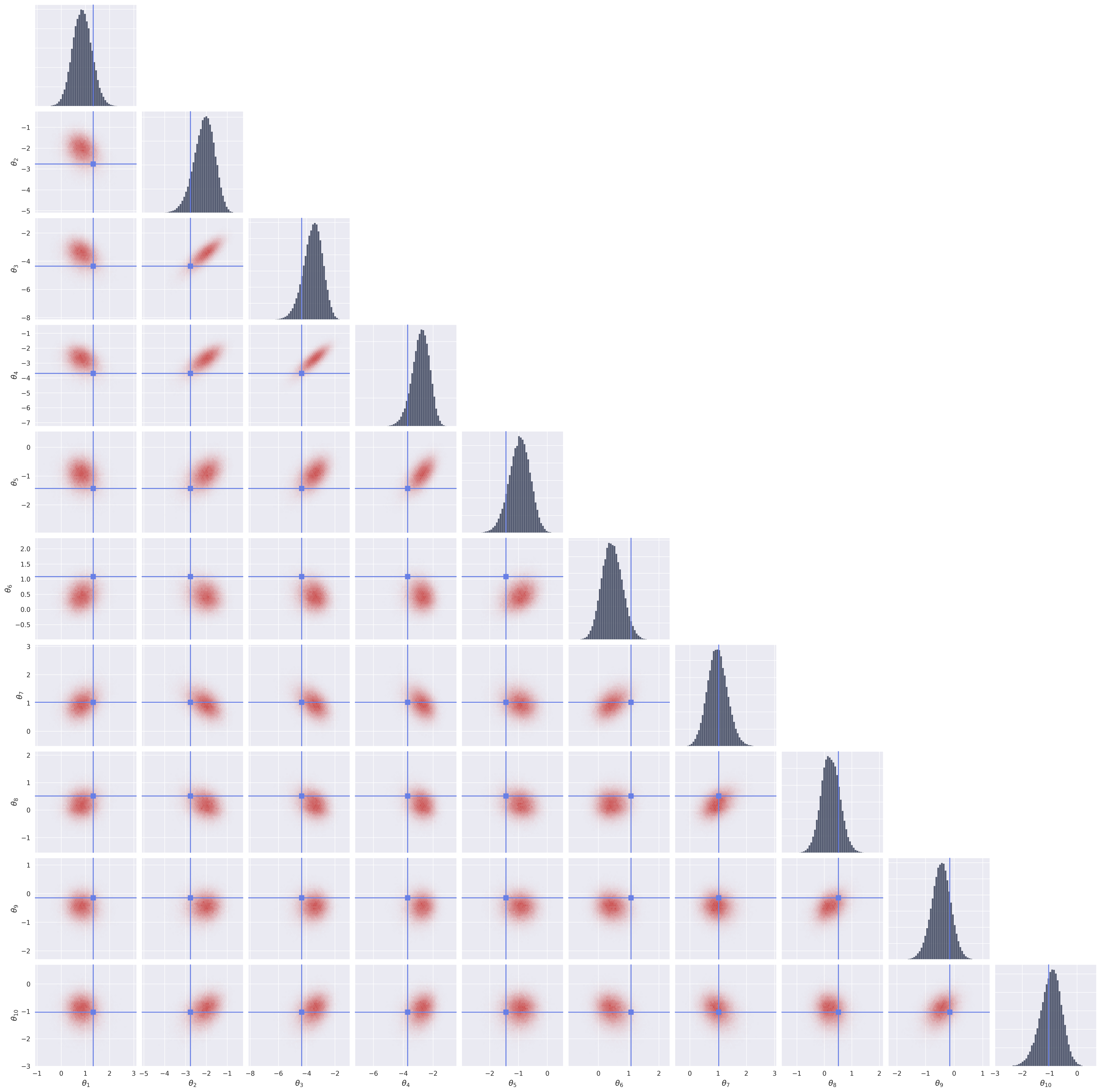}
    \caption{Marginal posterior distributions for the Bernoulli GLM task (Flux1Joint, flow matching, EMA). The correlated 10-dimensional posterior is faithfully reproduced.}
    \label{fig:marginals_bernoulli_glm}
\end{figure}

\begin{figure}[t]
    \centering
    \includegraphics[width=\columnwidth]{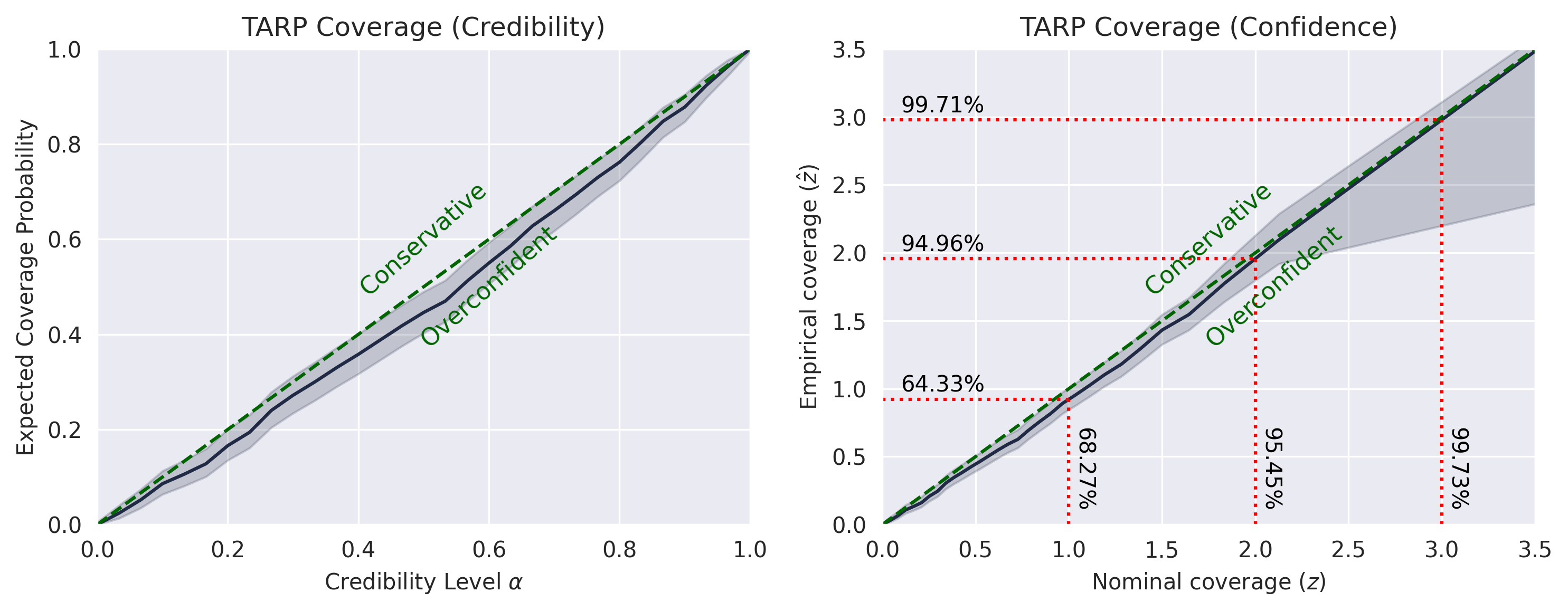}
    \caption{TARP expected coverage probability curve for the Bernoulli GLM task.}
    \label{fig:tarp_bernoulli_glm}
\end{figure}

This appendix presents the marginal posterior distributions and TARP calibration curves for the four SBIBM tasks not shown in the main text: Gaussian Linear (Figure \ref{fig:marginals_gaussian_linear} and \ref{fig:tarp_gaussian_linear}), Gaussian Mixture (Figure \ref{fig:marginals_gaussian_mixture} and \ref{fig:tarp_gaussian_mixture}), SLCP (Figure \ref{fig:marginals_slcp} and \ref{fig:tarp_slcp}), and Bernoulli GLM (Figure \ref{fig:marginals_bernoulli_glm} and \ref{fig:tarp_bernoulli_glm}). All results use the Flux1Joint model with flow matching and EMA parameters.

\subsection{C2ST Results Across Simulation Budgets}
\label{app:c2st_budget_tables}

Tables~\ref{tab:c2st_flux1} and~\ref{tab:c2st_flux1joint} report the best C2ST accuracy achieved by the Flux1 and Flux1Joint architectures, respectively, across three simulation budgets and three generative methods. These values correspond to the plots in Figures~\ref{fig:c2st_budget_flux1} and~\ref{fig:c2st_budget_flux1joint}. Best results per task and budget are highlighted in \colorbox{gray!20}{\textbf{bold}}.

\begin{table*}[t]
\centering
\small
\renewcommand{\arraystretch}{1.5}
\setlength{\tabcolsep}{3pt}
\caption{Best C2ST accuracy for the Flux1 architecture across simulation budgets and generative methods; best per task and budget highlighted.}
\label{tab:c2st_flux1}

\begin{subtable}[t]{\textwidth}
\centering
\begin{tabular}{M|CCC|CCC|CCC}
\Xhline{1.5pt}
\rowcolor{white!10}
Method
& \multicolumn{3}{c|}{Two Moons}
& \multicolumn{3}{c|}{Gaussian Linear}
& \multicolumn{3}{c}{Gaussian Mixture} \\
\rowcolor{white!10}
& $10\text{k}$ & $30\text{k}$ & $100\text{k}$ & $10\text{k}$ & $30\text{k}$ & $100\text{k}$ & $10\text{k}$ & $30\text{k}$ & $100\text{k}$ \\
\hline
Flow Matching
& $0.549$ \newline $\pm 0.013$ & $0.529$ \newline $\pm 0.014$ & $0.525$ \newline $\pm 0.007$ & $0.714$ \newline $\pm 0.043$ & \bestcell{$0.508$ \newline $\pm 0.007$} & $0.501$ \newline $\pm 0.006$ & $0.553$ \newline $\pm 0.025$ & $0.520$ \newline $\pm 0.014$ & $0.510$ \newline $\pm 0.006$ \\
Score Matching
& \bestcell{$0.529$ \newline $\pm 0.016$} & \bestcell{$0.516$ \newline $\pm 0.011$} & \bestcell{$0.507$ \newline $\pm 0.009$} & \bestcell{$0.666$ \newline $\pm 0.048$} & $0.509$ \newline $\pm 0.006$ & \bestcell{$0.500$ \newline $\pm 0.005$} & \bestcell{$0.524$ \newline $\pm 0.009$} & \bestcell{$0.510$ \newline $\pm 0.006$} & \bestcell{$0.502$ \newline $\pm 0.005$} \\
EDM
& $0.554$ \newline $\pm 0.018$ & $0.534$ \newline $\pm 0.015$ & $0.523$ \newline $\pm 0.009$ & $0.718$ \newline $\pm 0.128$ & $0.526$ \newline $\pm 0.007$ & $0.523$ \newline $\pm 0.005$ & $0.562$ \newline $\pm 0.024$ & $0.520$ \newline $\pm 0.007$ & $0.510$ \newline $\pm 0.005$ \\
\Xhline{1.5pt}
\end{tabular}
\end{subtable}

\vspace{0.6cm}

\begin{subtable}[t]{\textwidth}
\centering
\begin{tabular}{M|CCC|CCC}
\Xhline{1.5pt}
\rowcolor{white!10}
Method
& \multicolumn{3}{c|}{SLCP}
& \multicolumn{3}{c}{Bernoulli GLM} \\
\rowcolor{white!10}
& $10\text{k}$ & $30\text{k}$ & $100\text{k}$ & $10\text{k}$ & $30\text{k}$ & $100\text{k}$ \\
\hline
Flow Matching
& \bestcell{$0.833$ \newline $\pm 0.074$} & $0.725$ \newline $\pm 0.073$ & \bestcell{$0.602$ \newline $\pm 0.042$} & $0.722$ \newline $\pm 0.066$ & $0.696$ \newline $\pm 0.053$ & $0.550$ \newline $\pm 0.016$ \\
Score Matching
& $0.846$ \newline $\pm 0.048$ & \bestcell{$0.692$ \newline $\pm 0.035$} & $0.678$ \newline $\pm 0.035$ & \bestcell{$0.692$ \newline $\pm 0.046$} & \bestcell{$0.579$ \newline $\pm 0.019$} & \bestcell{$0.546$ \newline $\pm 0.021$} \\
EDM
& $0.848$ \newline $\pm 0.051$ & $0.741$ \newline $\pm 0.064$ & $0.617$ \newline $\pm 0.037$ & $0.701$ \newline $\pm 0.032$ & $0.673$ \newline $\pm 0.054$ & $0.575$ \newline $\pm 0.020$ \\
\Xhline{1.5pt}
\end{tabular}
\end{subtable}

\end{table*}

\begin{table*}[t]
\centering
\small
\renewcommand{\arraystretch}{1.5}
\setlength{\tabcolsep}{3pt}
\caption{Best C2ST accuracy for the Flux1Joint architecture across simulation budgets and generative methods; best per task and budget highlighted.}
\label{tab:c2st_flux1joint}

\begin{subtable}[t]{\textwidth}
\centering
\begin{tabular}{M|CCC|CCC|CCC}
\Xhline{1.5pt}
\rowcolor{white!10}
Method
& \multicolumn{3}{c|}{Two Moons}
& \multicolumn{3}{c|}{Gaussian Linear}
& \multicolumn{3}{c}{Gaussian Mixture} \\
\rowcolor{white!10}
& $10\text{k}$ & $30\text{k}$ & $100\text{k}$ & $10\text{k}$ & $30\text{k}$ & $100\text{k}$ & $10\text{k}$ & $30\text{k}$ & $100\text{k}$ \\
\hline
Flow Matching
& $0.542$ \newline $\pm 0.020$ & $0.529$ \newline $\pm 0.009$ & $0.524$ \newline $\pm 0.009$ & $0.564$ \newline $\pm 0.024$ & $0.512$ \newline $\pm 0.005$ & \bestcell{$0.500$ \newline $\pm 0.003$} & $0.522$ \newline $\pm 0.009$ & $0.515$ \newline $\pm 0.008$ & $0.513$ \newline $\pm 0.007$ \\
Score Matching
& \bestcell{$0.522$ \newline $\pm 0.014$} & \bestcell{$0.513$ \newline $\pm 0.010$} & \bestcell{$0.502$ \newline $\pm 0.007$} & \bestcell{$0.508$ \newline $\pm 0.005$} & \bestcell{$0.500$ \newline $\pm 0.006$} & $0.501$ \newline $\pm 0.003$ & \bestcell{$0.510$ \newline $\pm 0.008$} & \bestcell{$0.504$ \newline $\pm 0.005$} & \bestcell{$0.501$ \newline $\pm 0.005$} \\
EDM
& $0.785$ \newline $\pm 0.047$ & $0.580$ \newline $\pm 0.015$ & $0.573$ \newline $\pm 0.011$ & $0.620$ \newline $\pm 0.015$ & $0.564$ \newline $\pm 0.006$ & $0.562$ \newline $\pm 0.004$ & $0.825$ \newline $\pm 0.091$ & $0.513$ \newline $\pm 0.004$ & $0.516$ \newline $\pm 0.005$ \\
\Xhline{1.5pt}
\end{tabular}
\end{subtable}

\vspace{0.6cm}

\begin{subtable}[t]{\textwidth}
\centering
\begin{tabular}{M|CCC|CCC}
\Xhline{1.5pt}
\rowcolor{white!10}
Method
& \multicolumn{3}{c|}{SLCP}
& \multicolumn{3}{c}{Bernoulli GLM} \\
\rowcolor{white!10}
& $10\text{k}$ & $30\text{k}$ & $100\text{k}$ & $10\text{k}$ & $30\text{k}$ & $100\text{k}$ \\
\hline
Flow Matching
& \bestcell{$0.657$ \newline $\pm 0.062$} & $0.586$ \newline $\pm 0.028$ & $0.566$ \newline $\pm 0.039$ & \bestcell{$0.616$ \newline $\pm 0.025$} & $0.578$ \newline $\pm 0.021$ & \bestcell{$0.555$ \newline $\pm 0.016$} \\
Score Matching
& $0.663$ \newline $\pm 0.066$ & \bestcell{$0.568$ \newline $\pm 0.024$} & \bestcell{$0.534$ \newline $\pm 0.019$} & $0.630$ \newline $\pm 0.039$ & \bestcell{$0.576$ \newline $\pm 0.029$} & $0.560$ \newline $\pm 0.021$ \\
EDM
& $0.871$ \newline $\pm 0.073$ & $0.668$ \newline $\pm 0.057$ & $0.588$ \newline $\pm 0.022$ & $0.726$ \newline $\pm 0.040$ & $0.624$ \newline $\pm 0.014$ & $0.606$ \newline $\pm 0.011$ \\
\Xhline{1.5pt}
\end{tabular}
\end{subtable}

\end{table*}

\subsection{C2ST Comparison with Literature}
\label{app:c2st_comparison_table}

Table~\ref{tab:c2st_comparison_30k} provides a detailed comparison of all GenSBI model variants against three baselines from the literature at a simulation budget of $3 \times 10^4$. This budget is the largest common point reported in the OneFlowSBI benchmark~\cite{Nautiyal:2026oneflowsbi} and allows a direct comparison across all methods. Best values per task are highlighted in \colorbox{gray!20}{\textbf{bold}}; second-best in \textit{italics}.

\begin{table*}[t]
\centering
\renewcommand{\arraystretch}{1.25}
\setlength{\tabcolsep}{5pt}
\caption{C2ST accuracy at $3 \times 10^4$ simulations for all GenSBI model variants and three literature baselines, rounded to two decimal places to account for the C2ST undertainty. Best value per task in \textbf{bold} with gray background; second-best in \textit{italics}. OneFlowSBI and SimFormer values from Nautiyal et al.~\cite{Nautiyal:2026oneflowsbi}; NPE values from the \texttt{sbi} library~\cite{BoeltsDeistler_sbi_2025} as reported in Nautiyal et al.~\cite{Nautiyal:2026oneflowsbi}.}
\label{tab:c2st_comparison_30k}
\begin{adjustbox}{max width=\textwidth}
\begin{tabular}{>{\columncolor{white!10}}l|c|c|c|c|c}
\Xhline{1.5pt}
\rowcolor{white!10}
Method & Two Moons & Gauss.\ Linear & Gauss.\ Mixture & SLCP & Bernoulli GLM \\
\hline
\rowcolor{gray!5}
GenSBI FM (Flux1) & $0.53$ & \secondcell{0.51} & $0.52$ & $0.72$ & $0.70$ \\
\rowcolor{gray!5}
GenSBI SM (Flux1) & \secondcell{0.52} & \secondcell{0.51} & \secondcell{0.51} & $0.69$ & \bestcell{0.58} \\
\rowcolor{gray!5}
GenSBI FM (Flux1Joint) & $0.53$ & \secondcell{0.51} & $0.52$ & \secondcell{0.59} & \bestcell{0.58} \\
\rowcolor{gray!5}
GenSBI SM (Flux1Joint) & \bestcell{0.51} & \bestcell{0.50} & \bestcell{0.50} & \bestcell{0.57} & \bestcell{0.58} \\
\hline
OneFlowSBI & \bestcell{0.51} & \secondcell{0.51} & \secondcell{0.51} & $0.73$ & \bestcell{0.58} \\
SimFormer & \bestcell{0.51} & \bestcell{0.50} & \secondcell{0.51} & \bestcell{0.57} & \secondcell{0.59} \\
NPE & $0.57$ & $0.55$ & $0.56$ & $0.84$ & $0.65$ \\
\Xhline{1.5pt}
\end{tabular}
\end{adjustbox}
\end{table*}

\subsection{Training Configurations}
\label{app:training_configs}

Tables~\ref{tab:config_flux1_two_moons}--\ref{tab:config_flux1_slcp} and~\ref{tab:config_flux1joint_two_moons}--\ref{tab:config_flux1joint_slcp} report the complete training and architecture configurations for all combinations of task, generative method, and simulation budget used in the systematic comparison of Section~\ref{sec:posterior_quality}. Each table shows the hyperparameters for one architecture--task pair across the three generative methods (flow matching, EDM, score matching) and three budgets ($10^4$, $3 \times 10^4$, $10^5$). The last row of each table reports the best C2ST accuracy achieved by that configuration. The complete YAML configuration files are also available in the \texttt{GenSBI-examples} repository~\cite{GenSBI_examples}.

\begin{table}[t]
\centering
\caption{Training configuration for Flux1 — Two Moons.}
\label{tab:config_flux1_two_moons}
\scriptsize
\resizebox{\textwidth}{!}{%
\begin{tabular}{@{}l|ccc|ccc|ccc@{}}
\toprule
 & \multicolumn{3}{c|}{Flow Matching} & \multicolumn{3}{c|}{Diffusion (EDM)} & \multicolumn{3}{c}{Score Matching} \\
\textbf{Parameter} & \textbf{10k} & \textbf{30k} & \textbf{100k} & \textbf{10k} & \textbf{30k} & \textbf{100k} & \textbf{10k} & \textbf{30k} & \textbf{100k} \\
\midrule
Batch size & 256 & 256 & 1\,024 & 4\,096 & 1\,024 & 1\,024 & 256 & 256 & 4\,096 \\
Training steps & 50\,000 & 100\,000 & 100\,000 & 50\,000 & 90\,000 & 100\,000 & 100\,000 & 100\,000 & 50\,000 \\
Peak learning rate & $10^{-4}$ & $10^{-4}$ & $2 \times 10^{-4}$ & $4 \times 10^{-4}$ & $2 \times 10^{-4}$ & $2 \times 10^{-4}$ & $10^{-4}$ & $10^{-4}$ & $4 \times 10^{-4}$ \\
Minimum learning rate & $10^{-6}$ & $10^{-6}$ & $2 \times 10^{-6}$ & $4 \times 10^{-6}$ & $2 \times 10^{-6}$ & $2 \times 10^{-6}$ & $10^{-6}$ & $10^{-6}$ & $4 \times 10^{-6}$ \\
Warmup steps & 500 & 500 & 1\,000 & 500 & 1\,000 & 1\,000 & 500 & 500 & 500 \\
EMA decay & 0.999 & 0.9999 & 0.999 & 0.999 & 0.999 & 0.999 & 0.9999 & 0.9999 & 0.999 \\
Single-stream blocks & 8 & 8 & 16 & 16 & 16 & 16 & 8 & 8 & 16 \\
Double-stream blocks & 4 & 4 & 8 & 8 & 8 & 8 & 4 & 4 & 8 \\
Attention heads & 4 & 4 & 4 & 4 & 4 & 4 & 4 & 4 & 4 \\
ID merge mode & concat & concat & sum & sum & sum & sum & concat & concat & sum \\
Val.~emb.~dim & 10 & 10 & 10 & 10 & 10 & 10 & 10 & 10 & 10 \\
ID emb.~dim & 10 & 10 & --- & --- & --- & --- & 10 & 10 & --- \\
\midrule
Best C2ST & 0.549 & 0.529 & 0.525 & 0.554 & 0.534 & 0.523 & 0.529 & 0.516 & 0.507 \\
\bottomrule
\end{tabular}%
}
\end{table}

\begin{table}[t]
\centering
\caption{Training configuration for Flux1 — Bernoulli GLM.}
\label{tab:config_flux1_bernoulli_glm}
\scriptsize
\resizebox{\textwidth}{!}{%
\begin{tabular}{@{}l|ccc|ccc|ccc@{}}
\toprule
 & \multicolumn{3}{c|}{Flow Matching} & \multicolumn{3}{c|}{Diffusion (EDM)} & \multicolumn{3}{c}{Score Matching} \\
\textbf{Parameter} & \textbf{10k} & \textbf{30k} & \textbf{100k} & \textbf{10k} & \textbf{30k} & \textbf{100k} & \textbf{10k} & \textbf{30k} & \textbf{100k} \\
\midrule
Batch size & 256 & 256 & 256 & 256 & 4\,096 & 256 & 4\,096 & 256 & 256 \\
Training steps & 50\,000 & 50\,000 & 50\,000 & 50\,000 & 50\,000 & 50\,000 & 50\,000 & 50\,000 & 100\,000 \\
Peak learning rate & $10^{-4}$ & $10^{-4}$ & $10^{-4}$ & $10^{-4}$ & $4 \times 10^{-4}$ & $10^{-4}$ & $4 \times 10^{-4}$ & $10^{-4}$ & $10^{-4}$ \\
Minimum learning rate & $10^{-6}$ & $10^{-6}$ & $10^{-6}$ & $10^{-6}$ & $4 \times 10^{-6}$ & $10^{-6}$ & $4 \times 10^{-6}$ & $10^{-6}$ & $10^{-6}$ \\
Warmup steps & 500 & 500 & 500 & 500 & 500 & 500 & 500 & 500 & 500 \\
EMA decay & 0.999 & 0.999 & 0.999 & 0.999 & 0.999 & 0.999 & 0.999 & 0.999 & 0.9999 \\
Single-stream blocks & 8 & 8 & 8 & 8 & 8 & 8 & 8 & 8 & 8 \\
Double-stream blocks & 4 & 4 & 4 & 4 & 4 & 4 & 4 & 4 & 4 \\
Attention heads & 4 & 4 & 4 & 4 & 4 & 4 & 4 & 4 & 4 \\
ID merge mode & concat & sum & concat & concat & sum & concat & sum & sum & concat \\
Val.~emb.~dim & 20 & 20 & 20 & 20 & 20 & 20 & 20 & 20 & 20 \\
ID emb.~dim & 10 & --- & 10 & 10 & --- & 10 & --- & --- & 10 \\
\midrule
Best C2ST & 0.722 & 0.696 & 0.550 & 0.701 & 0.673 & 0.575 & 0.692 & 0.579 & 0.546 \\
\bottomrule
\end{tabular}%
}
\end{table}

\begin{table}[t]
\centering
\caption{Training configuration for Flux1 — Gaussian Linear.}
\label{tab:config_flux1_gaussian_linear}
\scriptsize
\resizebox{\textwidth}{!}{%
\begin{tabular}{@{}l|ccc|ccc|ccc@{}}
\toprule
 & \multicolumn{3}{c|}{Flow Matching} & \multicolumn{3}{c|}{Diffusion (EDM)} & \multicolumn{3}{c}{Score Matching} \\
\textbf{Parameter} & \textbf{10k} & \textbf{30k} & \textbf{100k} & \textbf{10k} & \textbf{30k} & \textbf{100k} & \textbf{10k} & \textbf{30k} & \textbf{100k} \\
\midrule
Batch size & 256 & 256 & 256 & 1\,024 & 4\,096 & 256 & 256 & 256 & 1\,024 \\
Training steps & 50\,000 & 50\,000 & 100\,000 & 100\,000 & 50\,000 & 50\,000 & 50\,000 & 50\,000 & 100\,000 \\
Peak learning rate & $10^{-4}$ & $10^{-4}$ & $10^{-4}$ & $2 \times 10^{-4}$ & $4 \times 10^{-4}$ & $10^{-4}$ & $10^{-4}$ & $10^{-4}$ & $2 \times 10^{-4}$ \\
Minimum learning rate & $10^{-6}$ & $10^{-6}$ & $10^{-6}$ & $2 \times 10^{-6}$ & $4 \times 10^{-6}$ & $10^{-6}$ & $10^{-6}$ & $10^{-6}$ & $2 \times 10^{-6}$ \\
Warmup steps & 500 & 500 & 500 & 1\,000 & 500 & 500 & 500 & 500 & 1\,000 \\
EMA decay & 0.999 & 0.999 & 0.9999 & 0.999 & 0.999 & 0.999 & 0.999 & 0.999 & 0.999 \\
Single-stream blocks & 8 & 8 & 8 & 8 & 8 & 8 & 8 & 8 & 8 \\
Double-stream blocks & 4 & 4 & 4 & 4 & 4 & 4 & 4 & 4 & 4 \\
Attention heads & 4 & 4 & 4 & 4 & 4 & 4 & 4 & 4 & 4 \\
ID merge mode & concat & sum & sum & sum & sum & concat & concat & concat & sum \\
Val.~emb.~dim & 8 & 10 & 10 & 10 & 10 & 8 & 8 & 8 & 10 \\
ID emb.~dim & 4 & --- & --- & --- & --- & 4 & 4 & 4 & --- \\
\midrule
Best C2ST & 0.714 & 0.508 & 0.501 & 0.718 & 0.526 & 0.523 & 0.666 & 0.509 & 0.500 \\
\bottomrule
\end{tabular}%
}
\end{table}

\begin{table}[t]
\centering
\caption{Training configuration for Flux1 — Gaussian Mixture.}
\label{tab:config_flux1_gaussian_mixture}
\scriptsize
\resizebox{\textwidth}{!}{%
\begin{tabular}{@{}l|ccc|ccc|ccc@{}}
\toprule
 & \multicolumn{3}{c|}{Flow Matching} & \multicolumn{3}{c|}{Diffusion (EDM)} & \multicolumn{3}{c}{Score Matching} \\
\textbf{Parameter} & \textbf{10k} & \textbf{30k} & \textbf{100k} & \textbf{10k} & \textbf{30k} & \textbf{100k} & \textbf{10k} & \textbf{30k} & \textbf{100k} \\
\midrule
Batch size & 256 & 256 & 256 & 4\,096 & 256 & 1\,024 & 256 & 256 & 1\,024 \\
Training steps & 50\,000 & 100\,000 & 100\,000 & 50\,000 & 50\,000 & 100\,000 & 50\,000 & 100\,000 & 100\,000 \\
Peak learning rate & $10^{-4}$ & $10^{-4}$ & $10^{-4}$ & $4 \times 10^{-4}$ & $10^{-4}$ & $2 \times 10^{-4}$ & $10^{-4}$ & $10^{-4}$ & $2 \times 10^{-4}$ \\
Minimum learning rate & $10^{-6}$ & $10^{-6}$ & $10^{-6}$ & $4 \times 10^{-6}$ & $10^{-6}$ & $2 \times 10^{-6}$ & $10^{-6}$ & $10^{-6}$ & $2 \times 10^{-6}$ \\
Warmup steps & 500 & 500 & 500 & 500 & 500 & 1\,000 & 500 & 500 & 1\,000 \\
EMA decay & 0.999 & 0.9999 & 0.9999 & 0.999 & 0.999 & 0.999 & 0.999 & 0.9999 & 0.999 \\
Single-stream blocks & 16 & 16 & 16 & 16 & 16 & 16 & 16 & 16 & 16 \\
Double-stream blocks & 8 & 8 & 8 & 8 & 8 & 8 & 8 & 8 & 8 \\
Attention heads & 4 & 4 & 4 & 4 & 4 & 4 & 4 & 4 & 4 \\
ID merge mode & sum & sum & sum & sum & sum & sum & sum & sum & sum \\
Val.~emb.~dim & 10 & 10 & 10 & 10 & 10 & 10 & 10 & 10 & 10 \\
ID emb.~dim & --- & --- & --- & --- & --- & --- & --- & --- & --- \\
\midrule
Best C2ST & 0.553 & 0.520 & 0.510 & 0.562 & 0.520 & 0.510 & 0.524 & 0.510 & 0.502 \\
\bottomrule
\end{tabular}%
}
\end{table}

\begin{table}[t]
\centering
\caption{Training configuration for Flux1 — SLCP.}
\label{tab:config_flux1_slcp}
\scriptsize
\resizebox{\textwidth}{!}{%
\begin{tabular}{@{}l|ccc|ccc|ccc@{}}
\toprule
 & \multicolumn{3}{c|}{Flow Matching} & \multicolumn{3}{c|}{Diffusion (EDM)} & \multicolumn{3}{c}{Score Matching} \\
\textbf{Parameter} & \textbf{10k} & \textbf{30k} & \textbf{100k} & \textbf{10k} & \textbf{30k} & \textbf{100k} & \textbf{10k} & \textbf{30k} & \textbf{100k} \\
\midrule
Batch size & 256 & 256 & 256 & 256 & 256 & 4\,096 & 256 & 4\,096 & 256 \\
Training steps & 50\,000 & 50\,000 & 50\,000 & 50\,000 & 100\,000 & 50\,000 & 50\,000 & 50\,000 & 100\,000 \\
Peak learning rate & $10^{-4}$ & $10^{-4}$ & $10^{-4}$ & $10^{-4}$ & $10^{-4}$ & $4 \times 10^{-4}$ & $10^{-4}$ & $4 \times 10^{-4}$ & $10^{-4}$ \\
Minimum learning rate & $10^{-6}$ & $10^{-6}$ & $10^{-6}$ & $10^{-6}$ & $10^{-6}$ & $4 \times 10^{-6}$ & $10^{-6}$ & $4 \times 10^{-6}$ & $10^{-6}$ \\
Warmup steps & 500 & 500 & 500 & 500 & 500 & 500 & 500 & 500 & 500 \\
EMA decay & 0.999 & 0.999 & 0.999 & 0.999 & 0.9999 & 0.999 & 0.999 & 0.999 & 0.9999 \\
Single-stream blocks & 8 & 8 & 8 & 8 & 8 & 8 & 8 & 8 & 8 \\
Double-stream blocks & 4 & 4 & 4 & 4 & 4 & 4 & 4 & 4 & 4 \\
Attention heads & 6 & 6 & 6 & 6 & 6 & 4 & 6 & 4 & 6 \\
ID merge mode & concat & concat & concat & concat & concat & concat & concat & concat & concat \\
Val.~emb.~dim & 20 & 20 & 20 & 20 & 20 & 20 & 20 & 20 & 20 \\
ID emb.~dim & 10 & 10 & 10 & 10 & 10 & 10 & 10 & 10 & 10 \\
\midrule
Best C2ST & 0.833 & 0.725 & 0.602 & 0.848 & 0.741 & 0.617 & 0.846 & 0.692 & 0.678 \\
\bottomrule
\end{tabular}%
}
\end{table}

\begin{table}[t]
\centering
\caption{Training configuration for Flux1Joint — Two Moons.}
\label{tab:config_flux1joint_two_moons}
\scriptsize
\resizebox{\textwidth}{!}{%
\begin{tabular}{@{}l|ccc|ccc|ccc@{}}
\toprule
 & \multicolumn{3}{c|}{Flow Matching} & \multicolumn{3}{c|}{Diffusion (EDM)} & \multicolumn{3}{c}{Score Matching} \\
\textbf{Parameter} & \textbf{10k} & \textbf{30k} & \textbf{100k} & \textbf{10k} & \textbf{30k} & \textbf{100k} & \textbf{10k} & \textbf{30k} & \textbf{100k} \\
\midrule
Batch size & 256 & 4\,096 & 1\,024 & 4\,096 & 256 & 4\,096 & 256 & 1\,024 & 4\,096 \\
Training steps & 100\,000 & 50\,000 & 100\,000 & 100\,000 & 100\,000 & 100\,000 & 100\,000 & 90\,000 & 50\,000 \\
Peak learning rate & $10^{-4}$ & $4 \times 10^{-4}$ & $2 \times 10^{-4}$ & $4 \times 10^{-4}$ & $10^{-4}$ & $4 \times 10^{-4}$ & $10^{-4}$ & $2 \times 10^{-4}$ & $4 \times 10^{-4}$ \\
Minimum learning rate & $10^{-6}$ & $4 \times 10^{-6}$ & $2 \times 10^{-6}$ & $4 \times 10^{-6}$ & $10^{-6}$ & $4 \times 10^{-6}$ & $10^{-6}$ & $2 \times 10^{-6}$ & $4 \times 10^{-6}$ \\
Warmup steps & 500 & 500 & 1\,000 & 500 & 500 & 500 & 500 & 1\,000 & 500 \\
EMA decay & 0.9999 & 0.999 & 0.999 & 0.9999 & 0.9999 & 0.9999 & 0.9999 & 0.999 & 0.999 \\
Single-stream blocks & 16 & 16 & 16 & 16 & 16 & 16 & 16 & 16 & 16 \\
Double-stream blocks & --- & --- & --- & --- & --- & --- & --- & --- & --- \\
Attention heads & 4 & 4 & 4 & 4 & 4 & 4 & 4 & 4 & 4 \\
\midrule
Best C2ST & 0.542 & 0.529 & 0.524 & 0.785 & 0.580 & 0.573 & 0.522 & 0.513 & 0.502 \\
\bottomrule
\end{tabular}%
}
\end{table}

\begin{table}[t]
\centering
\caption{Training configuration for Flux1Joint — Bernoulli GLM.}
\label{tab:config_flux1joint_bernoulli_glm}
\scriptsize
\resizebox{\textwidth}{!}{%
\begin{tabular}{@{}l|ccc|ccc|ccc@{}}
\toprule
 & \multicolumn{3}{c|}{Flow Matching} & \multicolumn{3}{c|}{Diffusion (EDM)} & \multicolumn{3}{c}{Score Matching} \\
\textbf{Parameter} & \textbf{10k} & \textbf{30k} & \textbf{100k} & \textbf{10k} & \textbf{30k} & \textbf{100k} & \textbf{10k} & \textbf{30k} & \textbf{100k} \\
\midrule
Batch size & 256 & 4\,096 & 256 & 256 & 256 & 256 & 256 & 256 & 256 \\
Training steps & 100\,000 & 50\,000 & 50\,000 & 50\,000 & 50\,000 & 100\,000 & 50\,000 & 100\,000 & 100\,000 \\
Peak learning rate & $10^{-4}$ & $4 \times 10^{-4}$ & $10^{-4}$ & $10^{-4}$ & $10^{-4}$ & $10^{-4}$ & $10^{-4}$ & $10^{-4}$ & $10^{-4}$ \\
Minimum learning rate & $10^{-6}$ & $4 \times 10^{-6}$ & $10^{-6}$ & $10^{-6}$ & $10^{-6}$ & $10^{-6}$ & $10^{-6}$ & $10^{-6}$ & $10^{-6}$ \\
Warmup steps & 500 & 500 & 500 & 500 & 500 & 500 & 500 & 500 & 500 \\
EMA decay & 0.9999 & 0.999 & 0.999 & 0.999 & 0.999 & 0.9999 & 0.999 & 0.9999 & 0.9999 \\
Single-stream blocks & 16 & 16 & 16 & 8 & 16 & 16 & 16 & 16 & 16 \\
Double-stream blocks & --- & --- & --- & --- & --- & --- & --- & --- & --- \\
Attention heads & 4 & 4 & 4 & 4 & 4 & 4 & 4 & 4 & 4 \\
\midrule
Best C2ST & 0.616 & 0.578 & 0.555 & 0.726 & 0.624 & 0.606 & 0.630 & 0.576 & 0.560 \\
\bottomrule
\end{tabular}%
}
\end{table}

\begin{table}[t]
\centering
\caption{Training configuration for Flux1Joint — Gaussian Linear.}
\label{tab:config_flux1joint_gaussian_linear}
\scriptsize
\resizebox{\textwidth}{!}{%
\begin{tabular}{@{}l|ccc|ccc|ccc@{}}
\toprule
 & \multicolumn{3}{c|}{Flow Matching} & \multicolumn{3}{c|}{Diffusion (EDM)} & \multicolumn{3}{c}{Score Matching} \\
\textbf{Parameter} & \textbf{10k} & \textbf{30k} & \textbf{100k} & \textbf{10k} & \textbf{30k} & \textbf{100k} & \textbf{10k} & \textbf{30k} & \textbf{100k} \\
\midrule
Batch size & 256 & 256 & 4\,096 & 256 & 256 & 256 & 256 & 256 & 1\,024 \\
Training steps & 100\,000 & 50\,000 & 50\,000 & 50\,000 & 50\,000 & 50\,000 & 50\,000 & 100\,000 & 100\,000 \\
Peak learning rate & $10^{-4}$ & $10^{-4}$ & $4 \times 10^{-4}$ & $10^{-4}$ & $10^{-4}$ & $10^{-4}$ & $10^{-4}$ & $10^{-4}$ & $2 \times 10^{-4}$ \\
Minimum learning rate & $10^{-6}$ & $10^{-6}$ & $4 \times 10^{-6}$ & $10^{-6}$ & $10^{-6}$ & $10^{-6}$ & $10^{-6}$ & $10^{-6}$ & $2 \times 10^{-6}$ \\
Warmup steps & 500 & 500 & 500 & 500 & 500 & 500 & 500 & 500 & 1\,000 \\
EMA decay & 0.9999 & 0.999 & 0.999 & 0.999 & 0.9999 & 0.9999 & 0.999 & 0.9999 & 0.999 \\
Single-stream blocks & 8 & 16 & 16 & 8 & 16 & 16 & 8 & 16 & 16 \\
Double-stream blocks & --- & --- & --- & --- & --- & --- & --- & --- & --- \\
Attention heads & 4 & 4 & 4 & 4 & 4 & 4 & 4 & 4 & 4 \\
\midrule
Best C2ST & 0.564 & 0.512 & 0.500 & 0.620 & 0.564 & 0.562 & 0.508 & 0.500 & 0.501 \\
\bottomrule
\end{tabular}%
}
\end{table}

\begin{table}[t]
\centering
\caption{Training configuration for Flux1Joint — Gaussian Mixture.}
\label{tab:config_flux1joint_gaussian_mixture}
\scriptsize
\resizebox{\textwidth}{!}{%
\begin{tabular}{@{}l|ccc|ccc|ccc@{}}
\toprule
 & \multicolumn{3}{c|}{Flow Matching} & \multicolumn{3}{c|}{Diffusion (EDM)} & \multicolumn{3}{c}{Score Matching} \\
\textbf{Parameter} & \textbf{10k} & \textbf{30k} & \textbf{100k} & \textbf{10k} & \textbf{30k} & \textbf{100k} & \textbf{10k} & \textbf{30k} & \textbf{100k} \\
\midrule
Batch size & 256 & 256 & 256 & 4\,096 & 256 & 4\,096 & 4\,096 & 256 & 1\,024 \\
Training steps & 100\,000 & 100\,000 & 100\,000 & 100\,000 & 100\,000 & 100\,000 & 50\,000 & 100\,000 & 100\,000 \\
Peak learning rate & $10^{-4}$ & $10^{-4}$ & $10^{-4}$ & $4 \times 10^{-4}$ & $10^{-4}$ & $4 \times 10^{-4}$ & $4 \times 10^{-4}$ & $10^{-4}$ & $2 \times 10^{-4}$ \\
Minimum learning rate & $10^{-6}$ & $10^{-6}$ & $10^{-6}$ & $4 \times 10^{-6}$ & $10^{-6}$ & $4 \times 10^{-6}$ & $4 \times 10^{-6}$ & $10^{-6}$ & $2 \times 10^{-6}$ \\
Warmup steps & 500 & 500 & 500 & 500 & 500 & 500 & 500 & 500 & 1\,000 \\
EMA decay & 0.9999 & 0.9999 & 0.9999 & 0.9999 & 0.9999 & 0.9999 & 0.999 & 0.9999 & 0.999 \\
Single-stream blocks & 8 & 8 & 8 & 16 & 16 & 16 & 16 & 8 & 16 \\
Double-stream blocks & --- & --- & --- & --- & --- & --- & --- & --- & --- \\
Attention heads & 4 & 4 & 4 & 4 & 4 & 4 & 4 & 4 & 4 \\
\midrule
Best C2ST & 0.522 & 0.515 & 0.513 & 0.825 & 0.513 & 0.516 & 0.510 & 0.504 & 0.501 \\
\bottomrule
\end{tabular}%
}
\end{table}

\begin{table}[t]
\centering
\caption{Training configuration for Flux1Joint — SLCP.}
\label{tab:config_flux1joint_slcp}
\scriptsize
\resizebox{\textwidth}{!}{%
\begin{tabular}{@{}l|ccc|ccc|ccc@{}}
\toprule
 & \multicolumn{3}{c|}{Flow Matching} & \multicolumn{3}{c|}{Diffusion (EDM)} & \multicolumn{3}{c}{Score Matching} \\
\textbf{Parameter} & \textbf{10k} & \textbf{30k} & \textbf{100k} & \textbf{10k} & \textbf{30k} & \textbf{100k} & \textbf{10k} & \textbf{30k} & \textbf{100k} \\
\midrule
Batch size & 256 & 256 & 256 & 256 & 256 & 256 & 4\,096 & 256 & 1\,024 \\
Training steps & 50\,000 & 100\,000 & 100\,000 & 50\,000 & 100\,000 & 100\,000 & 50\,000 & 100\,000 & 100\,000 \\
Peak learning rate & $10^{-4}$ & $10^{-4}$ & $10^{-4}$ & $10^{-4}$ & $10^{-4}$ & $10^{-4}$ & $4 \times 10^{-4}$ & $10^{-4}$ & $2 \times 10^{-4}$ \\
Minimum learning rate & $10^{-6}$ & $10^{-6}$ & $10^{-6}$ & $10^{-6}$ & $10^{-6}$ & $10^{-6}$ & $4 \times 10^{-6}$ & $10^{-6}$ & $2 \times 10^{-6}$ \\
Warmup steps & 500 & 500 & 500 & 500 & 500 & 500 & 500 & 500 & 1\,000 \\
EMA decay & 0.999 & 0.9999 & 0.9999 & 0.999 & 0.9999 & 0.9999 & 0.999 & 0.9999 & 0.999 \\
Single-stream blocks & 16 & 16 & 16 & 16 & 16 & 16 & 16 & 16 & 16 \\
Double-stream blocks & --- & --- & --- & --- & --- & --- & --- & --- & --- \\
Attention heads & 6 & 6 & 6 & 6 & 6 & 6 & 6 & 6 & 6 \\
\midrule
Best C2ST & 0.657 & 0.586 & 0.566 & 0.871 & 0.668 & 0.588 & 0.663 & 0.568 & 0.534 \\
\bottomrule
\end{tabular}%
}
\end{table}

\end{document}